# IMPROVED INDOOR LOCALIZATION WITH MACHINE LEARNING TECHNIQUES FOR IOT APPLICATIONS

## MADDUMA WELLALAGE PASAN MADURANGA

A Thesis Submitted to the School of Postgraduate Studies, IIC University of Technology in Fulfilment of the Requirements for the Degree of Doctor of Philosophy
in Computing

September 2022

# *ABSTRACT*

With the rapid development of the internet of things (IoT) and the popularization of mobile internet applications, the location-based service (LBS) has attracted much attention due to its commercial, military, and social applications. The global positioning system (GPS) is the prominent and most widely used technology that provides localization and navigation services for outdoor location information. However, the GPS cannot be used well in indoor environments due to weak signal reception, radio multi-path effect, signal scattering, and attenuation. Therefore, localization-based systems for indoor environments have been designed using various wireless communication technologies such as Wi-Fi, ZigBee, Bluetooth, UWB, etc., depending on the context and application scenarios. Received signal strength indicator (RSSI) technology has been extensively used in indoor localization technology due to it provides accuracy, high feasibility, simplicity, and deployment practicability features. Various machine learning algorithms have been employed to find the most accurate location from the RSSI-based indoor localization.

This study consists of three phases for the machine learning algorithms to solve the localization issue with different wireless technologies named the supervised regressors algorithms, supervised classifiers algorithms, and ensemble machine learning for RSSI-based indoor Localization.  In addition, a weighted least squares technique and pseudo-linear solution approach are proposed for the closed-form solution. The proposed methods approximate the original system of non-linear RSSI measurement equations with a system of linear equations. Then, the experimental testbed was designed using various wireless technologies with multiple anchor nodes to explore, test, evaluate and compare in a testbed data collection. In each testbed, RSSI values were collected using IoT cloud architectures. The collected data were pre-processed investigating appropriate filters before training the algorithm.

Finally, the received RSSI value is trained using several machine learning models named linear regression, polynomial regression, support vector regression, random forest regression, and decision tree regressor under various wireless technologies. It estimates the geographical coordinates of a moving target node and compares them with each supervised machine learning technique's performance using model evaluation matrices. Consequently, the experiment's performance outcome is expressed in terms of accuracy, root mean square errors, precision, recall, sensitivity, coefficient of determinant, and the f1-score.

# Declaration

I hereby declare that the thesis submitted in fulfilment of the PhD degree is my own work and that all contributions from any other persons or sources are properly and duly cited. I further declare that the material has not been submitted either in whole or in part, for a degree at this or any other university. In making this declaration, I understand and acknowledge any breaches in this declaration constitute academic misconduct, which may result in my expulsion from the programme and/or exclusion from the award of the degree.

Name: Madduma Wellalage Pasan Maduranga

Signature of Candidate:                                    Date: 01 September 2022

# *ACKNOWLEDGMENT*

First and foremost, I would like to express my most profound gratitude to my principle supervisor Professor Ruvan Abeysekara for his comments, criticism, advice, endless patience, inspiration and encouragement, all of which added considerably to my graduate experience. I would like to thank Professor Valmik Tilwari acting as co-supervisor for this thesis. I appreciate his vast knowledge and skills in a multitude of areas, as well as his guidance in directing me towards inspiring research projects.

My gratitude goes also to Professor Leow Chee Seng, head of graduate school and Professor Chanthan Chhuon, Rector, IIC University of Technology. I will never forget the individuals that have been a member of the IICUT group over the years. I'm surprised by how much I've learned from all of you and how far I've come when I think back to our first year together. I am also grateful to our administrative staff for helping me in all sorts of work duties.

Last, but not least, I am indebted to my wife Sarini and my son Niveith for supporting me and sharing my worries, frustrations, and happiness. Without my family's love, support, and understanding, my thesis most certainly would not have been feasible.

# *Table of Content*

## Contents

# List of Figures

# List of Tables

# List of Abbreviations

IoT- Internet of Things

ML-Machine Learning

AI-Artificial Intelligence

WSN-Wireless Sensor Networks

RSS- Received Signal Strength

RSSI- Received Signal Strength Indicator

TDOA-Time Difference of Arrival

AOA- Angle of Arrival

GPS- Global Positioning System

BLE- Bluetooth Low Energy

DT-Decision Tree

ETR-Extra Tree Regressor

LR- Linear Regression

RFR-Random Forest Regressor

RFID- Radio Frequency Identification

NFC- Near Field Communication

TOA- Time of Arrival

AOD- Angle of Departure

LoRaWAN-Long Range Wide Area Network

5G- $5^{th}$ Generation

3G- $3^{rd}$ Generation

V2I- Vehicle-to-Infrastructure

V2V- Vehicle-to-Vehicle

MAC- Media Access Control

kNN- K-Nearest Neighbor

DBM-Deep Boltzmann Machine

DBN- Deep Belief Network

CNN- Convolutional Neural Network

PR- Polynomial Regression

LR- Linear Regression

RL- Reinforcement Learning

PCA- Principal Components Analysis

LBS- Location-Based Services

AAL- Ambient Assistant Living

IPv4- Internet Protocol version 4

UDP- User Datagram Protocol

HTTP- Hypertext Transfer Protocol

MQTT- MQ Telemetry Transport

BAN- Building Area Network

GPRS- General Packet Radio Service

RMSE- Root Mean Squared Error

LoS- line-of-sight

FFNN- Feed Forward Neural Networks

LR- Learning Rate

# CHAPTER 1:

## *Introduction*

*"Information is power. The more you know the better it is. This is true in many cases of our lives. One such case is having good knowledge and understanding of our daily environments. This means to be continuously aware of our own presence in the surroundings around us and have knowledge about "what" & "where" things, people etc. are in our current environment. Context- and location-awareness can be easily achieved with help of deploying a localization system in our chosen surroundings."*

### 1.1. Indoor Localization

Indoor localization is referred to as finding the location of a moving object in location-based IoT applications. Recently, significant research is being carried out in the field of indoor localization. This has led to several indoor positioning systems using different signal technologies for research and commercial purposes. In indoor localization algorithms, signal measurements use as input to the localization algorithm to estimate the algorithm to predict the geographical location of the object. These algorithms could be deterministic, probabilistic, or machine learning-based algorithms. With the recent developments of Artificial Intelligence (AI), many researchers investigated AI applications in many areas, including wireless communications. Hence, machine learning-based indoor localization could consider one of the AI-inspired wireless communication applications. In recent related works, a Machine Learning (ML) based approach has taken place.

## 1.2. Motivation

Indoor localization is perceived as one of the upcoming significant applications used in a wide variety of essential location-based services, such as indoor navigation in airports, hospitals, shopping malls, tracking goods in warehouses, assisted living systems for elderly care…etc. While GPS has become an appropriate standard for outdoor environments, there is no appropriate system for indoor environments. However, the complexity of implementing and maintaining a reliable tracking system can be very challenging.

Traditional deterministic, probabilistic, or filter-based algorithms developed on indoor localization are less accurate, have less robustness, and have complex hardware designs. This thesis focuses on the study of machine learning techniques for improving the accuracy and robustness of indoor positioning systems.

Fluctuation in RSSI is the most challenging problem in localization systems and it effects the location accuracy adversely. The most signals can't advantage of ML is its ability to learn useful information from the input data with known or unknown statistics. For instance, recurrent neural networks could effectively exploit the sequential correlation of time-varying RSSI measurements and use the trajectory information to mitigate RSSI fluctuations.

Machine Learning algorithms have a significant potential to estimate a sensor node's geographical location based on its Received Signal Strength (RSS) measurements with high accuracy.

## 1.3. Background

An indoor localization system uses radio waves, magnetic fields, and acoustic signals to determine the position of people, animals, or moving objects inside indoor environments. The systems are usually linked to sensors that monitor the environment. WiFi is the most widely used platform for indoor localization. However, Bluetooth Low Energy (BLE), Zigbee, and LoRaWAN have been used in many state-of-art IoT applications. The wireless technology will depend on the range, complexity of the system, cost, etc. The methods used to get the position of the device or the user from the RSSI are divided into three main categories: proximity, fingerprinting, and triangulation. The first two use the RSSI values to determine the nearest possible locations of the device or the user.

fingerprinting-based techniques are based on the assumption that a signal map exists for a given indoor environment, and that it can be reconstructed to determine the exact locations of the signals in the area. The localization of an RSSI fingerprint involves two phases: the training phase and the positioning phase. During the training phase, the multi-dimensional vectors of the RSSI values are generated and linked to known locations.

Traditional location techniques such as GPS cannot be used in indoor applications. Its complex requirement of sophisticated equipment and high energy consumption has significantly constrained IoT systems' application scale. Energy-saving has always been a central issue in IoT systems, and long-term application to large-scale applications requires the cost and size of nodes to be as low as possible. To solve this problem, many positioning algorithms developed did not use GPS technology directly, but in some cases as an ancillary method, further focusing on mining the IoT system itself. As per related works, most of the localization algorithms proposed could be generally classified into two categories as range-based localization and range-free localization. The deterministic localization technique usually needs extra hardware to accomplish ranging and then utilizes some algorithm to calculate coordinates.

## 1.4. Problem Statement.

Based on the problem description, research and goal sections discussed earlier; the objective of this graduate project work is to answer the following main research question;

***How to use machine learning techniques to improve the accuracy of indoor localization?***

The following sub-questions are to be answered as well:

1. How to implement an IoT cloud-based architecture for RSSI data collection.

2. How to use linear and non-linear filters approaches to pre-process the RSSI data.

3. How to use supervised regressors to improve the accuracy of the indoor localization

4. How to use supervised classifiers to improve the accuracy of the indoor localization

5. What are the most suitable machine learning regressors for a Wi-Fi-based indoor localization systems?

6. What are the most suitable machine learning classifiers for a BLE-based indoor localization systems?

7. What are the most suitable machine learning regressors for Zigbee-based indoor localization systems?

8. What are the most suitable machine learning regressors for a LoRaWAN-based indoor localization systems?

9. How hyperparameters tuning affects in the localization accuracy?

10. How linear and non-linear filters can improve the localization accuracy.

11. How the number of reference nodes in the network affect to the localization accuracy?

12. How to develop ensemble learning methods to improve the localization accuracy?

## 1.5. Contributions

The main contributions of this research are:

- Design and implementation of state-of-art Wi-Fi and cloud-based testbed to collect RSSI data in an indoor environment.
- Study on using different linear and non-linear filters, weighted least squares technique, and pseudo-linear solution approach for the closed-form solution to improve the localization accuracy.
- Investigate supervised learning techniques(regressors) to improve the localization accuracy of Wi-Fi, Zigbee, and LoRawan-based applications
- Investigate supervised learning techniques(classifiers) to improve localization accuracy.
- Developing a novel hybrid algorithm for indoor localization using ensemble learning techniques.

## 1.6. Objectives

- To improve the localization accuracy using Machine Learning (ML) techniques.

    Location-based services can be considered one of the primary applications in IoT. In such systems, tracking a real-time location of a moving object such as a human, animal, or robot is crucial. Even Though deterministic, probabilistic, or filter-based algorithms exist for indoor localization, machine learning-based techniques deployed are relatively more minor. This research will be surveying opportunities to use machine learning-based techniques for the indoor localization problem over classical localization algorithms.

- To design and implement an experimental testbed.

    Designed and implemented an experimental testbed consisting of multiple sensor nodes designed using microcontroller modules in an indoor environment (8m x 8m room). This testbed will be designed based on Wi-Fi technology as the communication technology. And RSSI values received from sensor nodes will be collected using an IoT cloud-based architecture.

- To investigate using supervised regressors for indoor localization.

    It will be investigated the use of supervised learning regressors including Decision tree (DT), Extra Tree Regressor (ETR), Linear Regression (LR), Polynomial Regression, Random Forest Regression (RFR), Artificial Neural Networks (ANN), and Support Vector Machine (SVR) for accurate node localization problems in IoT systems. It will be considered using the precise

location of a sensor node based on the RSSI level of an IoT application. Collected data from the experimental testbed will be pre-processed using different types of filters, and linearization methods and train the machine learning models. Finally, comprehensively evaluate the performance of each algorithm.

- To investigate using supervised classifiers for indoor localization.

It will be investigated the use of supervised classifiers including Artificial Neural Networks (ANN), Support Vector Machine (SVM), and Nearest Niebuhr(k-NN) for accurate node localization problems in IoT applications. It will be considered using the precise location of a sensor node based on the RSSI level of an IoT application. Collected data from the experimental testbed will be pre-processed, trained, and comprehensively evaluate the performance of each algorithm.

- To introduce an improved hybrid Machine Learning algorithm for indoor localization applications.

It will be developed novel and hybrid localization methods using machine learning ensemble techniques. Such an algorithm will provide a better localization accuracy in robust indoor environments.

## 1.7. Hypotheses

Hypotheses of the research describe as follows.

$H_1$: The supervised regressors can be used for accurate localization in IoT applications in the indoor environment.

$H_0$: The supervised regressors cannot be used for accurate localization in IoT applications in the indoor environment.

$H_2$: Supervised classifiers can be used to improve the accuracy of indoor localization.

$H_0$: Supervised classifiers cannot be used to improve the accuracy of indoor localization.

$H_3$: The Accuracy of the node location depends on the number of reference nodes in the network.

$H_0$: The Accuracy of the node location does not depend on the number of reference nodes in the network.

$H_4$: A hybrid localization model can improve the accuracy of indoor localization.

$H_0$: A hybrid localization model cannot improve the accuracy of indoor localization.

## 1.8. Methodology

Figure 1 illustrates the proposed research methodology to be adopted to achieve the research objectives successfully. A literature review conducted at the initial stage of the research is used to identify opportunities for applying ML for indoor localization. Related works and white papers show that existing algorithms for sensor node localization are highly mathematical and impractical to implement on natural IoT systems. Therefore, it has been decided to investigate ML for the localization problem. To achieve the second objective, a Wi-Fi-based testbed will be designed to collect RSSI values for nodes. Collected data will be trained using different supervised ML algorithms and compared.

Further, it will be investigated ML-based indoor localization for other major wireless technologies such as Zigbee, Bluetooth Low Energy (BLE), and Lorawan. To achieve the third objective, it will be trained to collect data using ANN and evaluate their performances. Where the localization problem is formulated as a classification problem. Moreover, it will be evaluated the effect of the number of reference nodes on the localization accuracy.

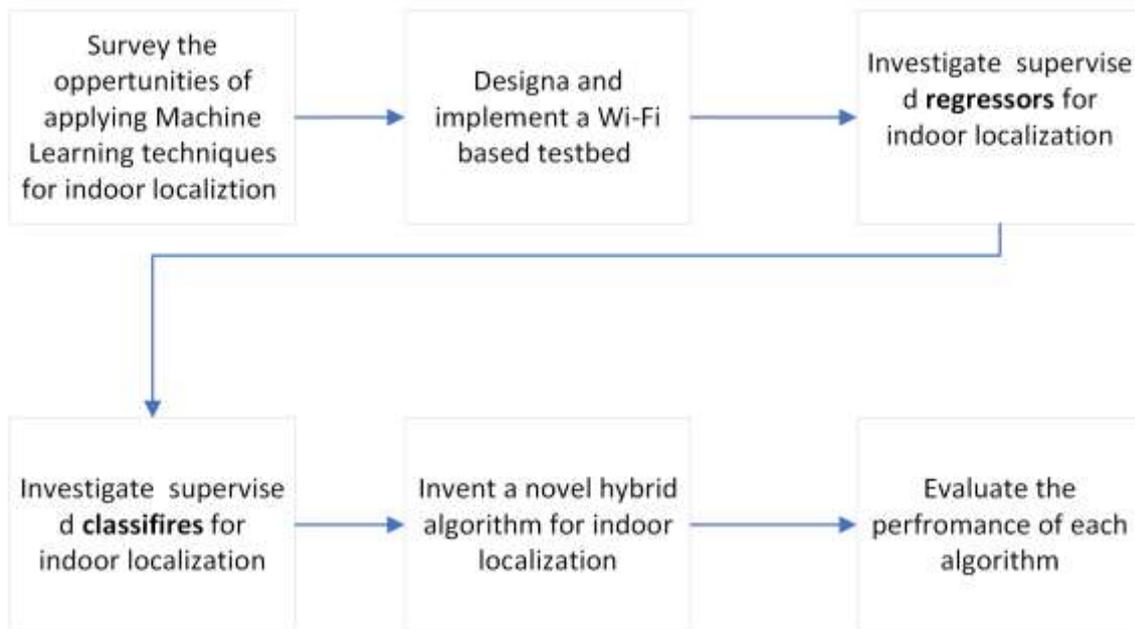

*Figure 1: Research Methodology*

## 1.9. Outline.

The structure of the thesis is given as follows;

**Chapter 2** presents the background theory and the related works to the topic. Further, it will discuss the fundamentals of both supervised and un-supervised machine learning techniques, signal filtering techniques uses in indoor localization, recent related works available in machine learning-based indoor localization, classical localization algorithms such as trilateration, multilateraion, Kalman filters, fingerprinting, etc., and finally, review potential wireless technologies used in indoor localization applications.

**Chapter 3** discusses the designed experimental testbed and other datasets used throughout this thesis. A Wi-Fi-based experimental testbed was designed and implemented using sensor nodes. The system architecture, protocols, and hardware used during the design will be explained step-by-step. Further, it will explain the RSSI data collection process. In addition to the RSSI dataset collected from our testbed, we have used another two publicly available BLE, Zigbee, and LoraWAN datasets in this research. These datasets were used to investigate the localization of the machine learning classifiers and regressors.

**Chapter 4** presents the investigation of machine learning regressors on indoor localization. In this work, RSSI values collected from the testbed discussed in chapter 3 were used to train supervised ML algorithms after pre-processing the data. The Linear Regression (LR), Polynomial Regression (PR), Decision Tree Regression (DTR), Support Vector Regression (SVR), and Random Forest Regressor (RFR) were trained to estimate the accurate positioning of IoT related localization applications. The error between the actual measured values of the position and the estimated values are compared to validate the system model presented.

**Chapter 5** presents a feasibility study of using supervised classifiers for indoor localization applications. Where three supervised classifiers ANN, SVM and kNN will be investigated. In this experiment, It has used an existing RSSI dataset to train the models. The signal strength values received from thirteen different BLE ibeacon nodes placed in an indoor environment were trained using the three algorithms mentioned above. During the experiment, the hyperparameters of each algorithm will be tuned, and observed the localization accuracy.

**Chapter 6** presents a novel algorithm developed for Received Signal Strength (RSS) based on indoor localization. The algorithm is named as TreeLoc(Tree-Based Localization). This novel method is based on ensemble learning trees. Popular Decision Tree Regressor (DTR), Random Forest Regression (RFR), and Extra

Tree Regressor have been investigated to develop the novel TreeLoc method. The results of the TreeLoc method are presented in this paper.

**Chapter 7** concludes the conclusion of all the above works mentioned in different chapters.

# CHAPTER 2:

## 2. Related Works

### 2.1. Internet of Things

The Internet of Things (IoT) technology has revolutionized every aspect of everyday life by making everything smarter. IoT has become more prevalent in recent years due to its vast applications. IoT applications such as smart cities, agriculture, healthcare, and the environment use intelligent data analysis. Today, the IoT concept has been greatly influenced by creating a new dimension in the Internet world. Intelligent processing and big data analysis are the key to developing smart IoT applications in many areas. The internet of things has broadened the scope of presently accessible internet services to include every physical item on the planet. The Internet of Things is based on using numerous items that can communicate with one another, such as RFID, sensors, actuators, mobile phones, and NFC. These things can hear, see, think, and execute different activities via conversing with one another (sharing information). Because of their enhanced functionality, these objects are referred to as "Smart Objects." The Internet of Things (IoT) is helping to enhance people's quality of life and a significant rise in the global economy [1-3].

Machine learning (ML) could be considered a novel way for machines to simulate human learning activities, gain new knowledge, continually improve performance, and achieve unique maturity. In the past few years, ML has been very successful in algorithms, theories, and applications, combined with other agricultural techniques to minimize crop costs and maximize yield. ML applications on agricultural farms can be widely used in disease detection, crop detection, irrigation planning, soil conditions, weed detection, crop quality, and weather forecasting. ML for analyzing the freshness of produce can be found after harvest (Freshness of fruits and vegetables), Shelf life, Product quality, Market analysis, etc.

ML application in IoT-based agriculture could be based on main ML algorithms such as support vector machine (SVM), naive Bayes, discriminant analysis, K-nearest neighbor, K-means clustering, fuzzy clustering, Gaussian mixture models, artificial neural networks (ANN), decision-making and deep learning [7-10].

A Hybrid of ML and the IoT provides an appropriate and controllable environment for growing crops through greenhouse technology. However, the Spatio-temporal variability of crop growth environmental parameters and their mutual effects in protected agriculture makes it difficult for traditional agriculture and environmental regulations to adapt to the growth of different types of plants at different stages of growth. Therefore, higher accuracy is needed from a monitoring and control perspective. Many works exist on designing and testing monitoring and control systems for adjusting temperature and humidity, brightness, $CO_2$ concentration, and other environmental parameters for the Internet of Things, technical and economic results.

It is proposed that controlling the environmental conditions for a specific type of plant can be controlled through IoT, sensors, and actuators. Here the rules of controlling conditions can be made according to an Artificial Neural Network (ANN) set up in the IoT cloud. [11-13]

ML-based yield mapping could apply in farms based on collected data over an IoT network through yield monitoring connected through GPS. The collected data, which reveals the yield details, will be mapped based on the types of farm land. Apart from that, ML systems, together with IoT, can use to predict and improve the yields in agriculture. Farmers rely mainly on agricultural experts to make decisions. Farmers and others use these systems without any knowledge of computer use. ML systems can be used for crop production. This is a knowledge-building system that generates information using existing knowledge. This enables farmers to make economically sound crop management decisions. Various such systems have been developed because of the success of expert systems. The Internet of Things plays an essential role in agriculture. Related works show that ML systems can be built on the IoT and can make recommendations on the use of input data collected in real-time [14-15]

Several ML-based approaches can be applied to soil management. Soil data can be collected from wireless sensor nodes deployed on site. Then, collected data can be fed into ML algorithms to predict and analyze soil properties or classify the types of soil using supervised ML algorithms. Moreover, most commonly used ML algorithms, K-nearest neighbor, support vector regression (SVR), Naive Bayes, etc., can be used to predict soil dryness based on precipitation and evaporative hydrology data [16-18]

Blended ML and IoT can identify and manage diseases in agricultural fields. ML methods further stimulate appropriate pesticides to protect crops from these infections and reduce labor. Such a system assists producers by obtaining statistics and planning fertilizers, pesticides, and irrigation accordingly. By accurately identifying the disease and providing accurate pesticide application and irrigation schemes, grape visibility and volume have been increased and excessive pesticide use reduced—furthermore, architecture with deep learning methods for identifying and classifying speech steps of various plants. The audio steps at these factories are based on real-time captured visual information and travel through different farm areas via IoT-based camera sensor nodes deployed in crop fields [19-20].

Weed management is essential for any farming. Advanced IoT technologies such as NB-IoT can use to handle and manipulate a large amount of data. To optimize this, we propose an unmanned flying machine to take images and map weed in a field. Where flying machines can be controlled through an IoT network. Weed mapping through ML has been investigated [21-22].

Several systems have been implemented to control the water supply for an agriculture field and analyze the water quality using ML [23-24]. It can be developed intelligent systems that detect ground parameters such as soil moisture, soil temperature, and environmental conditions using IoT sensors. Then, use the same data to predict outdoor relative humidity. In addition, we can use hybrid machine learning and IoT systems to control water temperature and adjust to ambient temperature intelligently.

Animal tracking in an agriculture field is really important. Several types of research have been conducted on tracking animals using IoT-based sensors, and independent research is carried out on animal type classification [25-26]. Together IoT and ML solutions could solve this problem efficiently. Sensing the presence of an animal can be detected through IoT sensors. The traced animal can be classified and study their living patterns and movements using ML techniques.

## 2.2. Ranging Technologies for Indoor Localization

The position estimation is a compulsory component of location-based IoT applications [2], especially in the tracking and monitoring applications such as [2]: animal behaviors monitoring, elderly care applications, and smart cities. Besides, the location information enables various emerging applications such as inventory management, intrusion detection, road traffic tracking, health monitoring, etc. [3]. Existing methods for determining position are usually based on geometric calculations such as trilateration (by measuring the angle to a fixed point or node with a known position) or trilateration (by measuring the distance between nodes) [4].

Several techniques can be employed to determine the distance between two nodes, such as synchronization, RSSI, and the physical characteristics of the carrying wave [5]. The localization techniques in the IoT systems can be free of a previous position determination in the network, relying on a few specific sensors' position information and their inter-measurements in the network, such as time difference of arrival, distance, angle of arrival, and connectivity [5]. Previously known location sensors, called anchor points or references, either obtain a position via the Global Positioning System (GPS) or place an anchor point at a point with known coordinates.

Traditional location techniques such as GPS cannot be used in indoor IoT applications. Its complex requirement of sophisticated equipment and high energy consumption has significantly constrained IoT systems' application scale. Energy-saving has always been a central issue in the Internet of Things systems, and long-term application to large-scale applications requires the cost and size of nodes to be as low as possible. To solve this problem, many positioning algorithms developed did not use GPS technology directly but, in some cases, as an ancillary method, further focusing on mining the IoT system itself. Up to now, most of the localization algorithms proposed in the literature could be generally classified into range-based localization [6-8] and range-free localization [9-11]. The former technique usually needs extra hardware to accomplish ranging and then utilizes some algorithm to calculate coordinates.

- Time of Arrival (TOA)

Time of Arrival (TOA)-based positioning systems require additional hardware to ensure synchronization between transmitting and receiving devices. Otherwise, minor timing errors can lead to significant distance estimation errors.

In ToA, however, processing and synchronization times impact distance measuring. A few approaches have been developed to eliminate the temporal synchronization error, such as the symmetric double-sided two-way ToA ranging [33]. This method averages out the error by considering multiple back-and-forth rounds of signal propagation between the nodes.

- Time Difference of Arrival (TDOA)

The Time Difference of Arrival (TDOA) system has the same drawbacks as the TOA system. That means expensive hardware is required. Besides, TDOA uses ultrasonic ranging technology. Ultrasound transmission distances are only 20 to 30 feet, so density deployment is needed. The AOA positioning system can be seen as a complementary technology to TOA and TDOA. This allows the node to estimate the distance based on the relative angle. This can be achieved by installing an angle measuring device [12-15].

- The angle of Arrival (AOA)

For position estimation, the AoA approach exploits the angle a signal creates with an antenna array. This is a more advanced range method. Because both angle and distance measurements are employed, two anchor nodes are usually sufficient for position estimation. However, one disadvantage of this technology is that it necessitates antenna arrays, which makes it complex and costly. This method can also use the time difference of arrival of the signal at specific antenna elements, but this requires much more sophisticated gear and precise calibration.

- Channel-State-Information (CSI)

An enhanced ranging technique is Channel-State-Information. The CSI technique can precisely approximate the received signal over the whole signal bandwidth. This is far superior to RSSI, which simply obtains a single amplitude value for the received signal. Multiple antennas are usually required for CSI, and the channel frequency response received by each antenna must be approximated. CSI can provide both the magnitude and phase of the channel response and can be used for both range-based and range-free localization.

- Received Signal Strength Indicator (RSSI)

On the contrary, RSSI technology overcomes most of the above drawbacks. Thus, it doesn't need additional hardware devices. However, this technology typically suffers from multipath fading, noise interference, and irregular signal propagation, which seriously impacts distance estimation accuracy. We believe that only by taking appropriate measures can the positioning accuracy be improved to meet the requirements of most application systems. This goal is achieved through regular node deployment, space partitioning, and localization optimization. Based on a particular signal propagation model, RSSI positioning systems must use received signal strength to pre-estimate the distance between transmitter and receiver. Several signal propagation models, whether theoretical or empirical, are used to convert signal strength to distance. The relationship between RSSI and the distance between transmitter and receiver is expressed as per Eq. 1 and 2.

The relationship between the Received Signal Strength Indicator (RSSI) and the distance is the key to any wireless-based ranging and localization system. RSSI-based position systems use three kinds of propagation models free space model, bidirectional surface reflection, and log-normal shadowing (LNSM) [9]. As an improved RSSI-based ranging model, we can use the LNSM, which can use for practical applications [11]. It can be defined as;

$$P_r \propto \frac{P_t}{d^\alpha} \quad (1)$$

where; Pr - Received Signal Pt - Transmit Signal Power d – distance

$\alpha$ – distance power gradient can be found using a table (specific for the environment). According to [8], Eq. 2 can be simplified to get the improved version of LNSM to show the relationship between the RSSI and distance.

$$RSSI = -(10n\log 10d + A) \quad (2)$$

Where; d – the distance from the mobile/target node to the reference node, n - Signal propagation constant, and A - Received Signal strength at 1m distance.

Where n is the signal propagation constant, d is the distance from the sender or Euclidean distance. A represents the received signal strength at a distance of 1 meter.

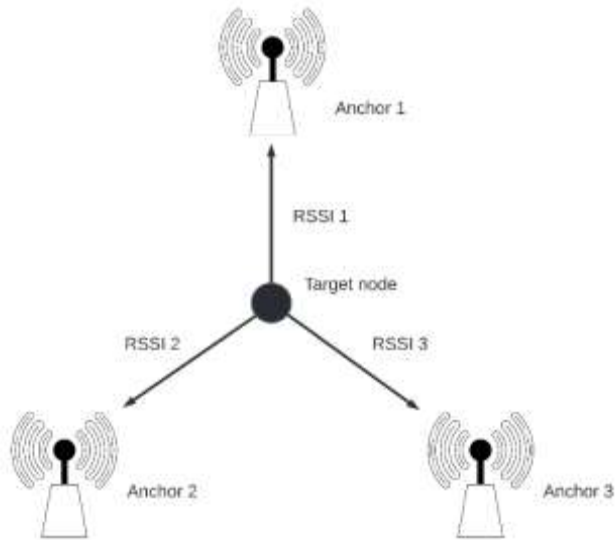

*Figure 2: RSSI measurements*

Fig.2 shows how anchor nodes and the target nodes are placed in indoor positioning systems. Moreover, multipath fading is responsible for the unreliability of wireless transmission. Multipath fading creates more interferences, such as link quality and path loss model predictions. Therefore its real impact on wireless communication is always taken into account. Although multipath fading is a deterministic event, analytical models show that it may also be described as a probabilistic phenomenon. Furthermore, in the case of static nodes, multipath fading is time-invariant. Because the nodes are frequently static, static multipath fading is critical. This is a common feature of wireless sensor networks. Shadowing is the significant divergence of a radio frequency signal from its mean caused by large obstructions. Deep fading occurs when a receiver enters a shadow zone. In the indoor environment, such as businesses and residences, multipath fading has a significant impact. Because radio waves are reflected by buildings, trees, and other aspects of the landscape, the outdoor environment is not immune if wireless sensor nodes are put outside of the structure. As a result, multipath fading and shadowing significantly impact wireless networks and must be taken into account while deploying a wireless sensor network. Table 1 explain the prons and cons of each distance measurement techniques.

*Table 1:Advantages and disadvantages of different distance measurement techniques*

| Technique | Advantages | Disadvantages |
|---|---|---|
| RSSI | A simple technique for distance measurement as there is no need for extra hardware. Cost-effective and less complex. | Provides low precision due to NLOS propagation of the signal. Robust optimization and positioning techniques are required for accurate results. |
| CSI | This has high granularity over RSSI due to both amplitude and phase information of channel frequency. | Complexity is higher than the RSSI-based and most other localization systems. |
| ToA | ToA provides the highest measurement precision among most range-based methods with strict clock synchronization between the transmitter and the receiver. | Expensive due to different devices and modules needed to lower synchronization error. |
| TDoA | TDoA can provide better precision than ToA | It is expensive due to extra hardware and module added to lower the synchronization error. |
| AoA | AoA provides more accuracy for the target node localization estimation for the short signal propagation distance. | Requires antenna arrays and extra hardware. It is often hard to implement AoA due to the multipath effect. |

## 2.3. Localization Algorithms

Localization algorithms can be classified as deterministic, probabilistic, or learning-based algorithms. Deterministic methods can estimate localization faster by using classical measurement techniques like Trilateration and Triangulation. They use a simple signal strength map in which each location has a list of objects, humans, or services within range and an average signal strength value. Deterministic methods, despite using a more straightforward calculation than probabilistic methods, estimate the location by precisely considering where the object is located in the current measurement. Because probabilistic methods require much information than deterministic methods, they perform more accurate location detection compared to the deterministic methods.

However, they are ineffective for small-capacity devices such as sensors because they have computing density. Particle Filters, Kernel Method, Histogram Method, and Hidden Markov Model Methods are some probabilistic methods [13-16].

Measurement algorithms that can be used for localization; Trilateration, Triangulation, Scene Analysis, Dead Reckoning, Proximity, and Hybrid Algorithms are researched

- Trilateration Algorithm

The trilateration method measures the distance to the station from mobile devices. They need to know at least three mobile devices' locations for the first step of an efficient localization [10-11]. Received Signal Strength (RSS), Time of Arrival (ToA), Time Difference of Arrival (TDoA) are some of the measurement techniques of Trilateration Algorithms.

According to Fg.3, suppose a point of interest is located on the surface of three intersecting spheres in a 3D space. In that case, it is possible to limit the number of potential locations for the point to no more than two by knowing the positions of the spheres' centers and their radii. As per Fig.3, The equation for a sphere with its center at ($x_a$, $y_a$, $z_a$) is shown in the general case first.[12]

$$r^2 = (x - x_a)^2 + (y - y_a)^2 + (z - z_a)^2 \quad (3)$$

As a result, the equations can be expressed using the sphere radii of $r_1$, $r_2$, and $r_3$.

$$r_1^2 = x^2 + y^2 + z^2 \quad (4)$$

$$r_2^2 = (x - x_2)^2 + y^2 + z^2 \quad (5)$$

$$r_3^2 = (x - x_3)^2 + (y - y)^2 + z^2 \quad (6)$$

Simplifying equations (4–6) yields the following solutions:

$$x = \frac{r_1^2 - r_2^2 + x_2^2}{2x_2} \quad (7)$$

$$y = \frac{r_1^2 - r_3^2 + x_3^2 + y_3^2 - (2x_3 x)}{2y_3} \quad (8)$$

$$z = \sqrt{r_1^2 - x^2 - y^2} \quad (9)$$

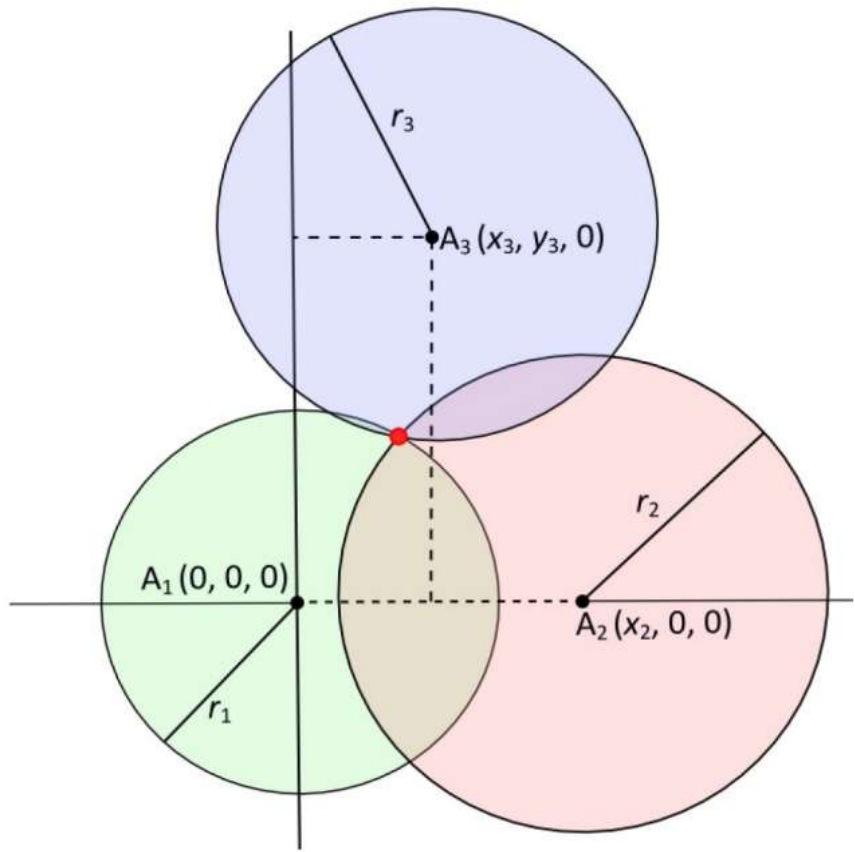

*Figure 3:Trilateration Algorithm*

- Triangulation Algorithm

These algorithms are particularly well-suited for usage in Line of Sight communication. The angle of Arrival (AoA) or Angle of Departure (AoD) measurements are used in many Triangulation Algorithms. The position is established by the angle between the mobile user's signal and the anchor nodes. If the signal's angle changes due to multipath, such methods may be inefficient [13-15].

- Fingerprinting Algorithm

Fingerprinting is a well-known method for Scene Analysis. Algorithms for fingerprinting are divided into two phases: training and localization. Typically, the training phase is regarded offline, whereas the detecting phase is considered online. Before localization, some offline location data is collected. Signal strength may vary in this technique due to communication issues such as fading, interference, or multipath.

Additionally, environmental variables may change, necessitating retraining. Localization is done during the online phase. The localization process compares the signal intensities during the online and offline phases [13,15].

- Proximity

Proximity measurements aim to determine if two devices are 'linked' or 'in-range.' Localization costs and effort may be reduced in local and interior areas by using locations based on proximity to recognized reference points combined with widely distributed wireless technologies [15].

- Dead Reckoning

Sensors utilize it in particular. The localization algorithm uses velocity or acceleration based on the last known or predicted position. It employs an internal navigation system that provides exact location data. Every measurement of a place is carried out utilizing outdated methods. As a drawback, measurement inaccuracies may rise with each step [15].

- Hybrid Algorithms

Probabilistic and deterministic algorithms are often combined to produce hybrid algorithms. Interference and fading are typical issues with indoor multipath localization. Furthermore, each structure has its own set of drawbacks. Using probabilistic and deterministic algorithms to create an efficient localization method may improve localization accuracy [15-17].

## 2.4. Wireless Technologies

### 2.4.1. Wi-Fi

Unlike other indoor positioning technology solutions like BLE beacons or RFID, Wi-Fi does not require additional hardware or infrastructure maintenance. Wi-Fi infrastructure is currently present in almost all locations, offering a basic level of locating capability without additional investment. Manually deployable hardware solutions can be expensive and time-consuming, especially if numerous locations need the functionality. As a result, other technologies require a larger budget in both money and time to create an effective environment for indoor positioning [18-19].

Wi-Fi is a global indoor positioning solution that can be scaled with little to no manual intervention. For instance, implementing Wi-Fi positioning in 25 warehouses compared to manually deploying BLE beacons or RFID in 25 warehouses saves time and resources. In addition, because Skyhook maintains a global map of over 5 billion Wi-Fi access points, your positioning system will work around the globe, indoors and outdoors. While Wi-Fi location allows for immediate access, companies have various alternatives for improving accuracy depending on the use case. The placement accuracy of the system may be improved by conducting a simple survey to establish the exact positions of all Wi-Fi APs inside a facility. The business may install more Wi-Fi access points across its premises if more precision is needed. While this is an extra cost, it does not need the client to learn, deploy, or maintain yet another technology inside their infrastructure. This may be done as part of a larger Wi-Fi infrastructure upgrade project or by simply installing more Wi-Fi access points in locations where positional precision is needed [18-22].

## *2.4.2. Bluetooth Low Energy (BLE)*

Bluetooth Low Energy (BLE) has recently attracted several researchers to exploit its capabilities beyond location-based services and applications [23-24]. The ability of the BLE to operate at multiple low transmission power levels, long transmission range, low energy consumption, long battery life, small size, and low cost [24] have allowed BLE to be embedded in several consumer electronics [25]. Due to its ubiquitousness, BLE is now widely considered for wireless Indoor Localization Systems (ILS) [26].

## *2.4.3. Zigbee*

ZigBee is an IEEE 802.15.4-based standard for tiny, low-power devices. ZigBee is a wireless technology used in applications with low data rates, long battery life, and secure networking. With a data rate of 250 kbit/s, ZigBee is best suited for transmitting periodic or intermittent data and a single signal from a sensor or input device. Depending on the power output and circumstances, the transmission range of ZigBee may vary from 10 to 100 meters [27].

Tree, star, and mesh network topologies are all supported by ZigBee. The mesh topology aids ZigBee's ability to create a stable network by allowing data to be sent through many routes. Even if one of the routers or paths is down, ZigBee may send data to its destination. In the most recent version of ZigBee, the addressing system can handle up to 64,000 nodes per network, and several network coordinators may be linked together to enable large networks [27-28].

The three functions of ZigBee are ZigBee coordinator, ZigBee Router, and ZigBee end devices. In the network system, each position has a distinct purpose. The most comprehensive gadget is the ZigBee coordinator. Every system or network must have at least one coordinator to manage the network and handle data transfer. The ZigBee router's job is to establish a link between the coordinator and the end device. It transmits data to the next router and executes code at the application level. The basic purpose of a ZigBee end device is to relay data to the coordinator and router [27-28].

### 2.4.4. LoRaWAN

Long Range Wide Area Network (LoRaWAN) is being increasingly seen [29] as a complementary technology to 5G. It allows Internet-of-Things (IoT) applications for smart cities, namely Intelligent Transportation Systems (ITS), as well as machine-to-machine (M2M) applications that don't need the high bandwidth or ultra-low latency of 5G networks. Furthermore, we must remember that LoRaWAN is currently accessible and ready to be implemented with various low-cost hardware and software options. This bridge solution may allow businesses to build their LoRaWAN-based solution while waiting for 5G to become generally accessible.

There are already a significant number of projects focusing on the applicability of LoRaWAN to the automotive sector. Li et al. [30], for example, examine the performance of this technology when it is utilized for Vehicle-to-Infrastructure (V2I) and Vehicle-to-Vehicle (V2V) communications. The authors of [4] offer a new architectural and communication design paradigm for integrating automotive networking clouds with IoT. The goal is to create real-world apps that offer IoT services through vehicle clouds. The author of this essay focuses on smart city applications that would be deployed, managed, and controlled through LoRaWAN-based vehicle networks. In [31], a solution in the field of vehicular IoT is presented that enables IPv6 end-to-end connection between a moving and an Internet node through an adaptation process conducted in a Multi-Access Edge Computing (MEC) node connected to the Low Power-Wide Area Network (LP-WAN) gateway.

## 2.5. Machine Learning for Indoor Localization

Machine learning algorithms can search for patterns and regularities in any data set and are widely used in various application areas. The more data available, the bigger the problem can be solved. These algorithms are usually implemented in two steps. In the first training phase, the data is collected and provided to the algorithm to train patterns, build models to classify the data, and predict data attributes. In the second phase (the test phase), new data is tested based on the model established in the training phase to reveal the model's validity. This two-step learning algorithm is called a supervised learning algorithm. It automatically learns from the data by generalizing the example. Some algorithms do not use the test phase. These algorithms are called unsupervised learning algorithms. These algorithms use unlabeled data to cluster the data into different classes. Machine learning algorithms can be used for classification or regression.

In classification, machine learning algorithms classify data into different categories, while regression learns from training data to predict continuous variables. To improve indoor the accuracy and reliability of indoor localization in the previous work, we propose investigating both supervised and unsupervised ML algorithms for indoor localization problems [31-36].

## 2.6. Machine Learning Algorithms.

A labeled training data set (i.e., predefined inputs and known outputs) is used to train the machine learning model in supervised learning. This model represents the learned relationship between the input, output, and system parameters. In this subsection, the major supervised learning algorithms are discussed in the context of IoT. In fact, supervised learning algorithms are extensively used to solve several challenges in WSN systems such as localization and objects tracking(e.g., [37]), event detection and query processing (e.g., [38-40]), media access control (MAC) (e.g., [41-43]), security and intrusion detection-based applications (e.g., [44], [45], [46]), and quality of service (QoS), data integrity and fault detection (e.g., [47], [48]).

### 2.6.1. K-nearest neighbor (k-NN):

K-Nearest Neighbor (kNN) is the most basic and most uncomplicated of all indoor localization algorithms. It is an enhanced version of the fingerprinting method that preform location determination utilizing RSSI value. The kNN algorithm requires a predefined table that references the strength of two or more different received signals from local routers. The predefined table is constructed during the offline phase [49].

For distance metrics, we will use the Euclidean metric

$$d(x, x') = \sqrt{(x_1 - x'_1)^2 + \cdots + (x_n - x'_n)^2} \quad (10)$$

Finally, the input x gets assigned to the class with the largest probability.

$$P(y = j | X = x) = \frac{1}{K} \sum_{i \epsilon A} I(y^i = j) \quad (11)$$

During the offline phase, specific coordinate points will be plotted on a sample indoor floor map. Several signal strength readings will be done at each specific point over a set period. The average signal strength for all routers referenced over a certain period is called a fingerprint vector for that particular location. The readings for all referenced routers (c1, c2, c3…) are the fingerprint vector Cj (cj1, cj2, cj3…). This process is then repeated for all specific coordinate points on the floor map.[50-52]. Fig.4 denotes how the distance measurement are trainined using the kNN algorithm.

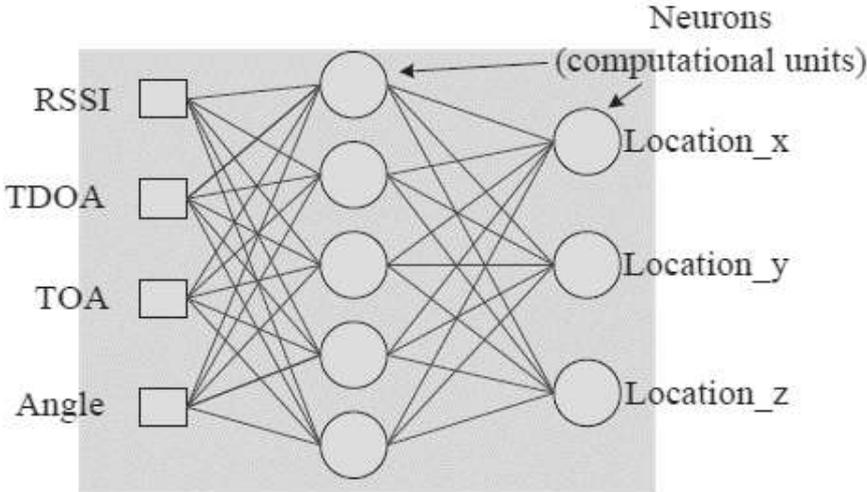

*Figure 4:K-nearest neighbor-based localization*

## 2.6.2. Neural networks (NNs):

This learning algorithm can be built by cascading a chain of decision-making units (e.g., perceptron's or radial basis functions) to recognize nonlinear and complex processes [53]. The use of neural networks in distributed schemes is still less common in Internet of Things systems due to the high computational complexity and increased management overhead for learning network weights. However, the neural network can simultaneously learn multiple outputs and decision boundaries [54]. This makes it suitable for solving various network challenges using the same model.

The fundamentals of working NNs are as follows. Multiply the input value $x_i$ by the weights $w_i$ For each input, then add up all of the multiplied values. A given input's influence on a neuron's output is determined by its weights, representing the strength of the connections between neurons. The input $x_i$ will have a more significant impact on the output than the weight $w_2$ if the weight $w_1$ is greater than the weight $w_2$ And vice versa.

$$\Sigma = (x_1 \times w_1) + (x_2 \times w_2) + \cdots + (x_n \times w_n) \quad (12)$$

The row vectors of the inputs and weights are x = [x$_1$, x$_2$, …, x$_n$] and w = [w$_1$, w$_2$, …, w$_n$], respectively, and their dot product is given by

$$w.x = (x_1 \times w_1) + (x_2 \times w_2) + \cdots + (x_n \times w_n) \quad (13)$$

Hence, the summation is equal to the *dot product* of the vectors *x* and *w*

$$\Sigma = x.w \quad (14)$$

The sum of the multiplied values with bias b added is what we'll refer to as z. Bias, often referred to as the offset, is typically required to shift the entire activation function to the left or right to get the desired output values.

$$z = x.w + b \quad (15)$$

Give a non-linear activation function the value of z. Without activation functions, the neural network would simply be a linear function. Activation functions are employed to add nonlinearity to the output of the neurons. They also significantly affect how quickly the neural network learns. The activation function of perceptrons is a binary step function. As our activation function, we'll utilize the sigmoid, also known as the logistic function.

$$y = \sigma(z) = \frac{1}{1+e^{-z}} \quad (16)$$

where **σ** denotes the s*igmoid* activation function, the output we get after the forward prorogation is known as the *predicted value* **y**.

Regarding indoor localization, node placement can be based on measurements of the propagation angle and distance of the received signal from the anchor node. Such measures may include the received signal strength indicator (RSSI), time of arrival (TOA), and time difference of arrival (TDOA). Neural network-related algorithms include self-organizing graphs. After supervised training, the neural network generates estimated node positions as vector-valued coordinates in 3D space.

These are the latest versions of Artificial Neural Networks (ANN) that can take advantage of today's massive data-driven problems in almost every field. It uses more extensive neural networks to solve semi-monitoring, where most of the large amounts of data are unlabeled or unclassified. Examples are Deep Boltzmann Machine (DBM), Deep Belief Network (DBN), Convolutional Neural Network (CNN), and Stacked Autoencoder. Deep learning algorithms have been investigated in many areas in WSN and IoT, such as data streaming, security, energy management, Detecting the multi-queue evolution pattern, Deciding the queue size for the RED zone, Combining the congestion control scheme with other schemes and protocols, outlier detection, routing, etc. [55-58].

## 2.6.3. Support Vector Machines (SVMs):

The SVM is a machine learning algorithm that uses labeled training samples to learn the classification of data points. [59]. For example, one way to detect malicious behavior on a node is to use SVMs to investigate data's temporal and spatial correlation.

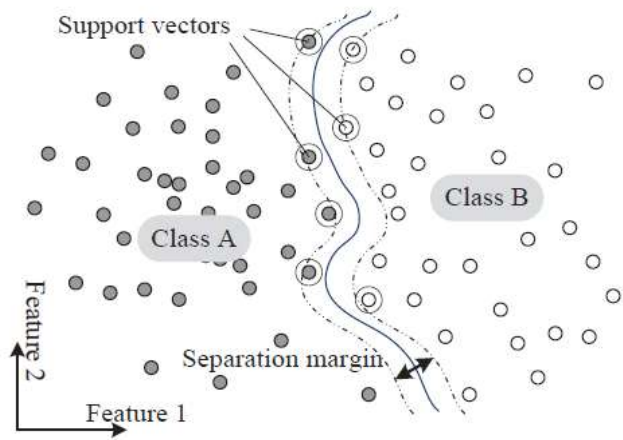

Figure 5:Support Vector Machine.

The WSN observation is used as a point in the element space, and the SVM divides it into multiple parts. These parts are separated by the widest possible margin (the separation gap), and the new readings are sorted according to which side of the gap they are on, as shown in Fig.5. An SVM algorithm uses linear constraints to optimize a quadratic function (the problem of building a set of hyperplanes). It provides an alternative to multi-layer neural networks with non-convex, unconstrained optimization problems. [60-63].

## 2.6.4. Support Vector Regression (SVR)

Support vector regression (SVR) has been extensively applied in precis location estimation applications. The method effectively solves non-linear problems even with a small sample of training datasets. SVR adopts the structural risk minimization principle, which minimizes an upper bound of the generalization error comprising the sum of the training error and a confidence level [45]. This principle is different from the traditional empirical risk minimization, which only minimizes the training error. The basic concept of SVR applied to regression problems is to introduce kernel function, map the input space into a higher dimensional feature space via a non-linear mapping, and perform a linear regression in this feature space.

The SVR model relies on a training subset, ignoring data close to the model's prediction (within a threshold ε). SVR depends on the choice of kernel and relevant parameters to solve the regression problem. The kernel used for this study is the radial basis function (RBF). One of the strengths of SVR is its high dimensional space, which relies on the input space dimensionality. SVR uses a linear function, also called the SVR equation, for non-linear mapping of the imported data into higher dimensionality. Assuming normalized input variables consist of a vector Xi, Yi is the valuable solar thermal energy (i represents the data-point in the dataset). In this case, a set of data points can be defined as $\{(X_i, X_j)\}_{i=1}^{N}$, where N is the total number of samples. An SVM regression approximates the function using the form given in Equation (16) The SVR equation is presented in Equation 5).

$$Y = f(X) = W.\Phi(X) + b \quad (16)$$

In Equation (1), $\Phi(X)$ denotes the high-dimensional space. A regularized risk function, given in Equation (17), is used to estimate coefficients W and b

$$\text{Minimize}: \frac{1}{2}\|W\|^2 + C\frac{1}{N}\sum_{i=1}^{N} L_\varepsilon(Y_i, f(X_i)) \quad (17)$$

$$L_\varepsilon(Y_i, f(X_i)) = \begin{cases} 0 & |Y_i - f(X_i)| \leq \varepsilon \\ |Y_i - f(X_i)| - \varepsilon, & \text{others} \end{cases} \quad (18)$$

$\|W\|^2$ is known as the regularized term, and C is the penalty parameter to determine the model's flexibility. The second term of Equation (17) is the empirical error and is measured by the ε-intensity loss function (Equation (18)). This defines an ε tube shown in Fig. 1. If the predicted value is within the tube, the loss is zero. If it is outside the tube, the loss is the magnitude of the difference between the predicted value and the radius ε of the tube [23]. To estimate W and b, the above equation is transformed into the primal objective function given by Equation (18)

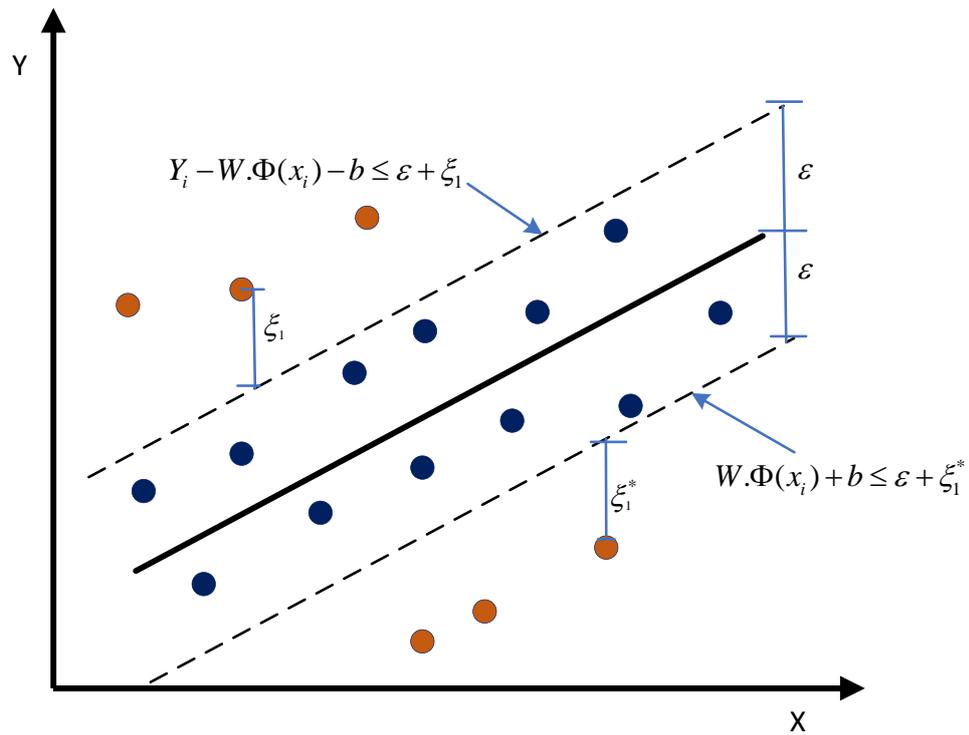

*Figure 6: Support Vector Regression (SVR)*

$$\underset{\xi_1,\xi_1^*,W,b}{\text{Minimize}}: \frac{1}{2}\|W\|^2 + C\frac{1}{N}\sum_{i-1}^{N}(\xi_1-\xi_1^*) \quad (19)$$

$$\text{Subject to}: \begin{cases} Y_i - W.\Phi(x_i) - b \leq \varepsilon + \xi_1 \\ W.\Phi(x_i) + b \leq \varepsilon + \xi_1^*, \quad i=1,2,...,N \\ \xi_1 \geq 0 \quad \xi_1^* \geq 0 \end{cases} \quad (20)$$

In the above equations, $\xi_1$ and $\xi_1^*$ are the slack variables. Equation (4) is written as below by introducing the kernel function.

$$\underset{\{\alpha_i\},\{\alpha_{Xi}^*\}}{\text{Minimize}}: \frac{1}{2}\sum_{i=1}^{N}\sum_{j=1}^{N}(\alpha_i-\alpha_i^*)(\alpha_j-\alpha_j^*).k(X_i,X_j) - \varepsilon\sum_{i=1}^{N}(\alpha_i-\alpha_i^*) + \varepsilon\sum_{j=1}^{N}(\alpha_j-\alpha_j^*) \quad (21)$$

$$\text{Subject to}: \begin{cases} \sum_{i=1}^{N}(\alpha_i-\alpha_i^*) = 0 \\ \alpha_i, \alpha_i^* \in [0,C] \end{cases}$$

In Equation (21) $\alpha_i, \alpha_i^*$ are Lagrange multipliers, i and j are different samples. Therefore, Equation (22) becomes.

$$Y = f(X) = \sum_{i=1}^{N}(\alpha_i-\alpha_i^*)k(X_i,X_j) + b \quad (22)$$

The design parameters of ϵ-SVR are the maximum tolerable error ϵ at the output, the regularization parameter C, the number of training patterns N, and the parameter σ of the kernel function.

## 2.6.5. Decision Tree Regressor (DTR):

Decision trees are supervised learning algorithms that can be used for classification and regression problems, but they are the recommended way to solve classification problems in most cases. This is a tree structure classifier, with internal nodes representing the characteristics of the dataset, branches representing decision rules, and each leaf node representing the result. The decision tree has two nodes, a decision node and a leaf node. Decision nodes are used to make decisions and have multiple branches, but leaf nodes are the output of these decisions and do not contain any other branches. Make decisions or tests based on the characteristics of a particular dataset, as shown in the Fig.7.

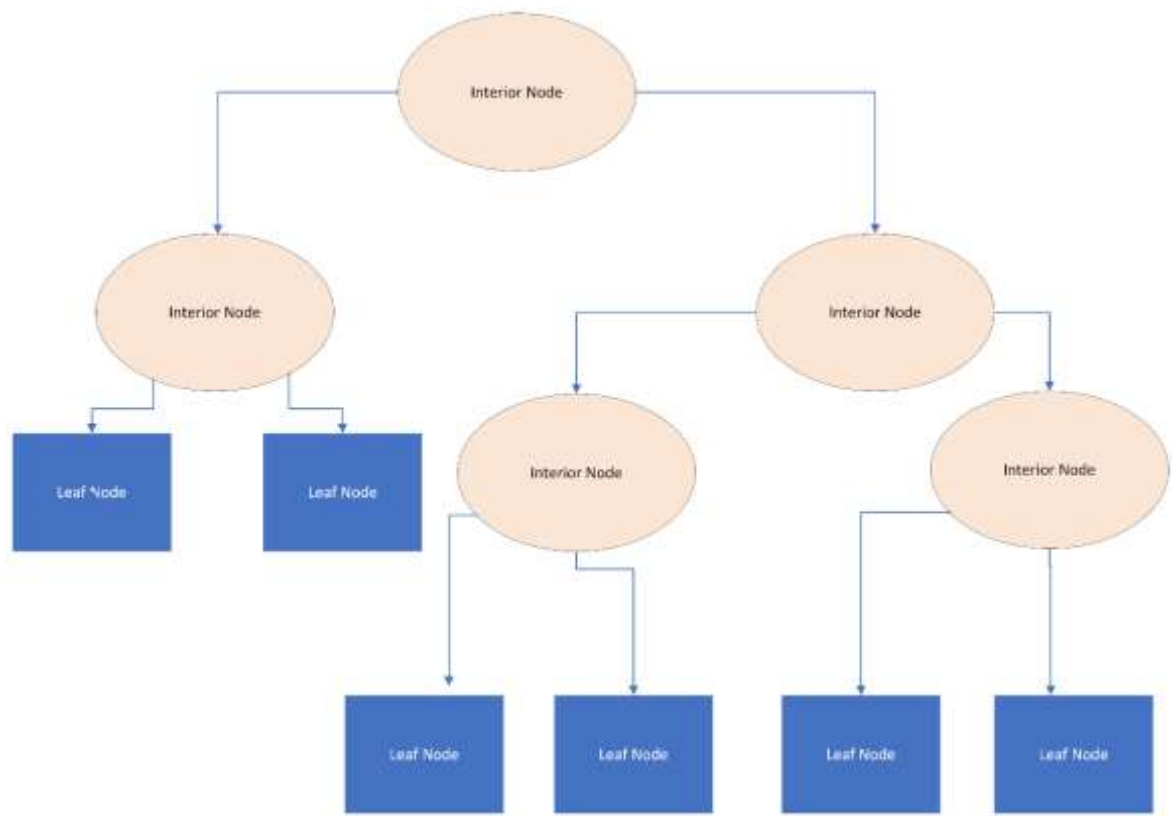

*Figure 7: Decision Tree Hierarchy*

This graphical representation is used to get all possible solutions to a problem/decision based on one specific condition. It is called a decision tree because it resembles a tree. Start with the root node and extend the branch to build a tree structure. This algorithm has been used for selecting cluster heads, real-time data mining, and anomaly detection in WSNs [64-67].

## 2.6.6. Linear Regression (LR)

The linear regression model can study the relationship between the output dependent and input variables. An essential premise of this method is that the output variable is a linear combination of a particular weight and an input variable. These linear models are established during the training phase and are used to make predictions during the testing phase.

A fundamental technique like Linear Regression (LR). Where $x$ is the independent variable and $y$ is the dependent variable as expressed in the equation?

$$y = \theta_0 + \theta_1 x + \varepsilon \quad (23)$$

Linear regression has been used for predictive modeling, data imputation in IoT systems [67-70]. For nonlinear relationships, these models provide inaccurate predictions. However, productive nonlinear models are also more brutal to derive. It has been observed that linear models may also provide a reasonable accuracy for location estimation.

## 2.6.7. Polynomial Regression (PR)

Polynomial regression is a form of regression analysis in which the relationship between the independent and dependent variables—is modeled on a polynomial of degree n as follows

$$y = \theta_0 + \theta_1 x + \theta_2 x^2 + \theta_3 x^3 .. + \theta_n x^n + \varepsilon \quad (24)$$

Polynomial regression models are usually suitable for least squares. In Gauss-Markov's theorem, the least-squares method minimizes the variance of the coefficients. Polynomial regression is a particular linear regression case that fits polynomial data and has a curvilinear relationship between the dependent and independent variables.

The polynomial regression has been used for data aggregation, intrusion detection applications, and security in WSN and IoT systems [70-72].

## 2.6.8. Random Forest Regressor (RFR)

The random forest algorithm is an integrated algorithm based on a cart decision tree [12]. Typically, the self-help sampling approach is used to get the training samples, and the decision tree is created by dividing the sample eigenvalues. The splitting value can be expressed using entropy, the Gini index, and variance.

Where B represents a feature in the sample data set, H(D) is the empirical entropy of the set D, Qi is the sample subset of the ith class in the training data set, yi represents the label of a sample instance, and μ represents the mean value of the sample instance. In the random forest regression model, each regression tree corresponds to a partition in the feature space and the output value on the partition unit. Assuming that each feature has a value, when a feature's value causes the loss function to be divided, it is doing so in accordance with the minimum loss function concept. The output value of each region is ci , and its expression is

$$\min_{j,s}[\min Loss(y_i, c_1) + \min Loss(y_i, c_2)] \quad [25]$$

Suppose the space is divided into f elements by partition R1, R2, · · · , RF , Then the expression of the regression tree is:

$$f(x) = \sum_{f=1}^{F} c_f I(x \in R_f) \quad [26]$$

Finally, multiple regression trees are formed. Assuming that the set of regression trees is {T1, T2, · · ·, Ts}, when new data is input, each tree will have a prediction value, and the average value of the prediction results of each tree is the final prediction result. The expression is:

$$f(x) = \frac{1}{s}\sum_{i=1}^{s} c_f(x) \quad [27]$$

To sum up, the random forest regression model integrates the prediction of each tree, thus reducing the variance of the model. In addition, the model has strong generalization ability, can tolerate a small number of outliers and missing values, and has a fast training speed. [73-75].

An example of a simple decision tree is as shown in Fig.8.

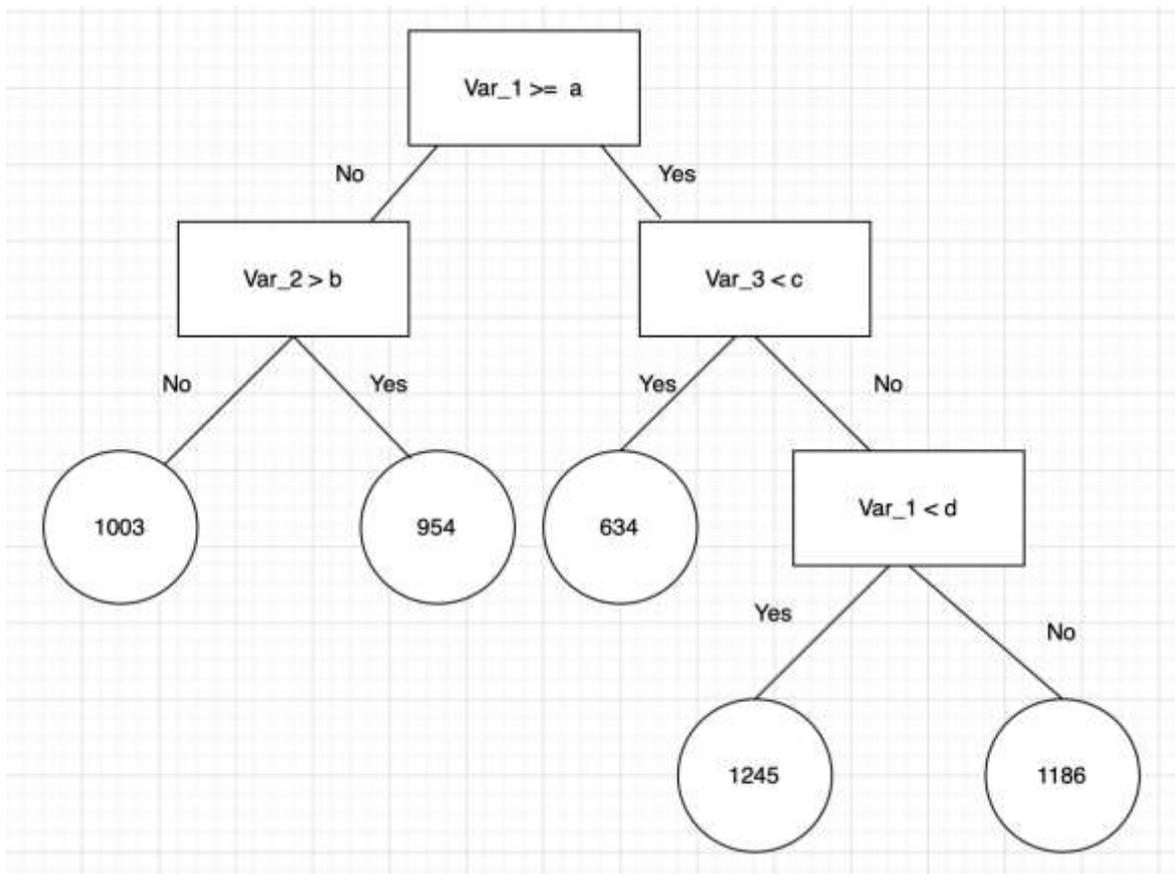

Figure 8: A simple decision tree

## 2.6.9. Unsupervised Learning

Unsupervised learners do not get the labeled data (there is no output vector) in the training phase. The unsupervised learning algorithm divides the sample sets into different groups by investigating the similarities between the sample sets. Not surprisingly, the subject of this learning algorithm is widely used in node clustering and data aggregation issues.

## 2.6.10 Clustering

Clustering is an important concept when it comes to unsupervised learning. This primarily involves finding structures or patterns within a collection of unclassified data. The clustering algorithm processes the data and looks for natural clusters (groups) in the data, if any. You can adjust the particle size of these groups [76]. You can also change the number of clusters that the algorithm recognizes. There are several types of clustering algorithms available:

Hierarchical Clustering: Hierarchical clustering is an algorithm to establish a clustering hierarchy. It starts with all the data assigned to the cluster. If there is only one cluster, the algorithm ends. Here, two close clusters are in the same cluster.

K-means Clustering: K means that it is an iterative clustering algorithm that helps find the maximum value for each iteration. First, select the required number of clusters. This clustering method requires clustering the data points into groups of k. The smaller k is, the larger the group, but the smaller the particle size—similarly, the larger k, the smaller the group and the smaller the granularity. K-means clustering has been used in routing optimization in IoT systems [77-78].

Assign data points to one of the k groups. In k-means clustering, each group is defined by creating a centroid for each group. Centroids are like the centers of a cluster, capturing the points closest to them and adding them to the collection. The output of the algorithm is a set of "labels."

Fuzzy Clustering: Fuzzy clustering is a method discussed under unsupervised learning. Also, Fuzzy clustering is called soft k-clustering. The clustering does assign the data points in different clusters based on their attributes. Clusters are classified based on their similarity—these similarities measurements include connectivity, distance, and intensity [79-80].

## *2.6.11 Reinforcement Learning*

Reinforcement Learning (RL) is training machine learning models to make a series of decisions. Agents learn to reach their goals in an uncertain and complex environment. In reinforcement learning, artificial intelligence faces a game-like situation. Computers use trial and error to solve this problem. Artificial intelligence rewards or punishes the actions it performs to complete the task the programmer wants. The goal is to maximize total rewards. The designer set a reward policy but did not provide any model tips or suggestions on how to solve the game. It's up to the model to decide how to perform the task to maximize rewards, starting with a random trial and completing it with good tactics and superhuman skills [80]. Reinforcement learning is currently the most effective way to suggest machine creativity, leveraging the power of search and many experiments. Unlike humans, running reinforcement learning algorithms on a sufficiently robust computer infrastructure allows artificial intelligence to gather experience from thousands of parallel games. This learning algorithm has been investigated in energy management, medium access control, routing optimization, and topology control [80-82]

## *2.6.12 Principal Components Analysis (PCA):*

PCA is one of the most common unsupervised learning techniques in statistical analysis, data compression, and feature extraction. This statistical technique defines a set of spindles that transform multiple related variables into unrelated variables, the so-called main components. The most relevant spindle holds the maximum data variance. This technique has been studied in many applications, including sensor network measurements, such as extracting features from noise samples and compressing and denoising high-dimensional datasets (time series of measurements). PCA has been used for data reduction and anomaly detection in IoT systems [82-84].

## 2.7. Signal Filtering Techniques for Indoor Localization.

Due to unreliable and inaccurate readings caused by multipath fading, RSSI is often not appropriate in indoor environments. Therefore, it is essential to filter the measured RSSI values to remove the signal's noise to get the accuracy of sound indoor location systems. This section further describes the three filtering techniques applied to the experiment's observed set of RSSI values.

### 2.7.1. Moving Average Filter

The moving average (MA) is the most common and widely used filter in digital signal processing because it is the most accessible digital filter to understand and use. This optimal filter is optimal to avoid random noise while retaining a sharp step response. It obtains N input samples at a time and calculates the average of those points to produce a single output point. In many marks has used moving average filter due to its simplicity and the robustness[34-40]

When the filter's length increases, the output's smoothness increases, and a moving average filter of length N for an input RSSI signal RSSI(MA) may be defined as follows:

$$\text{RSSI(MA)} = \frac{1}{N}\sum_{k=0}^{N-1} RSSI(n-k) \quad for\ n = 0,1,2,3,\dots \quad (28)$$

## 2.7.2. Gaussian Filter

Equation 29 depicts the Gaussian distribution of the RSSI measurement and its probability function. The Gaussian mathematical model chooses the effective value of data sampling as the RSSI value of the wide probability range, and the mean value is then computed as the filter's output. The accuracy of the data can be increased while reducing the influence of high interference and low probability on the entire data set. The RSSI measurement is noise-free due to the 2-D convolution operator known as the Gaussian smoothing operator.

$$G(x) = \frac{1}{\sqrt{2\pi\sigma^2}} e^{-\frac{x^2}{2\sigma^2}} \quad (29)$$

Where;

σ - Standard deviation of the distribution

Bell-shaped Gaussian distribution is assumed to have a mean of 0.

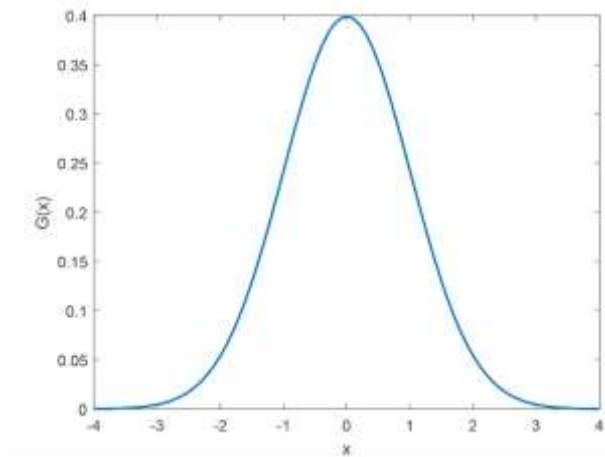

Figure 9: Gaussian distribution with mean 0 and σ = 1

## 2.7.3. Median Filter

In image processing, signal processing, and time-series processing, one-dimensional (1-D) median filters are frequently employed. The ability of the median filter to eliminate the impact of input noise values at gigantic magnitudes is its main benefit over linear filters. (In contrast, linear filters are sensitive to this kind of noise, which means that the output can be significantly damaged even by a small fraction of non-heterogeneous noise values.) The median filter's output may be determined using the Eq. 30.

$$x(n) = median\ [y(n-T)\dots, y(n)\dots, y(n+T)] \quad (30)$$

## 2.8. Related work on ML-based indoor localization

Several related works have been carried out on the localization of sensor nodes in Wireless Sensor Networks (WSN) using Machine Learning and few results for IoT. There are significant differences between WSN and IoT. In the Internet of Things system, all sensors send information directly to the Internet. For example, you can use sensors to monitor the temperature of your body of water. In this case, the data is sent directly to the Internet immediately or regularly, and the server can process the data and interpret it on the front-end interface.

In contrast, WSN does not have a direct connection to the Internet. Instead, various sensors are connected to some router or central node. IoT systems can take advantage of wireless sensor networks by communicating with routers and collecting data. If desired, you can route the data from your router or central node. A summary of related works has been presented in table 2.

*Table 2:Machine Learning algorithms used for IoT applications.*

| Proposed Methodologies for Localization | Machine Learning algorithms used | The complexity of the algorithm | Application |
|---|---|---|---|

| Localization of nodes [85] | Convolutional Neural Networks | Moderate | IoT |
| --- | --- | --- | --- |
| Human Localization [86] | Neural Networks | Moderate | WSN |
| Localization based on NNs[87] | Neural Networks | High | WSN |
| Localization using SVM [88] | Support Vector Machine | Moderate | WSN |
| Underwater Surveillance System [89] | Decision Tree-Based Localization | Low | WSN |
| Distributed Localization [90] | Self-Organizing Map | Moderate | WSN |
| Soft Localization [91] | Neural Networks | Moderate | WSN |

## 2.9. Indoor Localization Applications

### 2.9.1. Location-based on Services in Indoor Environment

Location-Based Services (LBS), which utilize a mobile device's location to provide contextually relevant information, may have significant economic advantages. Indoors, such services are becoming more popular. Obtaining safety information or current information about nearby theaters, concerts, or events are examples of indoor LBS. These apps may also help you find the appropriate shop in a mall or the right office in a public building. The location detection of goods inside a shop or warehouse is essential to both the owner and the consumers. Location-based advertising, billing, and local search services, in particular, have a significant economic value. There is a desire to direct people to the proper exposition booths at significant tradeshows and exhibits. Using localization for resource monitoring, fleet management, and user statistics adds value to the positioning provider.

### 2.9.2. Smart Home Applications

Ambient Assistant Living (AAL) devices help older persons in their homes with their everyday tasks. At home, applications include identifying misplaced objects, physical gesture games, and location-based services. A significant feature of AAL systems is location awareness, which necessitates indoor positioning. Medical monitoring, such as vital signs, emergency, and fall detection, are examples of residential applications, as are service and customized entertainment systems, such as intelligent audio systems. The precision required here may be very high, up to 1 meter [92-93].

### 2.9.3. Context Detection and Situational Awareness

Mobile devices provide a wide range of helpful capabilities, and an automatic adaption of the mobile device in response to a change in the user's environment is desired. Such a feature saves users time and effort by assisting them in certain circumstances. The mobile device itself must identify the mobile user's context in order for such automated adaption to be possible. The present geographical location is the most critical factor in determining the user's context. An intelligent conference guide, for example, may offer information on the subject being addressed in adjacent auditoriums. In this application, systems with the precision of a few meters perform well [94-95].

### 2.9.4. Health Care Applications

Medical personnel's location monitoring in emergency circumstances has become more essential in hospitals. In addition to patient and equipment monitoring, medical uses in hospitals include fall detection for patients. During operations, precise positioning is needed for robotic help. High precision is required for positioning systems used in hospitals [95-96]. Analytical instruments that are now in use may be replaced with more efficient surgical equipment.

### 2.9.5. Police and Firefighters

Indoor positioning systems may be very useful in law enforcement, rescue, and fire services, for example, in detecting the location of firefighters in a burning structure. Indoor positioning allows the police to profit from several valuable applications, including the immediate identification of theft or burglary, the detection of the location of police dogs trained to locate explosives in a building, and the finding and recovery of stolen goods. The precision requirements range from a few meters to a few meters [97-98].

### 2.9.6. Environmental Monitoring

Through environmental monitoring, it is simple to detect certain phenomena such as heat, pressure, humidity, air pollution, and deformation of objects and buildings. Multiple sensor nodes are arranged as a Wireless Sensor Network to monitor these metrics across a certain interior or outdoor area (WSN). A wireless sensor network (WSN) comprises tiny, low-cost, geographically dispersed autonomous nodes with minimal processing and computing capabilities and wireless communication radios. As a result, localization techniques are used to determine the location of nodes. Accuracy requirements range from a few meters to a few kilometers [99]. A thorough overview of the literature on WSNs may be found here.

### 2.9.7. Structural Health Monitoring

Performing strain measurements using advanced sensors integrated into steel reinforcements inside concrete may assist in identifying strain variations and deformation induced by loads at different locations. In this instance, centimeter precision is needed [100].

***Indoor positioning is critical for all future IoT applications.***

# CHAPTER 3:

## 3. Experimental Testbeds and Datasets

### 3.1. Wi-Fi Based Experimental Testbed Design and Implementation

We created an experimental setup for gathering RSSI data in a 293.8cmx274.6cm indoor setting. Three beacon nodes are physically installed in known fixed locations in our system and interact with the mobile node. The portable mobile node collects RSSI data from the three beacon nodes via Wi-Fi and sends it to the cloud server. ESP 8266 is used to create the beacon and mobile nodes. The IEEE 802.11 standard is included in the ESP-8266. It is low-cost and supports Internet Protocol version 4 (IPv4), Transmission Control Protocol (TCP), User Datagram Protocol (UDP), and Hypertext Transfer Protocol (HTTP) [101], [102]. The beacon nodes communicate with the mobile node using Wi-Fi, which uses the IEEE 802.11 standard in the 2.4 GHz range. The research's hardware design consists primarily of constructing a mobile node, a beacon node, and an IoT cloud architecture to transmit RSSI data to a distant server. The suggested system may also be utilized to analyze RSSI data for smart homes, healthcare, senior monitoring, and other indoor positioning applications.

## *Hardware Setup*

The system utilizes two distinct ESP 8266 modules, ESP 12E and ESP 01 as shown in Fig.10 and Fig.11 respectively . The software and hardware designs are identical in both modules. However, the number of GPIOs differs. ESP 12E is utilized for the mobile node, while ESP 01 is used for the beacon nodes. Figure 5 depicts the appearance of the two ESP8266 modules.

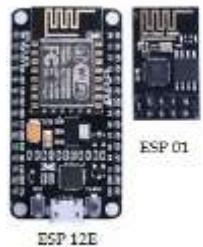

*Figure 10:ESP 12 E and ESP 01 modules*

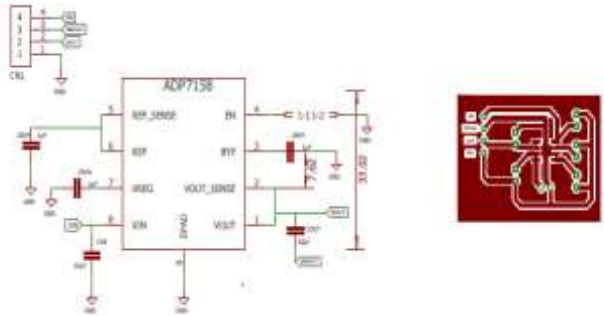

*Figure 11:Schematic diagram and PCB of regulator circuit for ESP 01*

ESP 12E uses a Li-Polymer rechargeable battery storage. The ESP-01 is the beacon node, using a separate 3.3 V DC supply with an ADP7158 linear voltage regulator. Figure 6 shows the schematic and PCB of the voltage regulator for the beacon node, which is powered by the RAC20-05SK/W series PCB-mount power conversion module. The module operates using 230V AC input, and the output is 5V, which can be directly used as the input to the regulator. The basic developed architecture is shown in figure 12(a). Figure 12(b) shows the arrangement of beacon nodes in star-connected WSN in the actual environment.

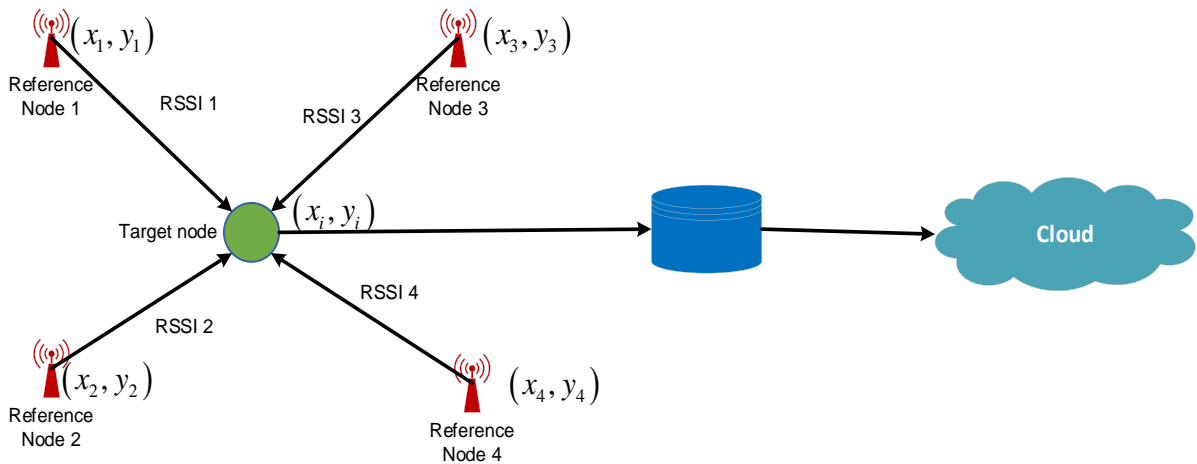

*Figure 12(a): Basic System Architecture*

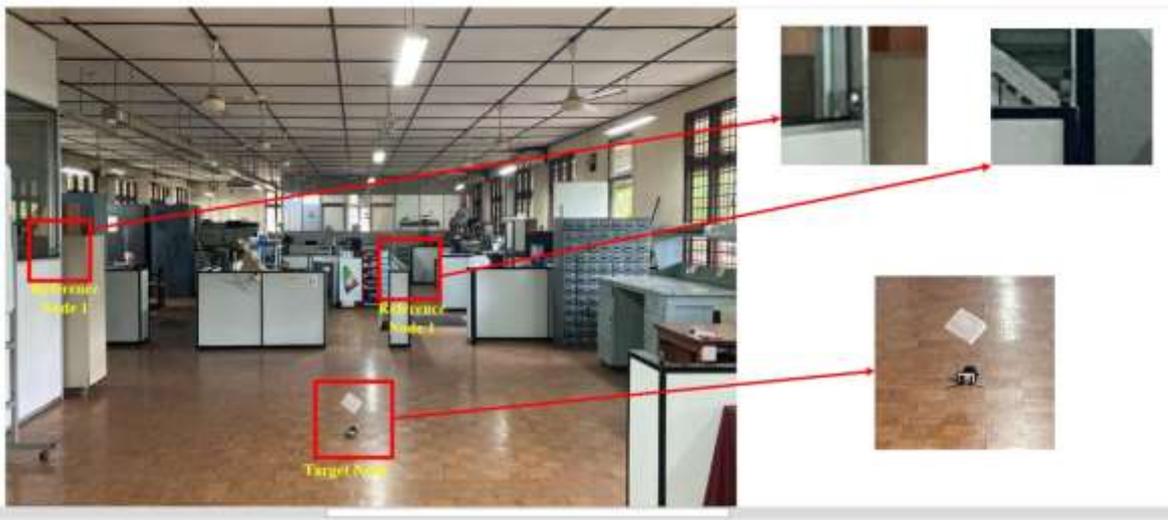

F*igure 13(b): Actual Experiment Area*

Using the trilateration method, this study may be used for interior positioning systems in the future. Three beacon nodes are put at known coordinates using the triangulation method. The private Wi-Fi network's three beacons and mobile nodes are installed in an interior setting known as a Building Area Network (BAN). The mobile node publishes the RSSI data to the MQTT broker in the Wide Area Network (WAN) via the public ADSL 2+ network. The data is published to a cloud storage server by the MQTT broker for further processing. A traditional broadband router with internet access is utilized as an interface between the BAN and the WAN.

## *IoT Cloud Architecture*

The IoT cloud architecture for data collecting and publication to a cloud storage server is shown in Fig.12. The RSSI data gathered by the mobile node over a private Wi-Fi network is made public through the internet. The public network used in the study is ADSL 2+ broadband technology. On the connection between hardware platforms, the IoT cloud platform is utilized. The Internet of the Things cloud platform is a distributed mosquito MQTT broker that sends data to a distant server [103]. Wi-Fi and internet technologies transmit gathered data between the hardware platform and the distant server. A test configuration with three beacon nodes and a mobile node was built to investigate the feasibility of seeing RSSI data in real time.

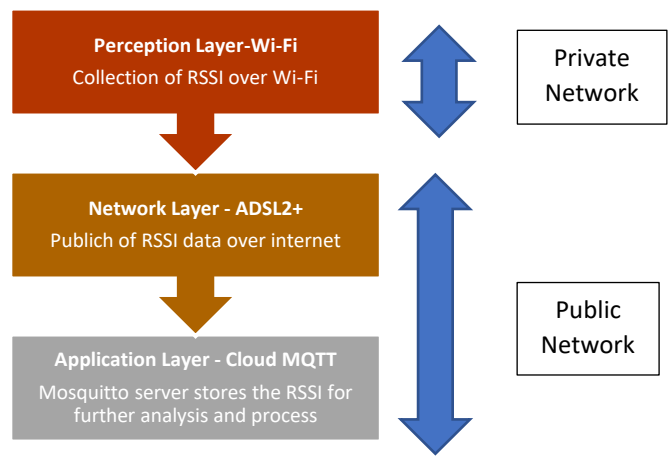

*Figure 14: The IOT Cloud Architecture for Data Collecting & Publishing to a cloud storage server*

THE MOBILE NODE PUBLISHES the RSSI data to CloudMQTT over the internet. RSSI is collected using the portable mobile node with the three beacon nodes. The beacon nodes are set up as soft access points (SoftAP), and the portable mobile node is set up as a client using CloudMQTT's application-specific interface (API). In Wi-Fi active mode, the mobile client node sends a probe request to all devices in the BAN and waits for a response from the beacon nodes. The beacon node sends the response probe as the access point, and the mobile node may use it to determine the received signal strength in dBm. CloudMQTT's Websocket UI displays the RSSI values for the three beacon nodes in real-time.

## 3.2. BLE, Zigbee and LoRaWAN Testbeds

This work used the data set by Sebestian and Petros in [104]. The original experiment was conducted in two different environments, and two datasets were available. However, this experiment used the dataset related to environment 1 [104]. The experiment setup has been implemented in a laboratory room, as shown in Fig.14. The environment is non-lone-of-sight (NLOS). In the evening, an experiment was conducted to eliminate interferences from other wireless devices such as Wi-Fi hotspots and mobile phones. Beacon nodes are placed at positions A, B, and C, as shown in Fig.15 , and mobile nodes are placed at positions D1, D2, and D3, respectively, to collect RSSI data. A series of tests were conducted to test positioning accuracy when positioning short and long distances between receivers and transmitters in all indoor systems.

When most people leave, all experiments are done at night to minimize interference caused by other devices using the same media for transmission. Because RSSI values are vulnerable to interference, a controlled environment can generate more consistent readings for all tests performed.

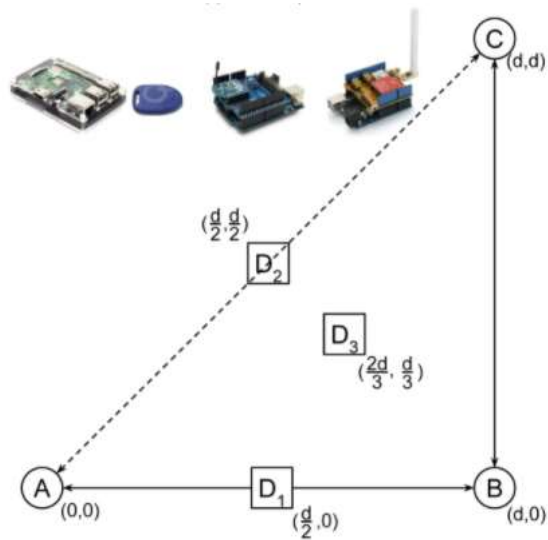

*Figure 15:Positioning of Beacon Nodes*

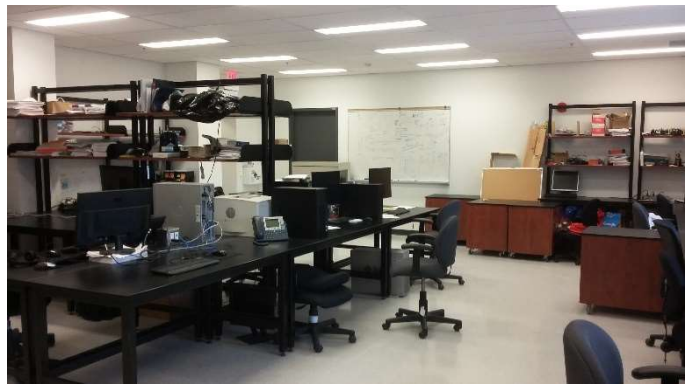

*Figure 16:The Experiment Setup*

## LoRAWAN Testbed

The dataset used in this experiment was taken from [105]. The RSSI readings of an array of 13 ibeacons on the first floor of Western Michigan University's Waldo Library were used to produce the dataset (see figure 16). The information was gathered using an iPhone 6S. A labeled dataset (1420 instances) and an unlabeled dataset (1420 instances) are included in the dataset (5191 cases). The recording was completed within the library's regular operating hours. More significant RSSI values indicate closer proximity to a given iBeacon.For out-of-range iBeacons, the RSSI is indicated by -200. Fig.16 depicts the layout of the arrangement of iBeacons in green color circles. This experiment has only used the labeled data. Further, a location labeled in the original data set includes the number of rows and columns. However, as per the zone classification approach, we have re-labeled the ibecon dataset by dividing it into four zones: A, B, C, and D.

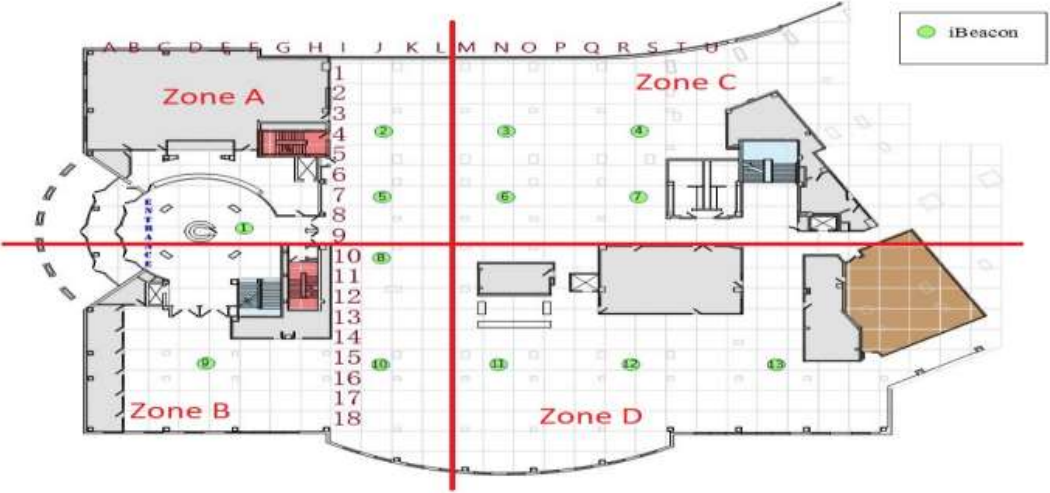

*Figure 17: Layout of the arrangement of iBeacons*

# CHAPTER 4:

## 4. Supervised Regressor for Indoor Localization

### 4.1. Introduction

In many indoor IoT applications, the localization approach must be used to determine the precise location of a sensor node. For instance, it uses wireless sensor data to determine the location of a moving human, vehicle, or animal. The broadcasting signals from the sensor node can estimate its location in these localization approaches, negating the need for extra gear to determine the location. Moreover, it is significant as a sensor node's position affects the performance and accuracy of the information. On the other hand, the localization method could help estimate a sensor node's accurate position based on other neighboring nodes' values. The IoT systems use many wireless technologies to communicate within the sensor nodes in their networks, such as Bluetooth, infrared, LoRaWAN, Zigbee, Wi-Fi, GPRS, and 3G. With these technologies, it can get the geographical location information of a node in different ways.

Location-based services are the primary service of the IoT. Therefore, localization accuracy is an important issue. Many localization algorithms were developed for Wireless Sensor Networks (WSNs) applications and IoT. Most of the algorithms proposed in the literature for indoor localization are statistical-based [13-17]. Different hardware devices are utilized by most of the existing statistical localization solutions, which increase the cost and significantly limit location-based applications.

Although there is a lot of related research on ML-based indoor localization for IoT systems, many times proposed approaches are assessed using just one ML algorithm or without the implementation of an experimental test-bed. Additionally, it may be challenging to install algorithms created for sensor node localisation on actual IoT devices and they may be ineffective. In this study, an indoor IoT testbed was created, deployed, and RSSI data was gathered. In the second phase, data were pre-processed, and supervised ML algorithms were investigated on estimating the accurate location of a sensor node and evaluating each proposed algorithm's performance.

The position estimation is a compulsory component of many IoT applications [5], especially in the tracking and monitoring applications such as [6-8]: animal behaviors monitoring, air quality monitoring, and smart cities.Besides, the location information enables various emerging applications such as inventory management, intrusion detection, road traffic tracking, health monitoring, etc. [9-10]. Methods for determining position are usually based on geometric calculations such as trilateration (by measuring the angle to a fixed point or node with a known position) or trilateration (by measuring the distance between nodes) [11]. Several techniques can be employed to determine the distance between two nodes in an IoT network. For example, synchronization, RSSI, and the physical characteristics of the carrying wave [12]. The localization techniques in the IoT systems can be free of a previous position determination in the network, relying on a few specific sensors' position information and their inter measurements in the network, such as time difference of arrival, distance, angle of arrival, and connectivity [13]. The relationship between RSSI and distance is essential for wireless-based indoor localization systems. The most common method is the location of fixed nodes based on triangulation [15]. According to [16-18] and [19], RSSI-based indoor positioning uses various algorithms to identify mobile users in an indoor environment.

These positioning algorithms are primarily trilateral measurements, arrival angle (AOA), arrival time (TOA), and arrival time difference (TDOA) [ 20] is based. Due to its simplicity and wide range of applications, the above algorithms' most popular is the trilateration algorithm.

Significant research works investigated ML for indoor localization problems in WSN and IoT [21-25]. Though significant related work exists on localization for WSNs, most of the related work on Machine Learning-based indoor localization is for WSNs, not IoT systems. The majority of studies on deterministic localization algorithms factor in noise and various conditions; using an ML-based approach, these factors and noise can be considered a part of the general environment in which the localization must be carried out. Existing works on ML-based localization for IoT are limited to only one type of specific ML algorithm. Moreover, a lack of performance evaluation on different algorithms has been conducted.

The RSSI-based localization of the target node is estimated by using multiple reference nodes. Let the target node is denoted as $(x_b, y_b)$ with the fixed reference node locations at $(x_i, y_i), i = 1, 2, ...., M$. i.e., M ≥ 3. The target node's RSSI measurement is included with noise due to signal fluctuation. The noisy reference location at the target node is represented as $(x_i, y_i)$ and the subsequent RSSI estimation is represented as $p_i$. An additive independent with zero-mean Gaussian noise affects the anchor node location information with a standard deviation indicated as $\sigma_{a_i}$ [36]. There is variation of $\sigma_{a_i}$ values due to the multiples reference nodes. On the other hand, it considers the identical for both the x and y coordinates of a targeted node.

$$x_i = \overline{x_i} + n_{x_i}$$

$$y_i = \overline{y_i} + n_{y_i}$$

$$n_{x_i}, n_{y_i} \sim N(0, \sigma_{a_i}^2)$$

Similarly, the RSSI measurement by log-normal shadowing system model of radio signal path-loss is also employed. So that the target node of the transmitted signal from the $i$th reference nodes is represented as $p_i$ (dBm) [26,37]. The perturbation $n_{\sigma_{p_i}}$ in $p_i$ is denotes an additive noises with independent zero-mean Gaussian and standard deviation is denoted as $\sigma_{p_i}$ (dB), such that.

$$p_i = \overline{p_i} + n_{\sigma_{p_i}}$$

$$n_{\sigma_{p_i}} \sim N(0, \sigma_{p_i}^2)$$

Moreover, the shadowing pathloss system model represents the correlation amongst the ith mean of the power and the distance amongst the target source and the ith reference nodes, i.e.,

$$d_i = \sqrt{(x_i - x_b)^2 + (y_i - y_b)^2}$$

as

$$p_i = p_0 - 10\eta \log_{10} \frac{d_i}{d_0}$$

here $d_0$ defines the reference nodes distance, $p_0$ defines received source power value at the reference distances, and $\eta$ is the pathloss exponent value, respectively. Assumed the perturbed value $p_i$, the RSSI-caused measure of the distance amongst the target source and the ith reference nodes is represented by $d_i$, and it is computed as

$$d_i = d_0 10^{\frac{p_0 - p_i}{10\eta}}$$

This study considers the challenges of computational efficiency and energy resources constraints for location estimation of the target node by using the reference nodes. In this manner, the RSSI location measurement from every reference node is accessible to the target node at any period for localization. To cope with the above-mentioned challenges, this study proposed a pseudo-linear solution (PLS) to solve the autonomous-localization issue is described as below:

The basic idea of the proposed algorithm is to find the near optimal position of the target node that decreases the sum of the squared error values. As denoted earlier, the reference nodes position $(x_i, y_i)$ and its subsequent distances $d_i$, $i = 1, 2, ....., M$, the target node location is computed by intersecting the circles described as

$$(x - x_i)^2 - (y - y_i)^2 = d_i^2, \quad i = 1, 2, ....., M.$$

To cope with the system non-linearization nature of equations (4), subtraction of the equation regarding from the i = 1 to the other outcomes in a system of linearization equations defined as

$$2As = b$$

here

$$A = \begin{bmatrix} x_1 - x_c & y_1 - y_c \\ x_2 - x_c & y_1 - y_c \\ \vdots & \vdots \\ x_i - x_c & y_1 - y_c \end{bmatrix}, b = \begin{bmatrix} d_c - d_1^2 + k_1 - k_c \\ d_c - d_2^2 + k_2 - k_c \\ \vdots \\ d_c - d_M^2 + k_M - k_c \end{bmatrix}, \ s = \begin{bmatrix} x \\ y \end{bmatrix}, \ k_i = x_i^2 + y_i^2$$

$$x_c = \frac{1}{M}\sum_{i=1}^{M} x_i, y_c = \frac{1}{M}\sum_{i=1}^{M} y_i, d_c = \frac{1}{M}\sum_{i=1}^{M} d_i^2, \text{ and } k_c = \frac{1}{M}\sum_{i=1}^{M} k_i,$$

It is observed that Eq. (5) is an over-determined set of non-linear equations, thus the objective is to find a solution **s** by decreasing the subsequent sum of the square-error function

$$J(s) = \arg\min_{s} \left[ \|b - As\|_2^2 \right]$$

The solution of (6) is

$$s = \frac{1}{2}(A^T A)^{-1} A^T b.$$

It is noted that, only noisy information $x_i$, $y_i$, and $d_i$ are accessible rather than actual $x_i$, $y_i$, and $d_i$. To factor in the change of the scale as well as numerical attribute values that included with multiple reference node's location and distance estimations of equation (6), the minimization of the sum of square errors as

$$J(s) = \arg\min_{s} \left[ \left\| W^{\frac{1}{2}}(b - As) \right\|_2^2 \right]$$

where

$$A = \begin{bmatrix} x_1 - x_c & y_1 - y_c \\ x_2 - x_c & y_1 - y_c \\ \vdots & \vdots \\ x_i - x_c & y_1 - y_c \end{bmatrix}, b = \begin{bmatrix} d_c - d_1^2 + k_1 - k_c \\ d_c - d_2^2 + k_2 - k_c \\ \vdots \\ d_c - d_M^2 + k_M - k_c \end{bmatrix}, \ \hat{s} = \begin{bmatrix} x \\ y \end{bmatrix}, \ k_i = x_i^2 + y_i^2$$

$$x_c = \frac{1}{M}\sum_{i=1}^{M} x_i, y_c = \frac{1}{M}\sum_{i=1}^{M} y_i, d_c = \frac{1}{M}\sum_{i=1}^{M} d_i^2, \text{ and } k_c = \frac{1}{M}\sum_{i=1}^{M} k_i,$$

and W denoted as $M \times M$ weighted matrix. Then, the explanation $\hat{s}$ of (8) is

$$\hat{s} = \frac{1}{2}\left(A^T W^{-1} A\right) A^T W^{-1} \tilde{b}$$

To evaluate the weight matrix (W), it is noted that the error vector $\tilde{b} - As$ in (8) contains two noise elements, one is in reference node's location and another one is in distance measurement. The vector $\tilde{b}$ comprises the squares of the noise elements, which basically lead the impact of noise in $A$ to the error vector covariance. Thus, it is considered that the W represents the covariance matrix of $\tilde{b}$. Thus, $\tilde{b}$ is simplified as

$$\tilde{b} = \left(I - \frac{1}{M} 11^T\right) b_1$$

Where

$$b_1 = \left[k_1 - d_1^2, k_2 - d_2^2, \ldots, k_M - d_M^2\right]$$

Hence, we have

$$W = \left(I - \frac{1}{M} 11^T\right) \text{cov}(b_1) \left(I - \frac{1}{M} 11^T\right)$$

where

$$Cov(b_1) = dig\left(Var\left(k_1 - d_1^2\right), Var\left(k_2 - d_2^2\right), \ldots, Var\left(k_M - d_M^2\right)\right)$$

Reflecting the assumptions mentioned above is independent features of the noises of the reference node's location and RSSI-induced distances, () is defined as

$$Var\left(k_i - d_i^2\right) = Var(k_i) + Var(d_i^2)$$

It is notable that the $k_i$ represent the summation of the square with independent normal distributed random variable $x_i$, and $y_i$ as well as a non-zero mean. Thus, variance $\dfrac{k_i}{\sigma_{a_i}^2}$ is defined as

$$Var\left(\frac{k_i}{\sigma_{a_i}^2}\right) = 2\left(2 + 2\frac{x_i^2 + y_i^2}{\sigma_{a_i}^2}\right)$$

And consequently

$$Var(k_i) = 4\sigma_{a_i}^2\left(\sigma_{a_i}^2 + \left(x_i^2 + y_i^2\right)\right).$$

Thus $Var\left(d_i^2\right)$ is computed as [9]

$$Var\left(d_i^2\right) = \exp\left(4\mu_{d_i}\right)\left[\exp\left(8\sigma_{a_i}^2\right) - \exp\left(4\sigma_{a_i}^2\right)\right]$$

where

$$\mu_{d_i} = \ln d_i \text{ and } \sigma_{p_i} = \frac{\ln 10}{10\eta}\sigma_{p_i}$$

The noisy values of $x_i$, $y_i$, and $d_i$ are used to compute equations (11) and (12) because of the actual values $x_i$, $y_i$, and $d_i$ are not accessible.

Moreover, it is noted that the equation (9) has multiple sources of bias. The matrix $A$ contains noise, the errors in $\tilde{b}$ are not additive as well as zero-mean, and there is a relationship among the errors in $A$ and $\tilde{b}$. To evaluate the bias into the system model algorithm taking an additive error, Equation (9) is simplified as

$$\tilde{A} = A + N, \tilde{b} \approx b + e$$

By using equations (13) and (9), the $\mathrm{E}\left[\hat{s}\right]$ is written as

$$\mathrm{E}\left[\hat{s}\right] = \underbrace{\mathrm{E}\left[\left(A^T W^{-1} A\right)^{-1}\right] A^T W^{-1} b}_{I} + \underbrace{\mathrm{E}\left[\left(A^T W^{-1} A\right)^{-1} N^T\right] W^{-1} b}_{II}$$
$$+ \underbrace{\mathrm{E}\left[\left(A^T W^{-1} A\right)^{-1} N^T W^{-1} e\right]}_{III} + \underbrace{\mathrm{E}\left[\left(A^T W^{-1} A\right)^{-1} A^T W^{-1} e\right]}_{IV}$$

In equation (14), the expansion of $A^T W^{-1} A$ to $(A+N)^T W^{-1}(A+N)$, to make the equation simpler has been avoided. It is assumed that part I in equation (14) is the correspond to the target node location $\hat{s}$ and the remaining of the parts, II, III, and IV are the bias parts owing to estimation errors.

Part II provides the bias owing to the noise in $A$. Part III provides the statistical dependence among $\tilde{A}$ and $\tilde{b}$ i.e., $E\left[N^T e\right] \neq 0$. Moreover, part IV provides the non-additive nature of perturbation in $d_i$ i.e., $E[e] \neq 0$. To compensate of the bias parts II, III, and IV, the expectation for concerning noise covariance is then subtraction in equation (9) is written as

$$\hat{s}_{bc} = \frac{1}{2}\left(A^T W^{-1} A - \mathrm{E}\left[N^T W^{-1} N\right]\right)^{-1} \times \left\{A^T W^{-1}\left(\tilde{b} - \mathrm{E}\left[\tilde{b}\right]\right) - \mathrm{E}\left[N^T W^{-1} \tilde{b}\right]\right\}$$

To compute $\mathrm{E}\left[N^T W^{-1} N\right]$ and $\mathrm{E}\left[N^T W^{-1} \tilde{b}\right]$, N can be written as

$$N = N_1 - N_2$$

where

$$N_1 = \begin{bmatrix} n_{x_1} & n_{y_1} \\ n_{x_2} & n_{y_2} \\ \dots & \dots \\ n_{x_M} & n_{y_M} \end{bmatrix}, N_2 = \begin{bmatrix} n_{x_c} & n_{y_c} \\ n_{x_c} & n_{y_c} \\ \dots & \dots \\ n_{x_c} & n_{y_c} \end{bmatrix}$$

$$n_{x_c} = \frac{1}{M} \sum_{i=1}^{M} n_{x_i}, \text{ and } n_{y_c} = \frac{1}{M} \sum_{i=1}^{M} n_{y_i}$$

Thus. We have

$$L = E[N^T W^{-1} N]$$
$$= E[N_1^T W^{-1} N_1] + E[N_2^T W^{-1} N_2] - E[N_2^T W^{-1} N_1] - E[N_1^T W^{-1} N_2]$$

Representing (i, j)th is the element of $W^{-1}$ by $w'_{ij}$, and the entries of (17) are estimated as

$$E[N_1^T W^{-1} N_1] = diag\left\{ \sum_{i=1}^{M} w'_{ii} \sigma^2_{n_{x_i}}, \sum_{i=1}^{M} w'_{ii} \sigma^2_{n_{y_i}}, \right\},$$

$$E[N_2^T W^{-1} N_2] = diag\left\{ \frac{1}{M^2} \sum_{i=1}^{M} w'_{ii} \sigma^2_{n_{x_i}}, \frac{1}{M^2} \sum_{i=1}^{M} w'_{ii} \sigma^2_{n_{y_i}}, \right\},$$

$$E[N_2^T W^{-1} N_1] = diag\left\{ \frac{1}{M} \sum_{i=1}^{M} \sigma^2_{n_{x_i}} \left( \sum_{j=1}^{M} w'_{ji} \right), \frac{1}{M} \sum_{i=1}^{M} \sigma^2_{n_{y_i}} \left( \sum_{j=1}^{M} w'_{ji} \right) \right\},$$

and

$$E[N_1^T W^{-1} N_2] = diag\left\{ \frac{1}{M} \sum_{i=1}^{M} \sigma^2_{n_{x_i}} \left( \sum_{j=1}^{M} w'_{ij} \right), \frac{1}{M} \sum_{i=1}^{M} \sigma^2_{n_{y_i}} \left( \sum_{j=1}^{M} w'_{ij} \right) \right\},$$

The bias owing to the dependence of noises in the $A$ and $\tilde{b}$ can be written as

$$E[N^T W^{-1} \tilde{b}] = E[N_1^T W^{-1} \tilde{b}] - E[N_2^T W^{-1} \tilde{b}]$$

where

$$\mathrm{E}\left[N_1^T W^{-1}\tilde{b}\right] = \begin{bmatrix} \dfrac{2}{M}\sum_{i=1}^{M} x_i \sigma_{n_{x_i}}^2 \left(\sum_{j=1}^{M} w'_{ji}\right) \\ \dfrac{2}{M}\sum_{i=1}^{M} y_i \sigma_{n_{y_i}}^2 \left(\sum_{j=1}^{M} w'_{ji}\right) \end{bmatrix}$$

and

$$\mathrm{E}\left[N_2^T W^{-1}\tilde{b}\right] = -\begin{bmatrix} \dfrac{2}{M^2}\sum_{i=1}^{M} x_1 \sigma_{n_{x_i}}^2 \mathbf{1}^T W^{-1}\mathbf{1} \\ \dfrac{2}{M^2}\sum_{i=1}^{M} y_1 \sigma_{n_{y_i}}^2 \mathbf{1}^T W^{-1}\mathbf{1} \end{bmatrix}$$

To compensate of the bias provided by the non-additive feature of the perturbation in the $d_i$ [part IV in equation (14)], $\mathrm{E}\left[\tilde{b}\right]$ with its i-*th* entry can be computed as

$$\mathrm{E}\left[\tilde{b}_i\right] = \mathrm{E}\left[\tilde{b}_c - \tilde{d}_i^2 + k_i - k_c\right].$$

It can be considered that the noise is independent of the reference 's location and RSSI-induced distances; thus equation (19) is expressed as

$$\mathrm{E}\left[\tilde{b}_i\right] = \mathrm{E}\left[d_c\right] - \mathrm{E}\left[\tilde{d}_i^2\right] + \mathrm{E}\left[k_i\right] + \mathrm{E}\left[k_c\right].$$

To compute $\mathrm{E}\left[\tilde{d}_i^2\right]$, it is noted that the $\tilde{d}_i^2$ employing in equation (3) is equal to

$$\tilde{d}_i^2 = d_i^2 \exp\left(\sqrt{2}un_{p_i}\right)$$

where

$$u = \frac{\ln 10}{5\sqrt{2}\eta}$$

Therefore,

$$E\left[d_i^2\right] = d_i^2 E\left[\exp\left(\sqrt{2}un_{p_i}\right)\right]$$
$$= d_i^2 \exp\left(u^2 \sigma_{n_{p_i}}^2\right)$$

It is noted that the value of $u^2 \sigma_{n_{p_i}}^2$ always small even though values of $\sigma_{n_{p_i}}$ is high. In this manner, by employing the second-order expansion of the Taylor-series for the function $\exp\left(u^2 \sigma_{n_{p_i}}^2\right)$ near to zero, (21) is estimated as

$$E\left[d_i^2\right] = d_i^2 + d_i^2 \left( u^2 \sigma_{n_{p_i}}^2 + \frac{u^4 \sigma_{n_{p_i}}^4}{2} \right)$$

By considering the assumption, $E\left[d_c^2\right]$ is correspond to

$$E\left[d_c\right] = \frac{1}{M}\sum_{i=1}^{M} E\left[d_i^2\right]$$

The term $E\left[k_i\right]$ in (20) corresponds to

$$E\left[k_i\right] = x_i^2 + y_i^2 + 2\sigma_{n_{p_i}}^2$$

And mentioned assumption $E\left[k_c\right]$ develop into

$$E\left[k_c\right] = \frac{1}{M}\sum_{i=1}^{M} E\left[k_i\right]$$

Employing (22) and (23) $E\left[\tilde{b}\right]$ is expressed as

$$E\left[\tilde{b}\right] = b + t$$

here the ith entry for the t is

$$t_i = \left( u^2 \sigma_{n_{p_i}}^2 + \frac{u^4 \sigma_{n_{p_i}}^4}{2} \right)\left( d_c^2 - d_i^2 \right) + 2\left( \sigma_{n_{a_i}}^2 - \frac{1}{M} \sum \sigma_{n_{a_i}}^2 \right)$$

It is noted that the $d_i$ is not available, thus the subsequent noise measurement values are employed in the estimation of the t.

Computation estimation shows that evaluation of the bias owing to the included of the noise in the $A$ and $\tilde{b}$ employing (18) is approximate actual value only when low noise exists in reference node's location. Thus, it is dependent of bias on the $(x_i, y_i)$ and becomes the poor evaluation performance is provided with higher values of the $\sigma_{a_i}$. The target node estimated location, that is bias compensated in the presented PLS algorithm, the bias-compensated solution $\hat{s}_{bc}$ in (15), is computed as a closed form equation as:

$$\hat{s}_{bc} = \frac{1}{2}\left( A^T W^{-1} A - L \right)^{-1} A^T W^{-1} \left( \tilde{b} - t \right)$$

## 4.2. Experiments on Wi-Fi Based Indoor Environments

We examined supervised ML techniques to estimate the precise location of the sensor node. Typically, these algorithms are implemented in two parts. Data is acquired and sent to the algorithm in the first training phase so that it can learn patterns and build a model to categorize or forecast data attributes. New data is tested against the model that was created during the training phase in the second testing phase, and the model's efficacy is revealed. Such two-phase learning algorithms are called supervised learning algorithms. In this work, we investigate Linear Regression (LR), Polynomial Regression (PR), Support ector Machine (SVM), Decision Tree

Regression (DTR), and Random Forest Regression (RFR). All the algorithms were coded using Python 3 on Jupiter Notebook.

### 4.2.1. Data Pre-processing

It has collected 4520 RSSI values from 32 known locations in the test-bed in one test. The dataset was filtered using the three best-outperformed filters available in the literature: moving average, median, and gaussian. The response from each filter is shown in figures 17,18 and 19. Then, the data set was trained using supervised ML algorithms Linear Regression (LR), Polynomial Regression (PR), Support Vector Machine (SVM), Decision Tree Regression (DTR), and Random Forest Regression (RFR). Further error histograms of each algorithm were plotted. Initially, the experiment started with three reference nodes, and step by step, the number of references nodes increased up to five. For each step, Root Mean Squared Error (RMSE) and Coefficient of Determination(R2) values were calculated to compare the performances. The Root Mean Squared Error (RMSE) and coefficient of determination, R2, were compared.

Root Mean Square Error:

Root Mean Squared Error is the square root of Mean Squared error, the average squared difference between the original and predicted values in the data set. It is defined as

$$RMSE = \sqrt{\frac{1}{N}\sum_{n=1}^{N}\|p_n - \hat{p}_n\|} \quad (30)$$

Where $p_n = [x_n, y_n, z_n]^T$ is the true position of the UAVs, and $\hat{p}_n$ is the estimated position?

## Mean Absolute error (MAE)

The Mean absolute error represents the average of the absolute difference between the actual and predicted values in the dataset. It measures the average of the residuals in the dataset and is calculated as:

$$MAE = \frac{1}{N} \sum_{n=1}^{N} |p_n - \hat{p}_n| \quad (31)$$

Where N is the total time interval for one localization procedure, the MAEs along different axes $MAE_x$, and $MAE_y$, Have been calculated by setting $p_n$ as $p_x$, $p_y$ respectively.

## Standard Deviation Error:

Standard deviation (STD) Error is calculated by dividing the standard deviation of the sample by the square root of the sample size. STD of the estimation errors can be calculated as

$$STD = \sqrt{\frac{1}{N} \sum_{n=1}^{N} \left(p_n - \bar{\hat{p}}_n\right)^2} \quad (32)$$

Where $\bar{\hat{p}}$ is the mean value of the estimation error $p_n$. A larger STD indicates that the means are more spread out; thus, it is more likely that the sample mean is an inaccurate representation of the true mean. On the other hand, a smaller STD error indicates that the means are closer together, and thus it is more likely that the sample mean is an accurate representation of the true position mean of the target node.

## Coefficient of determination

The coefficient of determination or $R^2$ method is the proportion of the variance in the dependent variable predicted by the independent variable. It indicates the level of variation in the given data set and is calculated as

$$R^2 = 1 - \frac{\sum_{n=1}^{N}(p_n - \hat{p}_n)^2}{\sum_{n=1}^{N}(p_n - \overline{\hat{p}}_n)^2} \quad (33)$$

The coefficient of determination is the square of the correlation (r), which tests for models' fitness using values between 0 and 1. Values nearer to 1 depict a mutual relationship, while values closer to 0 indicate a weaker association. The figure 16 shows the RSSI row data set extract from the cloud. And figure 17-19 show the filtered data from the moving average filter, median filter, and Gaussian filter respectively.

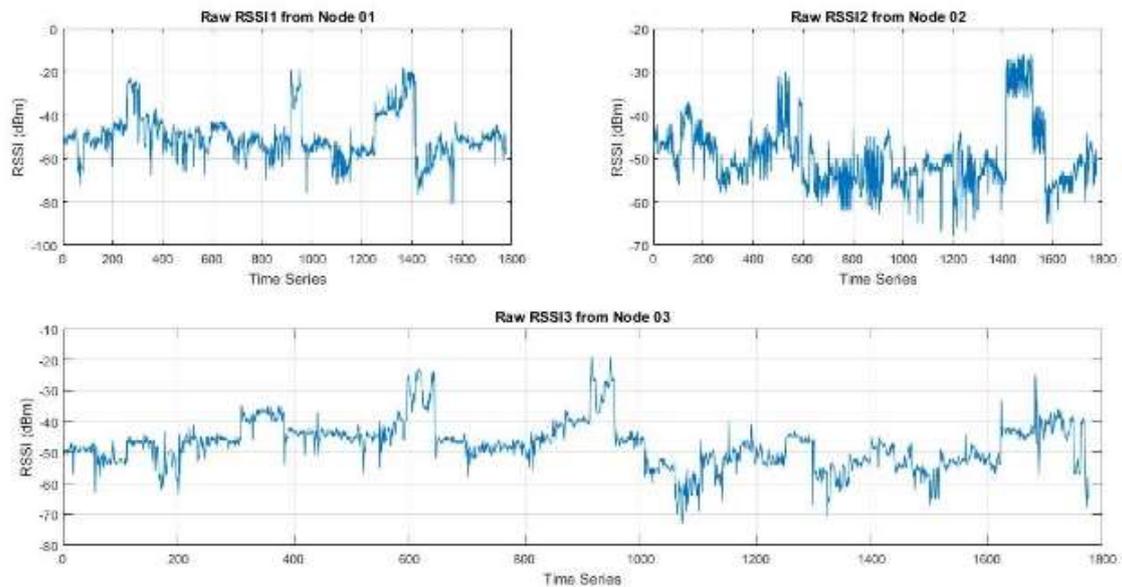

*Figure 18:RSSI data without filter*

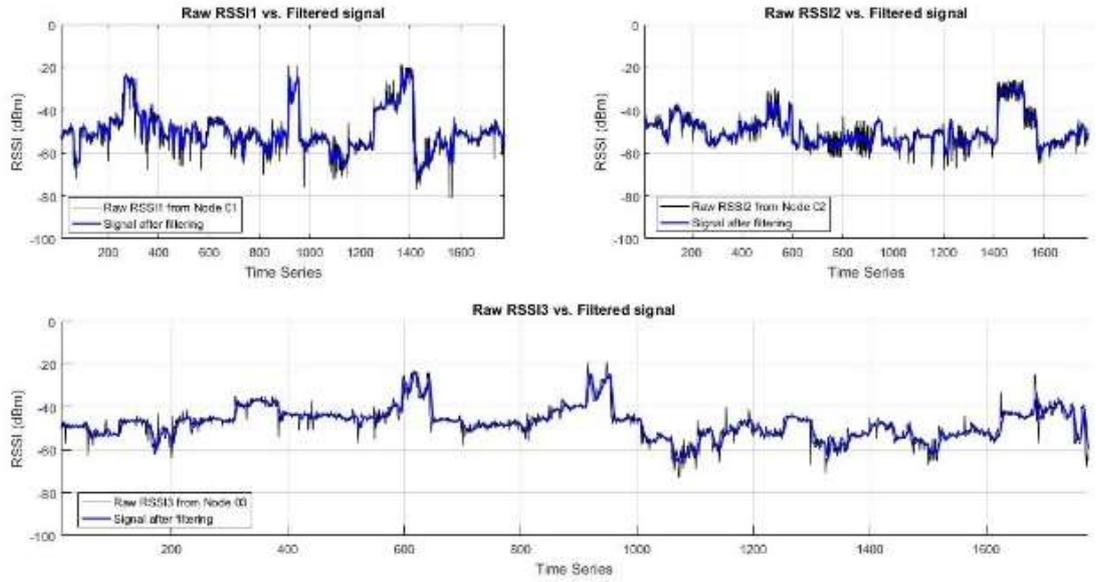

*Figure 19:RSSI data with moving average filter*

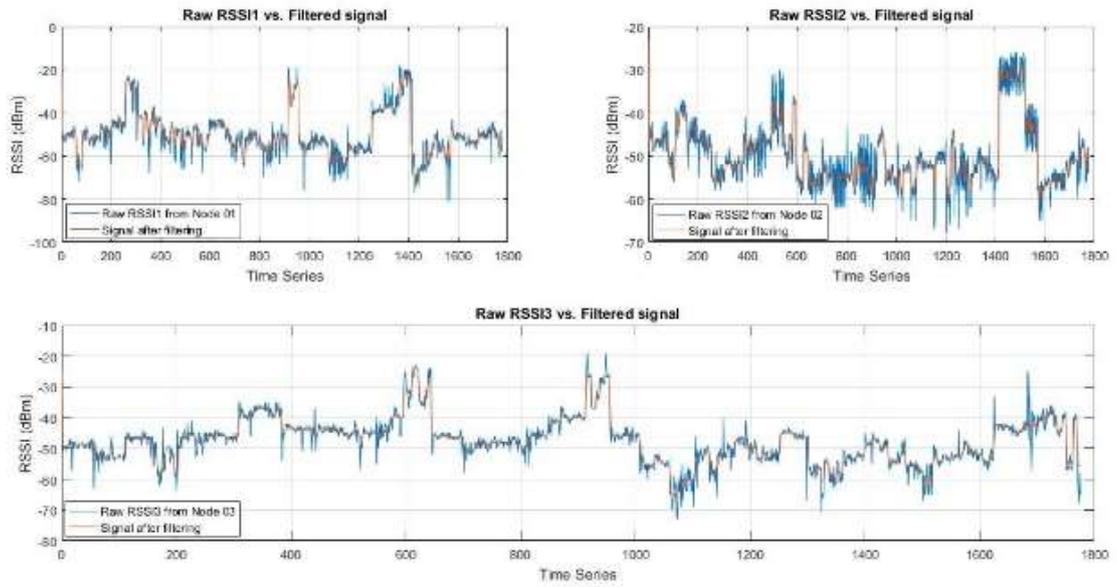

*Figure 20:RSSI data with median filter*

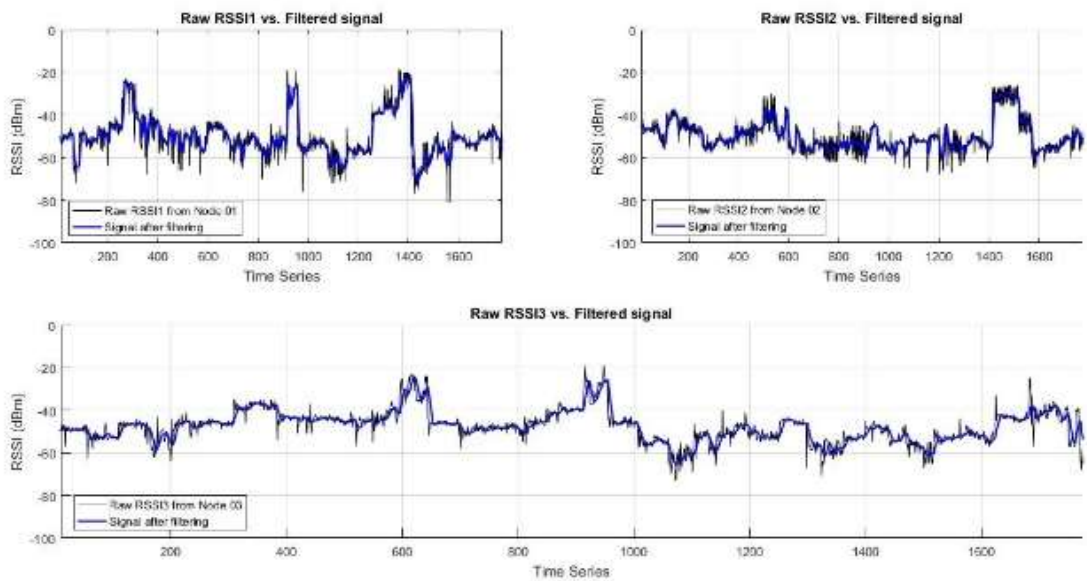

*Figure 21:RSSI data with gaussian filter*

## 4.2.2. Model Training Evaluation

Table 3 shows the results when the number of reference nodes equals 3. The RMSE value decreases in DTR as the number of trees increases and $R^2$ increases. Also, the error decreases in RFT as the number of forests increases. Tables 4 show the RMSE and R2 values when the number of reference nodes increases to 4 and 5, respectively. The RMSE and $R^2$ are improving as the number of reference nodes increases in the experimental test-bed. Figures 20 to 28 show each algorithm's error distribution of x coordinates and y coordinates. Interestingly, error distribution is optimal at DTR supervised ML algorithm.

*Table 3:Statistical comparison of each algorithm (Number of reference nodes=3)*

| ML Algorithm | Root Mean Squared Error (RMSE) | $R^2$ |
| --- | --- | --- |
|  |  |  |

|  |  | x | y | x | y |
|---|---|---|---|---|---|
| Linear Regression |  | 77.55 | 71.10 | 0.2910 | 0.4292 |
| Polynomial Regression |  | 65.93 | 52.14 | 0.4393 | 0.6502 |
| Support Vector Regression |  | 71.11 | 65.22 | 0.3600 | 0.4802 |
| Decision Tree Regression | No. of Trees=15 | 27.31 | 27.10 | 0.8959 | 0.9191 |
|  | No. of Trees=25 | 27.03 | 25.93 | 0.9139 | 0.9123 |
|  | No. of Trees=35 | 24.12 | 23.17 | 0.9598 | 0.9478 |
| Random Forest Regression | No. of Forests=40 | 28.92 | 29.11 | 0.8403 | 0.9191 |
|  | No. of Forests=70 | 28.01 | 28.31 | 0.8923 | 0.9298 |
|  | No. of Forests=100 | 28.11 | 28.02 | 0.8945 | 0.9187 |

Table 4: Statistical comparison of each algorithm (Number of reference nodes=4)

| ML Algorithm | | Root Mean Squared Error (RMSE) | | $R^2$ | |
|---|---|---|---|---|---|
| | | x | y | x | y |
| Linear Regression | | 77.73 | 71.84 | 0.2664 | 0.4183 |
| Polynomial Regression | | 69.16 | 57.54 | 0.4192 | 0.6268 |
| Support Vector Regression | | 73.37 | 66.84 | 0.34642 | 0.49634 |
| Decision Tree Regression | No. of Trees=15 | 29.33 | 28.52 | 0.8955 | 0.9083 |
| | No. of Trees=25 | 28.12 | 27.84 | 0.9040 | 0.9126 |
| | No. of Trees=35 | 28.16 | 27.87 | 0.9037 | 0.9124 |
| Random Forest Regression | No. of Forests=40 | 31.14 | 29.09 | 0.8822 | 0.9046 |
| | No. of Forests=70 | 30.84 | 28.80 | 0.8844 | 0.9064 |
| | No. of Forests=100 | 30.90 | 28.70 | 0.8840 | 0.90714 |

Table 5: Statistical comparison of each algorithm (Number of reference nodes=5)

| ML Algorithm | | Root Mean Squared Error (RMSE) | | $R^2$ | |
|---|---|---|---|---|---|
| | | x | y | x | y |
| Linear Regression | | 77.73 | 71.84 | 0.2664 | 0.4183 |
| Polynomial Regression | | 69.16 | 53.29 | 0.4192 | 0.6268 |
| Support Vector Regression | | 73.37 | 66224 | 0.34642 | 0.49634 |
| Decision Tree Regression | No. of Trees=15 | 27.33 | 27.52 | 0.8955 | 0.9083 |
| | No. of Trees=25 | 27.12 | 26.84 | 0.9040 | 0.9126 |
| | No. of Trees=35 | 28.16 | 27.87 | 0.9037 | 0.9124 |
| Random Forest Regression | No. of Forests=40 | 30.14 | 28.09 | 0.8822 | 0.9046 |
| | No. of Forests=70 | 29.44 | 28.30 | 0.8844 | 0.9064 |
| | No. of Forests=100 | 29.40 | 28.10 | 0.8840 | 0.90714 |

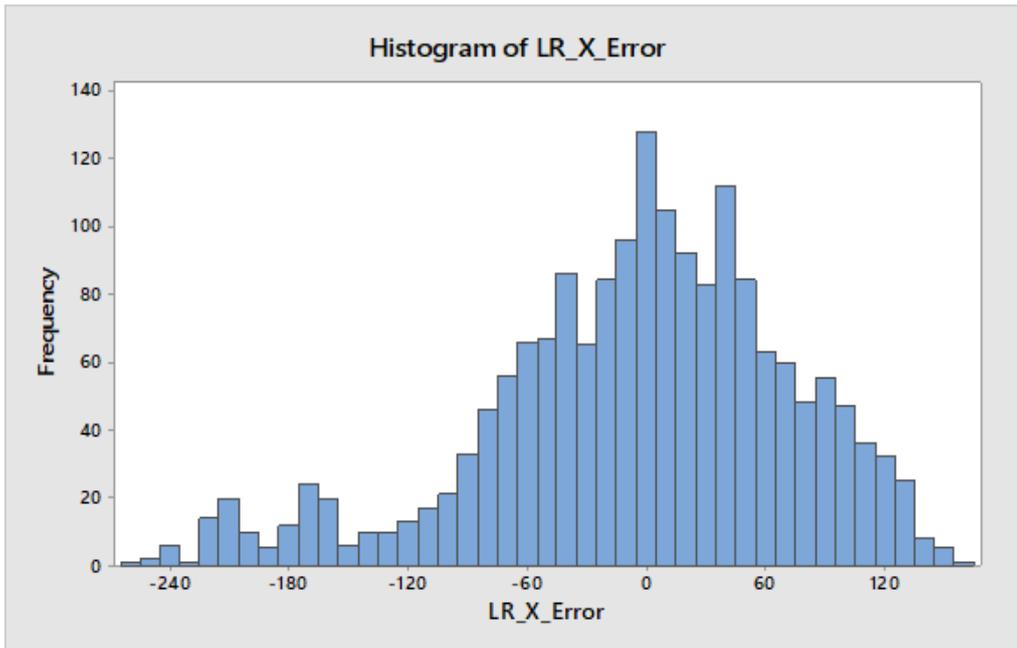

*Figure 22: Error distribution of x coordinate in LR*

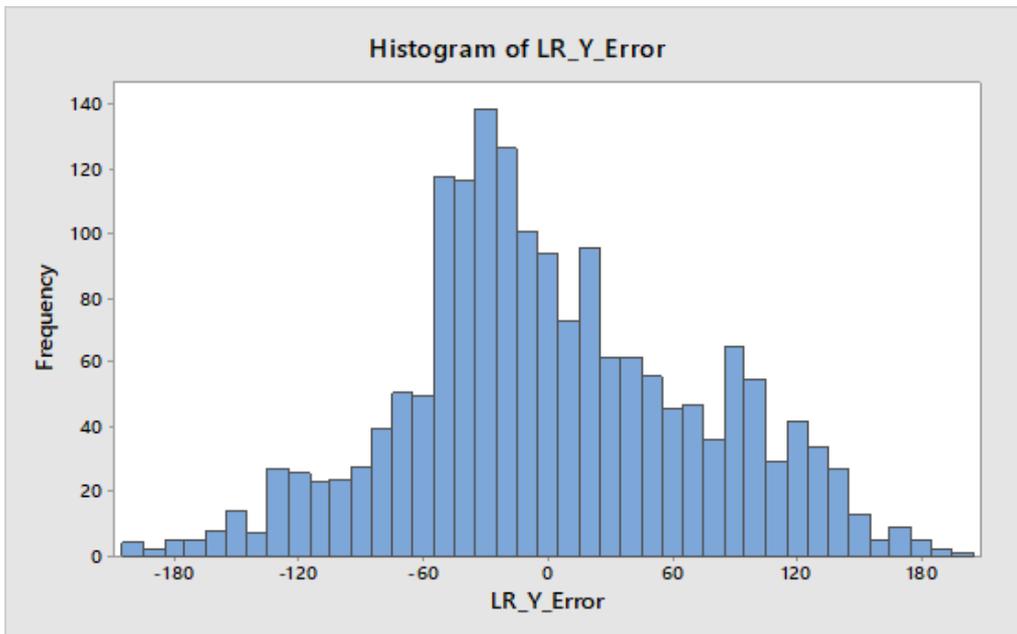

*Figure 23:Error distribution of y coordinate in LR*

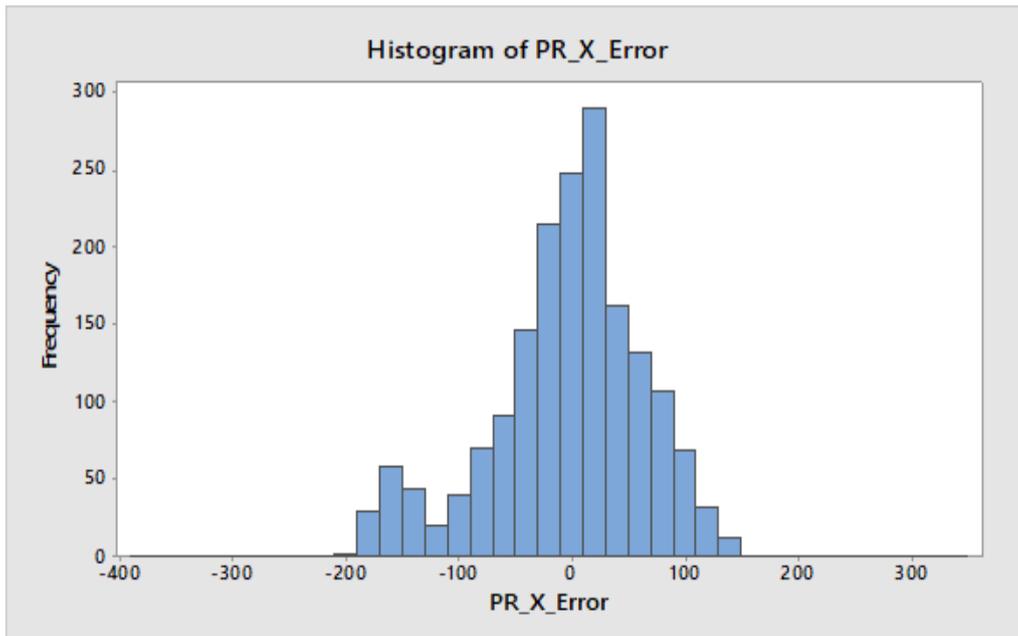

*Figure 24:Error distribution of x coordinate in PR*

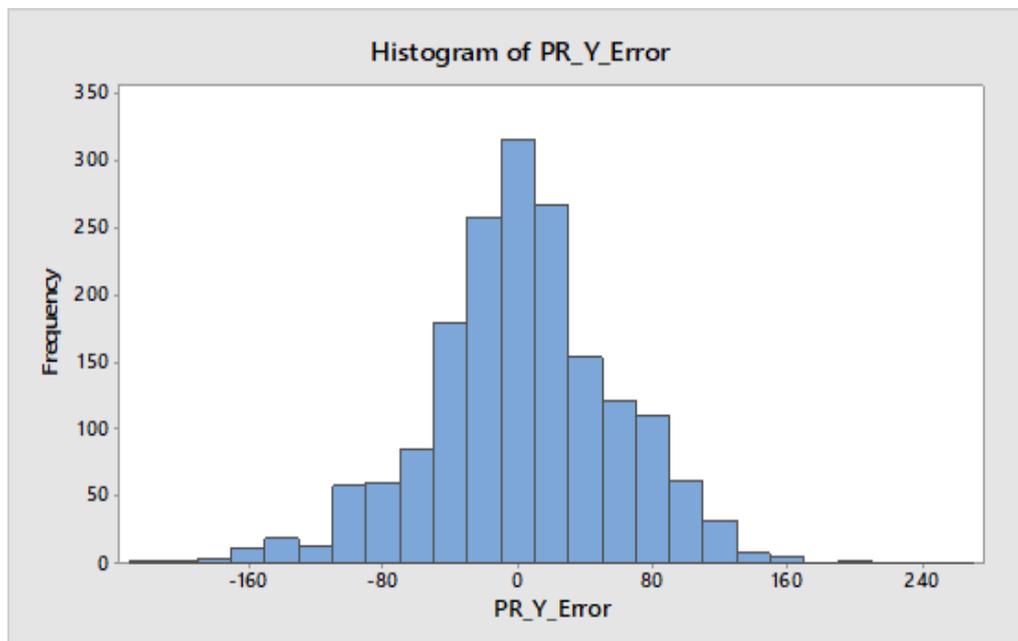

*Figure 25: Error distribution of y coordinate in PR*

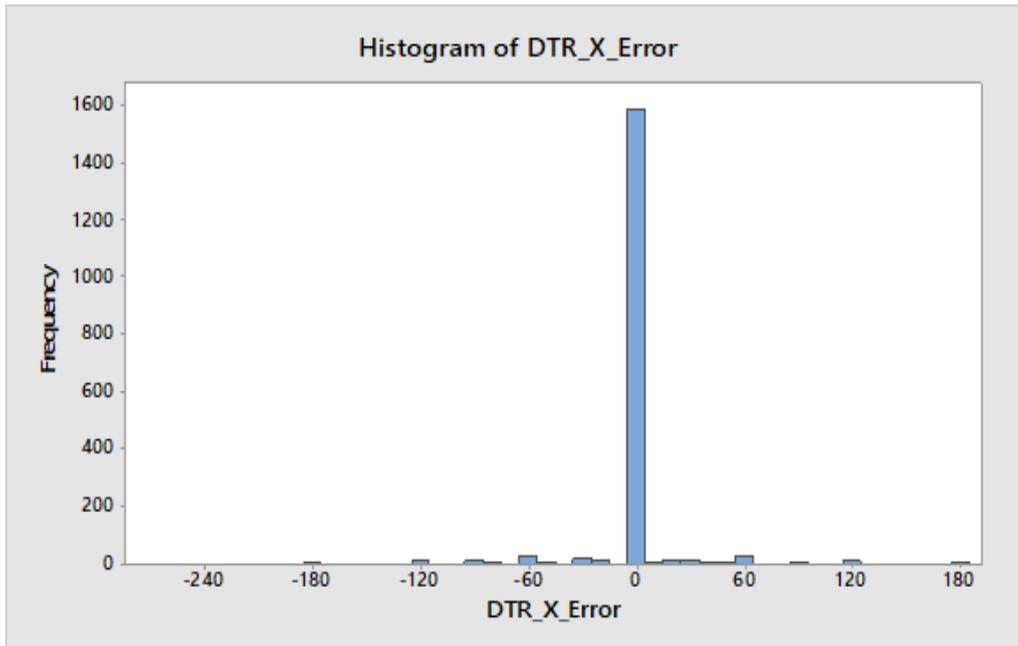

*Figure 25:Error distribution of x coordinate in DTR*

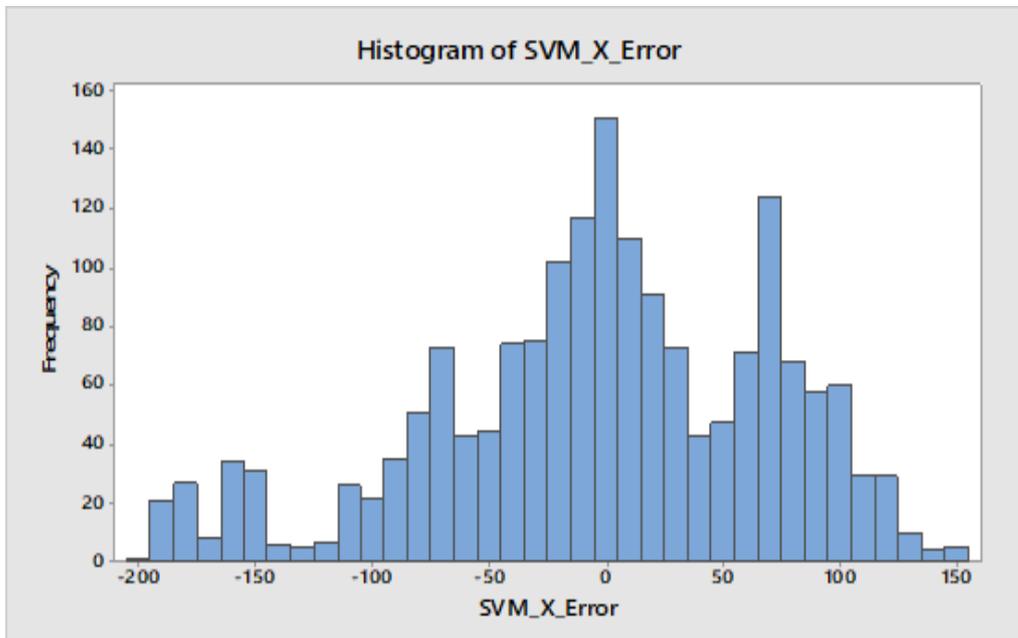

*Figure 26:Error distribution of y coordinate in SVR*

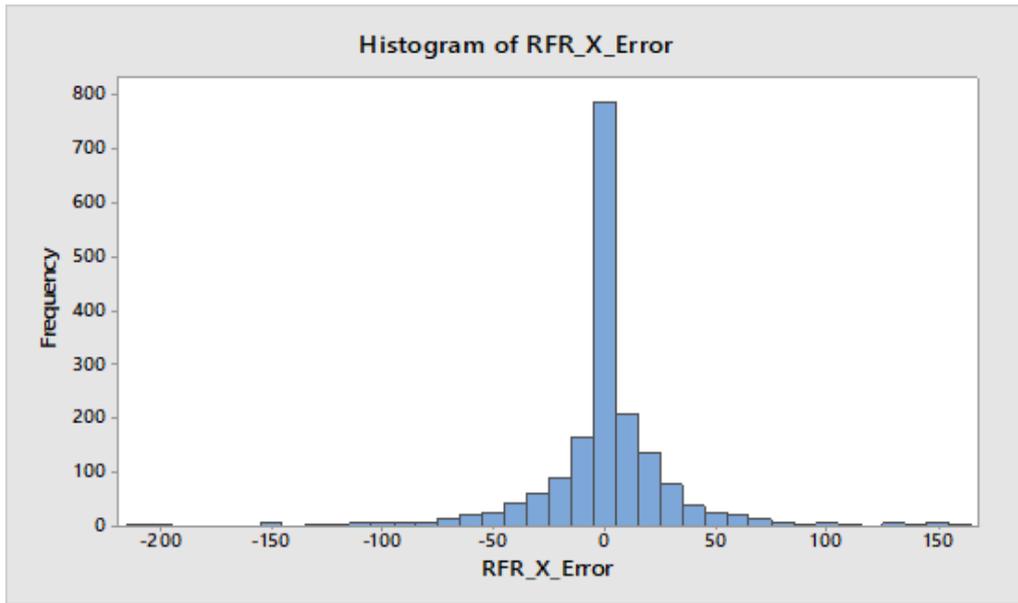

*Figure 27:Error distribution of x coordinate in SVR*

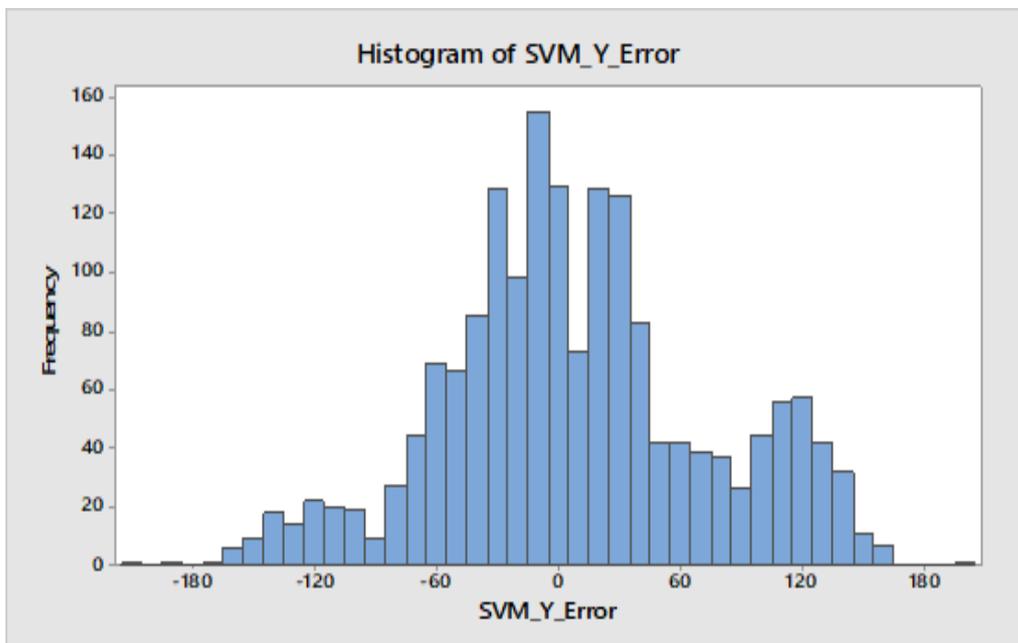

*Figure 28:Error distribution of x coordinate in RFR*

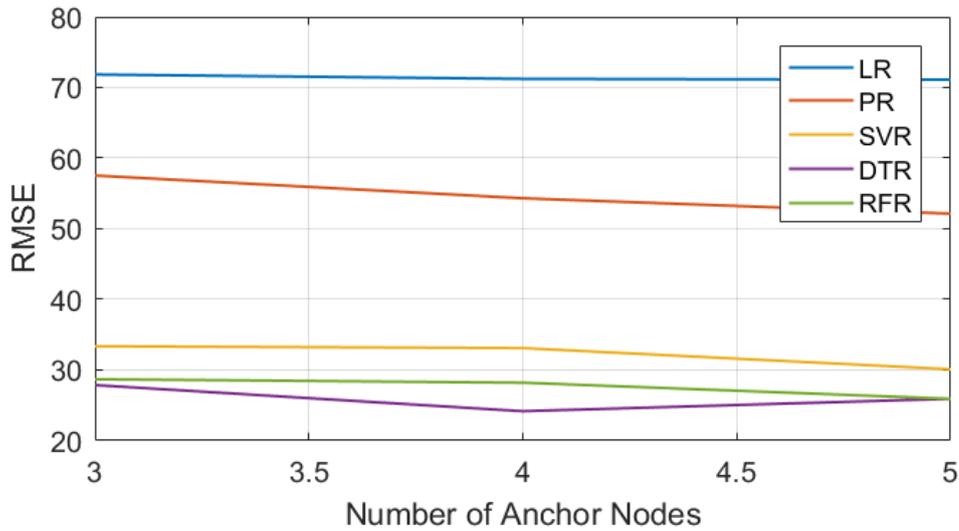

*Figure 30: Number of Anchor nodes Vs. RMSE x-coordinate*

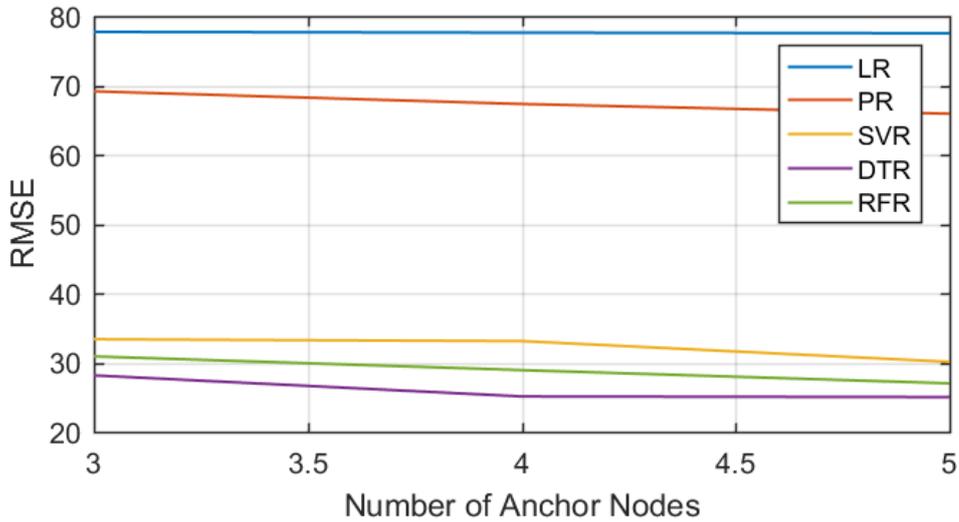

*Figure 31: Number of Anchor nodes Vs. RMSE y-coordinate*

Fig.30 and 31 above denote the changes of RMSE values in the x coordinate as we change the number of anchor nodes for the x coordinate and y coordinate respectively. In the experimental setup, we changed the number of anchor nodes 3, 4, and 5 respectively. In each case, RSSI values were collected and trained using ML models. It observed that as the number of anchor nodes increases, there is a significant reduction in RMSE values for all the models. The LR and PR show the higher RMSE values and SVR, DTR, and RFR show relatively lower RMSE values. Where DTR outperformed in terms of RMSE. The reason for this trend is, that when the number of trainable parameters increases, the model train very well.

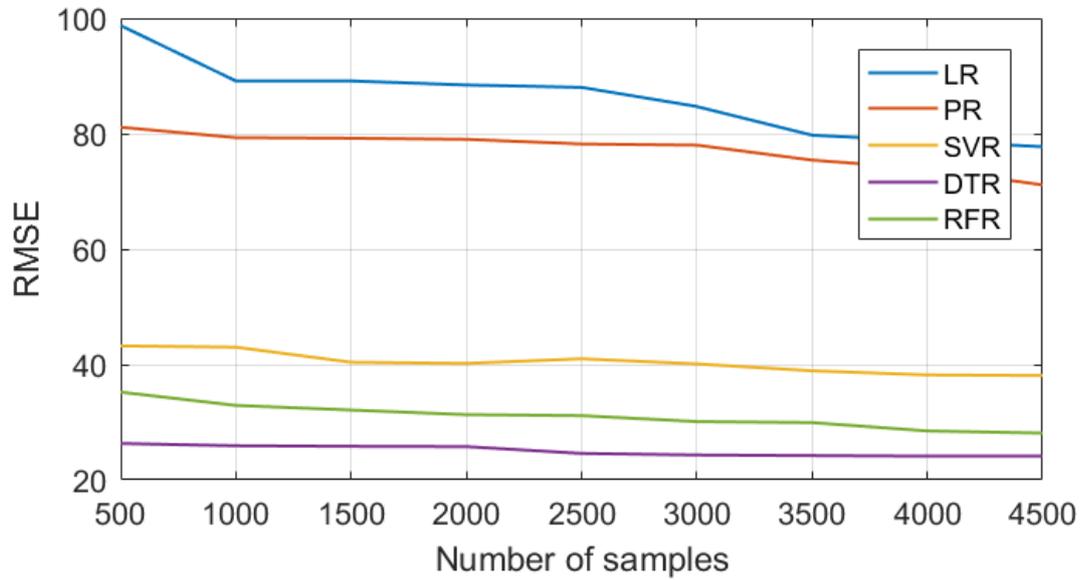

Figure 29:Number of Samples Vs. RMSE, x coordinate

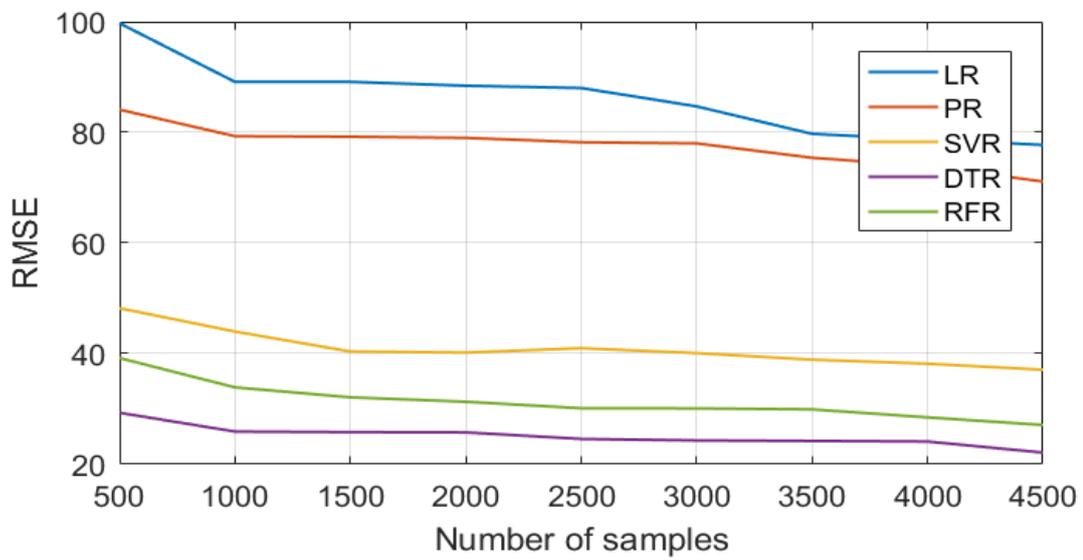

Figure 30:Number of samples vs. RMSE y coordinate

Fig.32 and 33 above denote the variation of RMSE value against the sample size for x coordinate and y coordinate respectively. It's observed that RMSE in decreasing as the number of samples increases in all the models. LR and PR showed relatively high RMSE and SVR, and DTR and RFR have shown lowest RMSE values. Where DTR is outperformed for both coordinates giving the lowest RMSE value. For all the models RMSE value is decreasing as the number of samples increases. In ML models, as the number of samples increases, the standard deviation is decreasing.

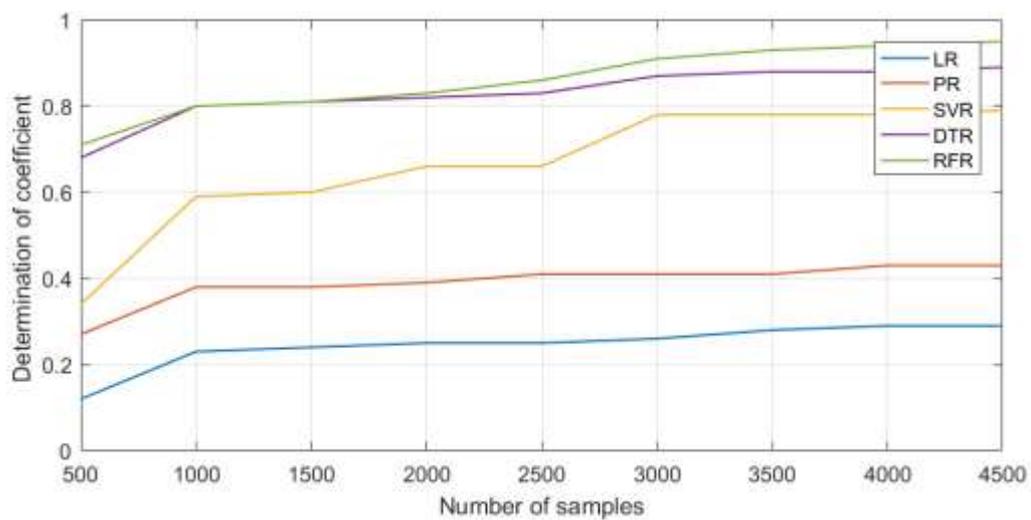

*Figure 31:Number of Samples vs. R2 x coordinate*

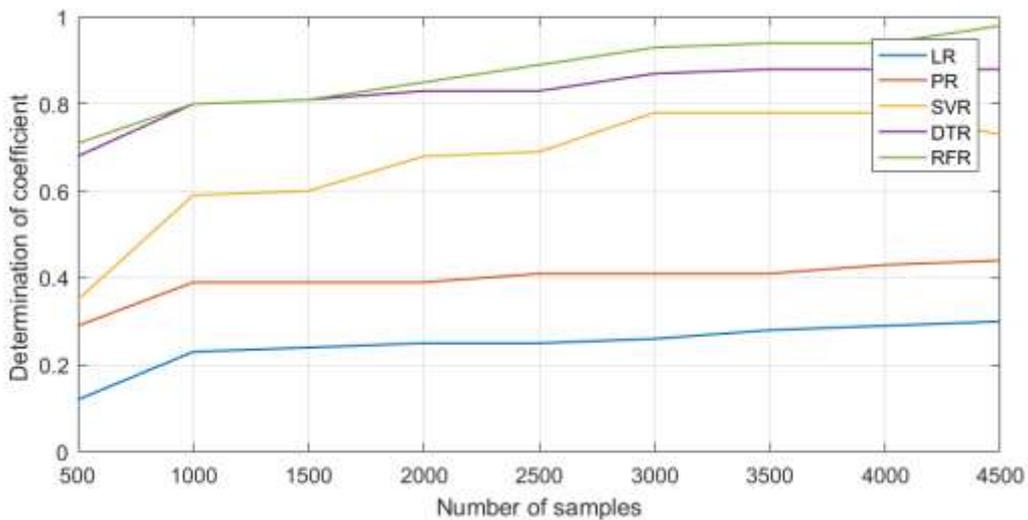

*Figure 32:Number of Samples vs. R2 y coordinate*

Fig.34 and 35 show the change of determination of coefficient value against the number of samples for x coordinate and y coordinate respectively. For machine learning models, the coefficient of determination, or R-squared value, ranges from 0.0 to 1.0 and reflects the correlation of the variance proportionate to the real and estimated node position. All dataset points perfectly lie at the estimated line of best fit when the R-squared values are closer to 1.0, indicating that the estimated position is entirely defined concerning the higher accuracy. For all the models, $R^2$ values rapidly increase till 1000 samples and after 1000, they increase normally. DTR and RFR show better $R^2$ score which is closer to 1. LR and PR are showing less than 0.5, which means that models are not fit well with the data.

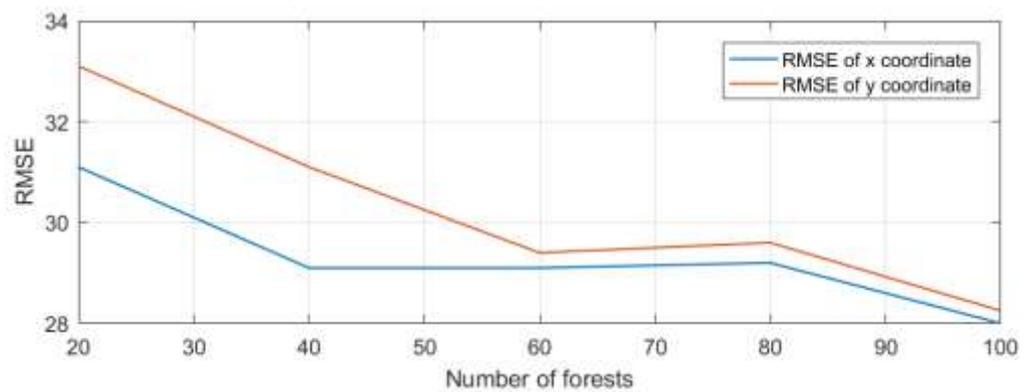

*Figure 33:Number of forest vs. RMSE*

Fig. 36 shows the impact of the hyper-parameter and the number of forests in RFR against the accuracy of the estimation. It can be observed that as the number of forests increases, RMSE is significantly decreasing. In RFR as the number of forests increases, the model is well trained with the data and it gives better accuracy. However, with a high number of forests, the model required a higher computational power in hardware devices.

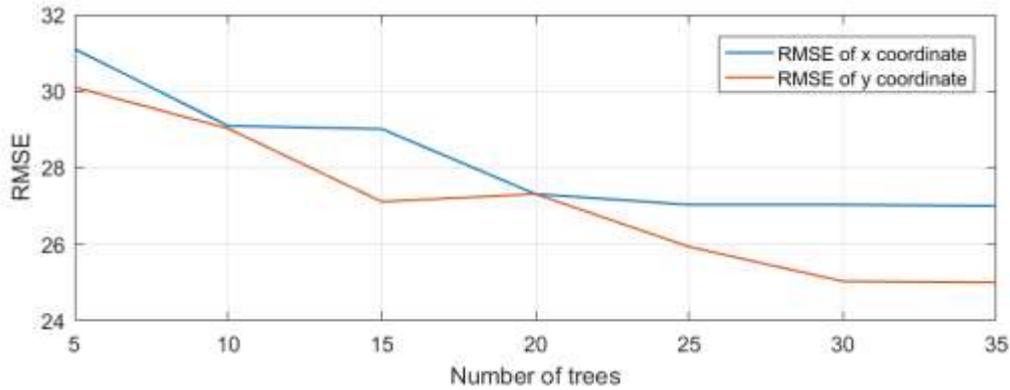

*Figure 34:Number of trees vs. RMSE*

The number of tree hyper-parameters used in tree-based ensemble methods must be adjusted, and this has a direct bearing on the computational cost. To find a trade-off between forecast accuracy and computational time, sufficient trees must be chosen. According to the foundations of tree-based algorithms, a model with more trees will be optimized and have the lowest possible prediction error. It shows that model performance is dependent on the maximum tree depth and that deeper trees perform better. The figure above illustrates the impact of the number of trees vs. RMSE in the DTR algorithm. It can observe that RMSE is significantly decreasing as the number of trees increases.

In conclusion, a mobile sensor node produces RSSI data, which is collected in an experimental test bed, and several fixed sensor nodes in the test bed receive those RSSI values. An MQTT server was used to collect RSSI data while the mobile node was maintained in 32 places that were known to exist. To prepare collected data for supervised ML training, filters and pre-processing were applied. After the experiment with different supervised algorithms under different conditions, it becomes clear that Decision Tree Regressor (DTR) was the best-outperformed algorithm compared to the rest of the algorithms tested. The number of forests in DTR matters to improve the location estimation accuracy and its significance in reducing the error. It was observed that once the number of reference nodes increased in the test bed, the accuracy and error were significantly improved. We foresee using supervised ML algorithms to improve results rather than deterministic localization based on our experiments.

## 4.3. Experiments on BLE, Zigbee, and LoRaWan-based Indoor Environment

Integrating technological advances into the building can be combined with many applications to improve humans' living standards. For example, they are tracking a person's location in a shopping complex, tracking the daily activity of an elder person living alone in a house, tracking autonomous robots in an indoor environment, etc. In the recent development of the Internet of Things (IoT), wearable smart devices built on wireless technologies such as Wi-Fi, Bluetooth Low Energy (BLE), Zigbee, LoRaWAN, etc. These devices can communicate data with the IoT network. Such data transmitted through the web could be information on building health, weather conditions, or other sensing information. When a sensor and the base station are connected, the signal strengths of each wireless link can be measured. In indoor localization, it uses the signal strength as an input to compute the geographical location of that mobile sensor.

An indoor positioning system is used to locate stationary or moving objects and devices in an environment where the Global Positioning System (GPS) cannot be applied. GPS is appropriate when it is used in outdoor positioning-related applications. However, it consumes much energy, and implementation is costly for each node in an extensive network. Moreover, GPS is highly dependent on line-of-sight (LOS) and cannot be used indoors. In addition, GPS is allowed only a maximum of 5 meters. Therefore, this may be suitable for the outdoors. Many applications initiate indoor positioning systems in the following areas, such as hospitals that can perform indoor positioning to track the patient. Where the doctor will know accurately place the patient in a location within the building.

Another example is the real-time tracking of older people when their life is inside the home. The gradients can monitor the real-time location of their elderly parents using their mobile phones through IoT servers. In the farming industry [18], indoor positioning can be used for animal tracking, military applications, etc. [12-13]. This technique's implementation cost is very low compared to the other monitoring mechanisms, such as image processing-based systems. In image processing-based systems always, the camera has to be focused on objects, and always object and camera should be in the line of sights

IoT devices tend to be compact in size. Hardware is typically modest as a result. They have minimal processing power, little storage capacity, and basic communication skills. As a result, the apparatus of these features must be taken into account by the localization method. If successful, creating an indoor positioning system needed simultaneously tracking several targets.

Various wireless technologies have been proposed and tested when performing indoor positioning literature. The most commonly used technologies are Wi-Fi, Bluetooth, Radio Frequency Identification (RFID), Bluetooth Low Energy (BLE), Zigbee, and LoRaWAN. But each of them has its strengths and weaknesses. Due to the building's high availability of access points, Wi-Fi has become the most straightforward option in such solutions. However, the purpose of deploying

Wi-Fi access points is usually to provide maximum coverage to Internet users. In this case, signal coverage is not sufficient for a localization application.

Zigbee and LoRaWAN have an ideal sensing range in comparison to Wi-Fi. However, employing these devices comes with a rather steep expense. Additionally, Wi-Fi uses a lot of the batteries' power.

This section compares indoor positioning accuracy using multiple supervised algorithms for the IoT systems developed using Zigbee, BLE, and LoRaWAN. Zigbee is a long-range and low-power technology typically used in IoT applications. LoRaWAN is a new technology that is not as popular as the previous one, transmitting at 915MHz with high data Speed. LoRaWAN node can reach a distance of 15000 meters, Limiting the number of nodes required for the sequence.

Based on the related literature, indoor localization primarily uses time-based, angle-based, RSS-based, or a combination of these technologies to obtain their signal measurements. The relationship between RSSI and distance is the key to wireless ranging and localization systems, where length is measured based on the signal strength received from each transmitting node. According to RSSI-based indoor positioning applications, triangulation and trilateration techniques primarily achieve mobile node position estimation. The Time of Arrival time (TOA) and Time Difference of Arrival (TDOA) are time-based measurements related to transmission time. The Angle of Arrival (AOA) -based position estimation system requires a complex directional antenna as a beacon node for angle measurement [13]. In literature, RSS-based multilateration positioning technology was the most popular algorithm due to its simplicity.

Machine learning is very suitable for predicting the expected target output using sample data, so when machine learning begins, researchers use neural networks to identify WSNs. Moreover, Kalman and extended Kalman filters have been used to filter RSSI data, and several Bayesian algorithms have been investigated for estimating the locations. Furthermore, Payal et al. used FFNN to develop WSN-based ANN localization techniques, a cost-effective localization framework[14].

Sebastian and Petros contributed to indoor positioning based on Zigbee, LoRaWAN, Wi-Fi, iand iBLE. They have designed individual systems in indoor environments and obtained RSSI ivalues. They have used a deterministic algorithm in the localization phase, trilateration to get the accurate location and presented error comparisons [16-17].

The RSSI measurements are volatile in terms of time and position, Consequently, it is challenging to suggest a reliable and precise positioning technique for all interior localization applications. Furthermore, the accuracy of related efforts for localization-based deterministic algorithms is poor. By streamlining the hardware design and reducing the complexity of the deterministic algorithms used to locate mobile nodes in an indoor setting, the suggested study investigates unanswered questions in the literature.

The proposed solutions for indoor localization based on deterministic and probabilistic algorithms are impractical to implement on real hardware devices. This is due to the complexity of proposed algorithms and hardware incompatibility. However, recently developed hardware devices such as programmable sensor nodes and single-board computers for IoT support machine learning computations.

This work has considered three types of wireless technologies used in IoT systems to collect RSSI data.

Bluetooth Low Energy (BLE) is a low-power wireless communication technology used over short distances. Specific smart wireless devices that work every day (smartphones, smartwatches, fitness trackers, wireless headphones, computers, etc.) use BLE to create a seamless experience between devices.

For the experiment testbed in [7], the ten beacon nodes were designed using Gimbal Beacon. The Gimbal Beacon is from the Apple iBeacon protocol. IBeacon data packet structure defines three fields: a universal unique identifier (UUID), a 16-byte lot used to identify a group of beacons. The second and third fields are the "primary" and "secondary" values.

Zigbee is low-cost, energy-saving, and can create mesh networks. It is a communication protocol based on the IEEE 802.15.4 standard for creating personal area networks with small antennas. The XBee is a type of sensor node based on Zigbee technology. Whereas XBee has low latency requirements and is easy to use, a device that allows you to quickly create a multipoint Zigbee network. In the experimental testbed in [6], it has used 2mW wired antenna XBees is used. Due to the limited processing power of XBees, Microcontrollers are essential for controlling the flow of information. Therefore, the microcontroller selected is Arduino Uno, with easy integration with XBee and a low power consumption Claim [26-27].

At lower transmission speeds, this technology was initially developed as LongRange by the LoRa Alliance Local Area Network (LoRaWAN) Protocol. The frequency is 915MHz [18]. Benefits of frequency of use lower than 2.4GHz because longer wavelengths are possible for traffic lights to pass through walls and obstacles. Then this makes the signal reachable far distance. The frequency of 915MHz is LoRaWAN is relatively free and does not interfere

LoRaWAN is more secure than other IoT wireless technologies because it can send encrypted data to numerous locations on a regular basis. As a result, the node interacts with other transmission tools. It is less vulnerable to noise when in use. It is ideally suited for applications like smart cities because of its vast transmission range. The data rate between nodes is decreased as a result of adopting such low frequencies.

In terms of cost, it's pretty high for LoRaWAN-based devices. Moreover, a large antenna and additional hardware are needed to access the media. Very effective for remote outdoor positioning, but short-range Indoor positioning may present some challenges. In terms of range, each wireless technology has its sensing range, as shown in table 5.

*Table 5: Transmission Range of the Wireless Communication Technologies*

| Wireless Technology | Range(m) |
|---|---|
| LoRaWAN | 10,000 |
| BLE | 60 |
| Zigbee | 100 |

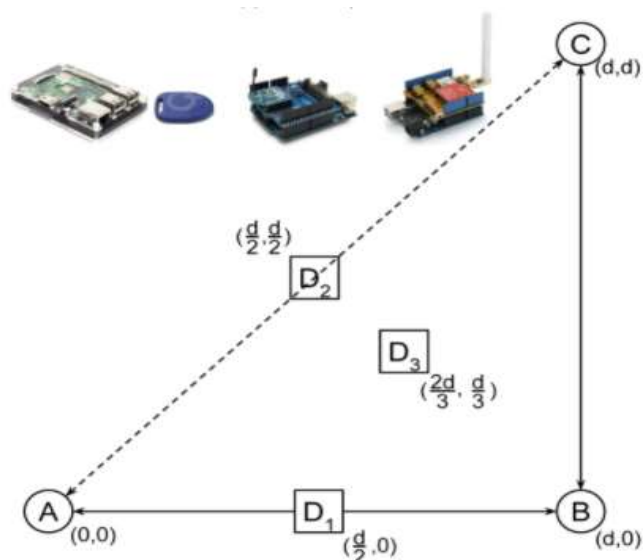

*Figure 35: Arrangement of sensor nodes and positions [9]*

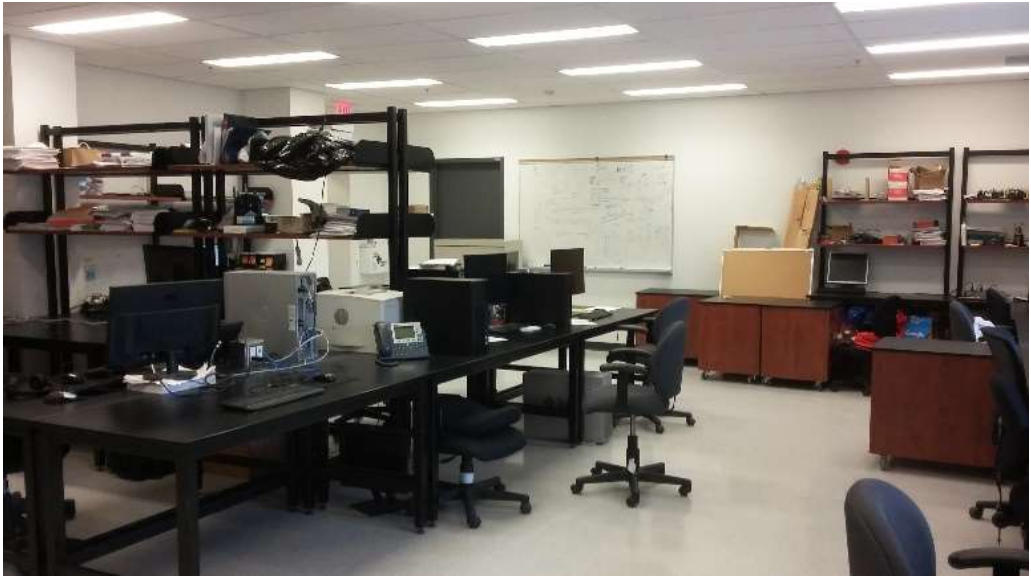
*Figure 36: Experiment environment [9]*

RSSI is recommended as one of the best approaches for indoor localization [7]. Its prominent popularity is that RSSI does not require additional hardware for signal measurement. The RSSI levels are measured by the receiver from the device's transmitter end. It is often used to determine the distance between a transmitter and a receiver because the signal strength decreases as the signal moves outward from the transmitter.

Because the propagated signal is susceptible to environmental noise, RSSIs usually lead to inaccurate values and errors in positioning systems—the relationship between the distance and RSSI is express in equation 33 [6]. In the localization scenario, we need to estimate the location of the reference node detecting the RSSI levels received from the mobile sensor node.

$$RSSI = -(10n)\,log10(d) + A, \quad (33)$$

Where n is the signal propagation constant, d is the distance in meters, and A is the offset RSSI reading at one meter from the transmitter.

### *4.3.1 Model Training*

Support Vector Regression (SVR) uses the same classification principles as SVM, with some differences. First, because the output is accurate, the information at hand is difficult to predict and has endless possibilities. SVR is a robust supervised learning algorithm that allows selecting an error tolerance by accepting the margin of error and adjusting the margin of error that exceeds the margin of error. For regression, the margin of error ($\varepsilon$) is set to approximate the SVM requested by the problem[5-10].

In Decision Tree Regressor, decision trees form a learning tree structure for solving classification or regression problems. The model divides the training data into several labels according to the creation rules. After creating the tree structure, predict the new data label by traversing the input data in the training tree. The information flow in the decision tree is so transparent that users can easily correlate assumptions without any analysis background [33-38].

Random Forest Regression (RFR) is a supervised machine learning algorithm that uses ensemble learning methods for classification and regression It works by creating many decision trees during training and testing each tree's class (classification) or average prediction (regression) model. This is one of the most accurate learning algorithms available. Many datasets produce very accurate classifiers when using this algorithm. It could be run efficiently on large databases. It can handle thousands of input variables without removing them [10-11].

### *4.3.1 Experimental Results*

The RSSI values received from the mobile sensor node at positions D1, D2, and D3 are used as the feature to train models. These RSSI values are collected by reference nodes placed at fixed points, as shown in figure. In this work, RSSI data were trained using supervised algorithms DTR, RFR, and SVR, and a comparison of errors of each location D1, D2, and D3 is shown in table 6, table 7, and table 8, respectively. The errors of positioning are calculated based on equation above. The Jupyter Notebook (Python 3) was used to train the algorithms [12]. The experimental results present valuable insights in terms of accuracy. BLE was the most accurate wireless technology compared to the other two. However, BLE has a minimal distance of operation. Therefore, BLE is suitable for short-range indoor localization applications.

Further, BLE consumes very little power [7]. Thus, it was prolonging the senor uptime. While Zigbee shows average errors, LoRaWAN has the highest estimation errors.

$$Error = \sqrt{(x_{predict} - x_{real})^2 - (y_{predict} - y_{real})^2} \quad (34)$$

*Table 6: Error comparison for BLE*

| Test Point | Actual Coordinates | | Errors (m) | | |
|---|---|---|---|---|---|
| | x | y | DTR | RFR | SVR |
| D1 | 0.500 | 0.000 | 0.116 | 0.089 | 0.189 |
| D2 | 0.500 | 0.500 | 0.013 | 0.011 | 0.602 |
| D3 | 0.667 | 0.333 | 0.167 | 0.124 | 0.478 |
| Average | | | 0.432 | 0.323 | 0.423 |

*Table 7: Error comparison for Zigbee*

| Test Point | Actual Coordinates | | Errors (m) | | |
|---|---|---|---|---|---|
| | x | y | DTR | RFR | SVR |
| D1 | 0.500 | 0.000 | 0.193 | 0.223 | 0.394 |
| D2 | 0.500 | 0.500 | 0.113 | 0.299 | 0.403 |
| D3 | 0.667 | 0.333 | 0.303 | 0.982 | 0.384 |
| Average | | | 0.536 | 0.501 | 0.393 |

*Table 8: Error comparison for LoRaWAN*

| Test Point | Actual Coordinates | | Errors (m) | | |
|---|---|---|---|---|---|
| | x | y | DTR | RFR | SVR |
| D1 | 0.500 | 0.000 | 0.993 | 0.523 | 1.932 |
| D2 | 0.500 | 0.500 | 1.093 | 0.521 | 0.928 |
| D3 | 0.667 | 0.333 | 0.890 | 0.732 | 1.993 |
| average | | | 0.992 | 0.592 | 1.617 |

## 4.4 Summary

This chapter compared RSSI-based indoor localization based on the wireless technologies BLE, LoRaWAN, and Zigbee for use in indoor localization systems. The experiments involved the RSSI data from three reference nodes built on the above wireless technologies. Supervised learning techniques were investigated to estimate a mobile node's geographical location. When comparing the localization accuracy, all algorithms tested in this experiment give fairly good error values of less than one meter. When comparing the technologies, BLE outperformed based on the results, achieving the lowest error among all the supervised algorithms. It is observed that one algorithm cannot propose as the best because different algorithms are outperformed for each technology.

Moreover, BLE considers the most inadequate power-consuming technology. This experiment has only considered the 2D environments. A study on localization for a 3D environment would be an interesting future research direction.

# CHAPTER 5:

# 5  Supervised Classifiers for Indoor Localization

## 5.1. Introduction

Machine learning algorithms can be categorized into supervised learning and unsupervised learning. The supervised learning algorithms required a labeled dataset for predictions, and unsupervised algorithms used unlabeled data sets for their predictions. The ML lifecycle has several stages: feature engineering (data pre-processing), model training, hyperparameter tuning, and model evaluation. Machine learning algorithms can be used for classification or regression. In classification, machine learning algorithms classify data into different categories, while regression learns from training data to predict continuous variables. There are existing works on using ML as both a classification and regression problem [13-20]. Where the object's location can be estimated as a numerical value(regression) or as a classification (classifiers), for example, consider a person living inside a house and needing to get their location through an IoT system. The location of this person can be estimated geographically with (x,y) coordinates or can predict the location as the bedroom, kitchen, etc.

## 5.2. Supervised Classifiers for Indoor Localization

Artificial neural networks are a branch of artificial intelligence inspired by biology and fashioned after the brain. A computational network based on biological neural networks, which create the structure of the human brain, is typically referred to as an artificial neural network. Artificial neural networks also feature neurons that are linked to each other in different layers of the networks, just as neurons in a real brain. Nodes are the name for these neurons. A vast number of artificial neurons, also known as units, are placed in a hierarchy of layers to form what is known as a neural network. Let's examine the many layers that can be found in an artificial neural network.

Feed-Forward Neural Network (FFNN) is considered a sub-branch of ANN. A feed-forward network is a type of neural network that consists of at least one layer of neurons and input and output layers. The network's intensity can be observed based on the collective behavior of the connected neurons. The output is chosen by evaluating the network's output in its input context. The main benefit of this network is that it learns to assess and identify input patterns. Generally speaking, an artificial neural network has three layers named input layer, the hidden layer, and the output layer. Few works in the literature have used FFNN for indoor localization, giving considerably good results [18-20].

In this work, we have used a publicly available RSSI data set collected from BLE ibeacon nodes [18]. On the first floor of Western Michigan University's Waldo Library, there were thirteen ibeacon nodes. Data was gathered with an iPhone 6S. There are essentially four zones in the library space. The FFNN was used to train the data and test the neural network's classification accuracy in a variety of scenarios.

Received signal strength Indicators (RSSI) is a well-known location estimation approach in which power intensity value is assessed from the beacons and measured in dBm. The RSSI measurements for one particular location of the target node are collected, as shown in Figure 40.

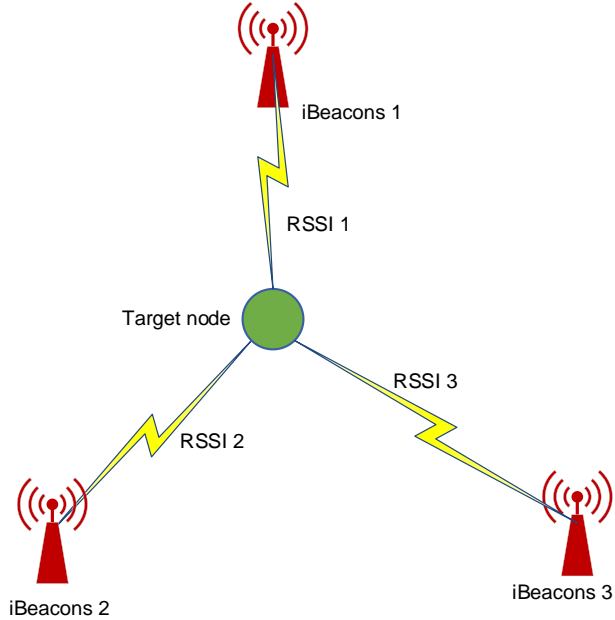

*Figure 37:Arrangement of Anchor node and target nodes.*

The RSSI value defines the distance between beacons and the target node, and the value increases accordingly as the distance increases. The logarithmic distance loss model is calculated as:

$$RSSI_r(p) = RSSI_r(p_0) - 10\eta \log \frac{p}{p_0} + X_\sigma \qquad (35)$$

where $RSSI_r(p)$ defined as received power [dBm] at a distance $p$ [m] from the transmitter, $X_\sigma$ defined as zero-mean Gaussian random variable with the variance of $(\mathbb{N}(0,\sigma^2))$, $RSSI_r(p_0)$ refers as received power [dBm] at the reference distance $p_0$ from the transmitter, and $\eta$ is the path loss exponent which varies from 2 in free space to 4 in indoor environments. The localization algorithm is applied to estimate the location of the target node with the distances to different beacons. In this study, we proposed a weighted hyperbolic algorithm for RSSI-based localization. The basic idea of the proposed algorithm is to find the position $(a,b)$ of the target node that decreases the sum of the squared error values. Let $p_n = (a_n, b_n)$ be the location of beacons such that $n$ ($n = 1, 2, \ldots, N$ where $N$ is the total number of beacons and $p_n$ is the estimated location of the beacons. Therefore, the square of the distance between the target node and the beacons node $n$ is calculated as

$$p_n^2 = (a_n - a)^2 + (b_n - b)^2 \quad (36)$$

The starting point is used at the beacon $n=1$, such that, $a_1 = b_1 = 0$. Therefore, for $n>1$

$$p_n^2 - p_1^2 = a_n^2 - 2aa_n + b_n^2 - 2bb_n \quad (37)$$

Therefore, equation (3) can be written in Matrix as

$$\begin{bmatrix} 2a_2 & 2b_2 \\ \vdots & \vdots \\ 2a_N & 2b_N \end{bmatrix} \begin{bmatrix} a \\ b \end{bmatrix} = \begin{bmatrix} a_2^2 + b_2^2 - p_2^2 + p_1^2 \\ \vdots \\ a_N^2 + b_N^2 - p_N^2 + p_1^2 \end{bmatrix} \quad (38)$$

The above equation (4) can be written as

$$M.A = C \quad (39)$$

where $M = \begin{bmatrix} 2a_2 & 2b_2 \\ \vdots & \vdots \\ 2a_N & 2b_N \end{bmatrix}$, $A = \begin{bmatrix} a \\ b \end{bmatrix}$ and $C$ defines random vector as

$$C = \begin{bmatrix} a_2^2 + b_2^2 - p_2^2 + p_1^2 \\ \vdots \\ a_N^2 + b_N^2 - p_N^2 + p_1^2 \end{bmatrix} \quad (40)$$

Thus, the location of the target node is estimated by using the linear least square method as:

$$A = (M^T M)^{-1} M^T C \quad (41)$$

The localization is dependent on $(M^T M)^{-1}$. The expression $M^T M$ is a square, non-singular, and invertible matrix if the beacons are reasonably far from the target node. By applying weighted least-squares in equation (5) as

$$A = (M^T R^{-1} M)^{-1} M^T R^{-1} C \quad (42)$$

where R defines the covariance matrix of vector $C$ and is written as

$$R = \begin{bmatrix} Var(p_1^2) + Var(p_2^2) & Var(p_1^2) & \ldots & Var(p_1^2) \\ Var(p_1^2) & Var(p_1^2) + Var(p_3^2) & \ldots & Var(p_1^2) \\ \vdots & \vdots & \ddots & \\ \overline{Var(p_1^2)} & \overline{Var(p_1^2)} & \ldots & \overline{Var(p_1^2) + Var(p_N^2)} \end{bmatrix} \quad (43)$$

Therefore, the value of covariance matrix R is evaluated as

$$Var(p_n^2) = E\left[p_n^4\right] - \left(E\left[p_n^4\right]\right)^2 \quad (44)$$

Now, by using the channel condition model, the estimated location $p_n$ of equation (44) is

$$p_n = p_n . 10^{\frac{N(0,\sigma)}{10\eta}} = 10^{N\left(\log 10 p_n, \frac{\sigma}{10\eta}\right)} = e^{N\left(\log 10 p_n, \frac{\sigma}{10\eta}\right)} = e^{N\left(\ln p_n, \frac{\sigma \ln 10}{10\eta}\right)} \quad (45)$$

where, $p_n$ defines lognormal random variable along with parameters $\sigma_p = \ln p_n$ and $\sigma_p = \dfrac{\sigma \ln 10}{10\eta}$. The jth moment of a lognormal random variable of parameters $(\mu_p, \sigma_p)$ is given by $\mu_j = e^{j \cdot \mu_p + \frac{1 j^2 \sigma_p^2}{2}}$. Therefore

$$E\left[ p_n^4 \right] = \exp\left(4\mu_p + 8\sigma_p^2\right) \quad (46)$$

$$E\left[ p_n^2 \right] = \exp\left(2\mu_p + 2\sigma_p^2\right) \quad (47)$$

Then, putting equation (12) and (13) in equation (10) as:

$$Var\ (p_n^2) = E\left[ p_n^4 \right] - \left(E\left[ p_n^4 \right]\right)^2 = \exp\left(4\mu_p\right) \cdot \left(\exp\left(8\sigma_p^2\right) - \exp\left(4\sigma_p^2\right)\right) \quad (48)$$

Here, $\mu_p$ dependent on the actual location $p_n$ and estimation location $p_n$ of the target node and the iBeacons. Therefore, the proposed weighted hyperbolic algorithm solves the non-linear problem of RSSI-based localization and provides a precise estimation of the target location.

K-Nearest Neighbor (kNN) is the most basic and simplest of all indoor localization algorithms. It is an enhanced version of the fingerprinting method that performs location determination utilizing RSSI value. Here, nearest neighbors are those data points with a minimum distance in feature space from our new data point.

And K is the number of such data points we consider in implementing the algorithm. For example, a number k is selected by KNN algorithms as the nearest neighbor to the data point that needs to be categorized. If k is 3, it will search for the three closest neighbors to that data point. The kNN algorithm requires a predefined table that references the strength of two or more different received signals from local routers. The predefined table is constructed during the offline phase [69-73].

## 5.3. Experimental Setup

The dataset used in this experiment was taken from [15]. The dataset was created using the RSSI readings of an array of 13 ibeacons on the first floor of Waldo Library, Western Michigan University, as shown in Fig.41. Data has been collected using iPhone 6S. The dataset contains two sub-datasets: a labeled dataset (1420 instances) and an unlabeled dataset (5191 cases). The recording was performed during the general operating hours of the library. More significant RSSI values indicate closer proximity to a given iBeacon. For out-of-range iBeacons, the RSSI is indicated by -200.

The attached figure depicts the layout of the iBeacons as well as the arrangement of locations. This experiment has only used the labeled data. Further, a location labeled in the original data set includes the number of rows and columns. However, as per the zone classification approach, we have re-labeled the ibecon dataset by dividing it into four zones: A, B, C, and D.

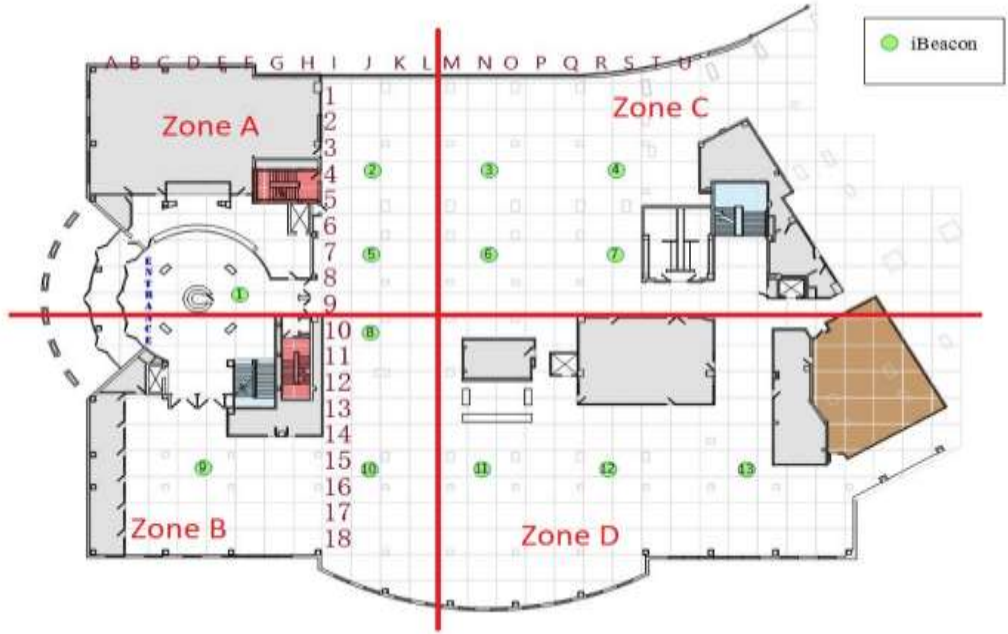

*Figure 38:Experimental Setup [24]*

The RSSI dataset consisted of 1420 data and was filtered using the moving average filters. The Eq.35 shows the mathematical expression for the moving average filter. The moving average (MA) is the most common and widely used filter in digital signal processing because it is the most accessible digital filter to understand and use. Even many works in related works also have used this filter [36-44]. This optimal filter is optimal to avoid random noise while retaining a sharp step response. It obtains N input samples at a time and calculates the average of those points to produce a single output point. When the filter's length increases, the output's smoothness increases, and a moving average filter of length N for an input RSSI signal RSSI(MA) may be defined as follows:

$$\text{RSSI(MA)} = \frac{1}{N}\sum_{k=0}^{N-1} RSSI(n-k) \quad for\ n = 0,1,2,3,\dots \quad (49)$$

The RSSI signals in both time-domain and the frequency-domain are shown in Fig. 42. Where zones A, B, C, and D are as per Fig.41.The Table 9 shows the original dataset.

Table 10 shows the dataset that was modified for classification purposes. Where new four columns are added.

*Table 9: Few data samples from the original dataset*

| location | b3001 | b3002 | b3011 | b3012 | b3013 |
|---|---|---|---|---|---|
| O02 | -200 | -200 | -200 | -200 | -200 |
| P01 | -200 | -200 | -200 | -200 | -200 |
| P01 | -200 | -200 | -200 | -200 | -200 |
| P01 | -200 | -200 | -200 | -200 | -200 |

*Table 10: Few samples of the modified dataset*

| location | b3001 | b3002 | b3011 | b3013 | zone A | Zone B | Zone C | Zone D |
|---|---|---|---|---|---|---|---|---|
| O02 | -200 | -200 | -200 | -200 | 0 | 0 | 1 | 0 |
| P01 | -200 | -200 | -200 | -200 | 0 | 0 | 1 | 0 |
| P01 | -200 | -200 | -200 | -200 | 0 | 0 | 1 | 0 |
| P01 | -200 | -200 | -200 | -200 | 0 | 0 | 1 | 0 |

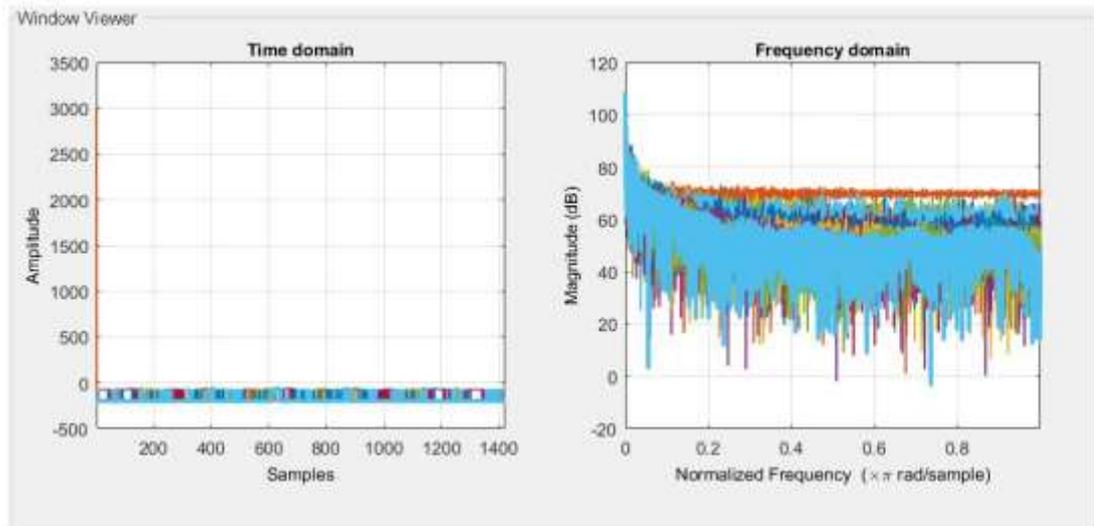

*Figure 39:RSSI data with frequency and time domain*

## 5.4. Model Development and Training

The filtered RSSI data set is trained using FFNN, SVM, and kNN. Where the FFNN consists of two fully connected layers and one output layer. There are thirteen inputs at the input layer as it has thirteen beacons node inputs. The first fully connected layer has 20 neurons, and the second fully connected layer has 17 neurons, as shown in figure 3. The network architecture included one classification layer and four fully connected levels for all simulations. Using the neural network toolkit in MATLAB 2020, the FFNN was implemented. During training, 30% of the dataset was utilized for testing, while 70% was used for training. Models were trained under three different hyper-parameters values of Learning Rate (LR), batch size, and epochs. During the simulations, the ReLU activation function was used in fully connected layers and the SoftMax function in the last layer.

SVM and kNN models were implemented on Jupiter Notebook using Python. All three algorithms evaluated the accuracy, precision, sensitivity, and the F1 score. In SVM, the number of support vectors changed from 25 to 75 as a hyperparameter tuning [12].

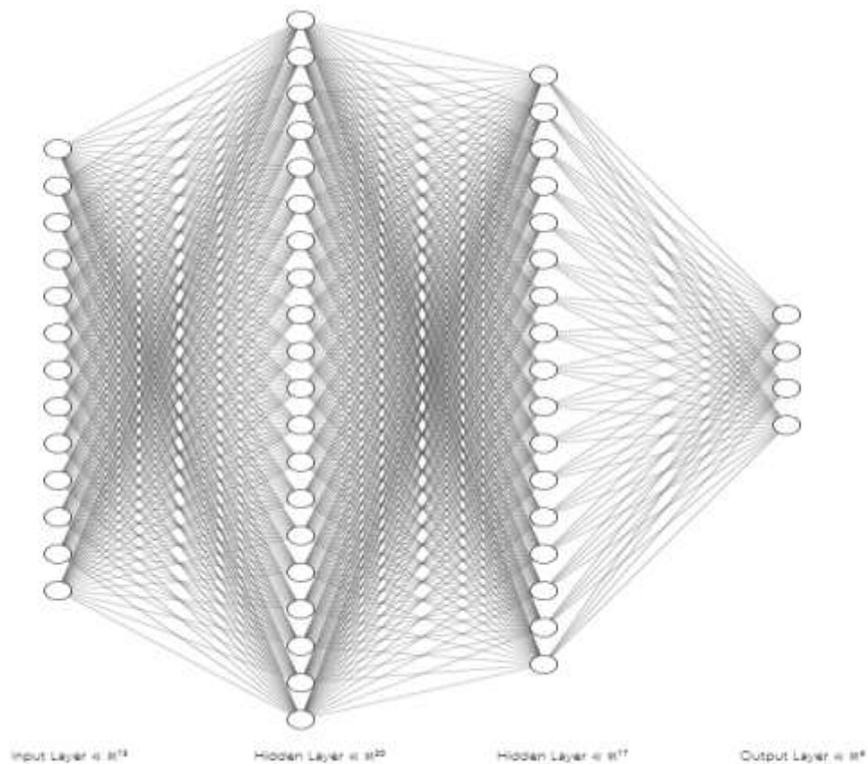

*Figure 40:Network architecture of the FFNN*

## 5.5. Model Evaluation

For FFNN, SVM and kNN, calculated the performance evaluation matrices precision, recall, and the f1-score. The definitions for each matrices are as follows;

True Positive (TP): These successfully predicted positive values imply that both the actual class value and the expected class value are true.

True Negative (TN): These precisely predicted negative values indicate that both the actual class value and the expected class value are zero.

Falese Positives (FP): when the estimated class is true but the actual class is false.

False Negatives (FN): when estiamted class is negative but actual class is positive

And evaluation matrices are defined as follows;

$$Accuracy = \frac{TP + TN}{TP + FP + FN + TN} \quad (50)$$

$$Precision = \frac{TP}{TP + FP} \quad (51)$$

$$Sensitivity = \frac{TP}{TP+FN} \quad (52)$$

$$F1 - Score = \frac{2 \times (Sensitivity \times Precision)}{Sensitivity + Presition} \quad (53)$$

Table 11-13 shows perforamce matrices for each algorithm and the Table 14 shows the summary of the evaluation matrix for all algorithms. There is a significant improvement in the accuracy of SVM when the number of vectors increases. The kNN gives 0.8511 accuracies, and it outperformed.

By adjusting the hyper-parameters, learning rate, batch size, and the number of epochs, the implemented FFNN was put to the test. It was noted that training accuracy is consistently higher than testing accuracy. Three distinct batch sizes—10, 50, and 100—were used to test the training and testing accuracy. Epochs ranged from 0 to 100—figures 3 and 4 display the outcomes.

*Table 11:Evaluation matrices for FFNN*

| Zone | precision | recall | f1-score | support |
|---|---|---|---|---|
| Zone A | 0.71 | 0.76 | 0.73 | 482 |
| Zone B | 0.50 | 0.46 | 0.48 | 69 |
| Zone C | 0.81 | 0.85 | 0.83 | 797 |
| Zone D | 0.00 | 0.00 | 0.00 | 71 |

Table 12:Evaluation matrices for SVM

| | precision | recall | f1-score | support |
|---|---|---|---|---|
| Zone A | 0.75 | 0.83 | 0.79 | 482 |
| Zone B | 0.89 | 0.74 | 0.81 | 69 |
| Zone C | 0.90 | 0.85 | 0.87 | 797 |
| Zone D | 0.88 | 0.90 | 0.89 | 71 |

Table 13:Evaluation matrices for KNN

| | precision | recall | f1-score | support |
|---|---|---|---|---|
| Zone A | 0.78 | 0.81 | 0.79 | 482 |
| Zone B | 0.88 | 0.91 | 0.89 | 69 |
| Zone C | 0.89 | 0.87 | 0.88 | 797 |
| Zone D | 0.96 | 0.96 | 0.96 | 71 |

Table 14:Summery of comparison of all the algorithms

| Algorithm | Sensitivity | Specificity | Precision | Accuracy | F1 Score |
|---|---|---|---|---|---|
| FFNN | 0.8976 | 0.5176 | 0.8344 | 0.7651 | 0.6949 |

| SVM | Vectors=25 | 0.9083 | 0.5176 | 0.8505 | 0.8401 | 0.8479 |
| | Vectors=50 | 0.9083 | 0.5176 | 0.8505 | 0.8408 | 0.8480 |
| | Vectors=75 | 0.9083 | 0.5176 | 0.8505 | 0.8412 | 0.8485 |
| KNN | | 0.9587 | 0.4215 | 0.9062 | 0.8511 | 0.8817 |

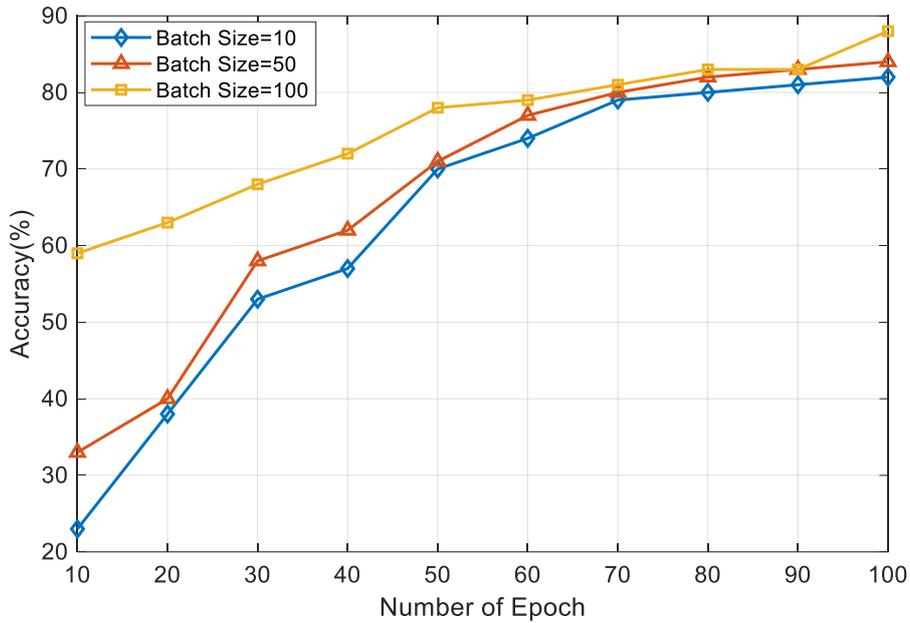

*Figure 41: Training accuracy versus no. of epoch*

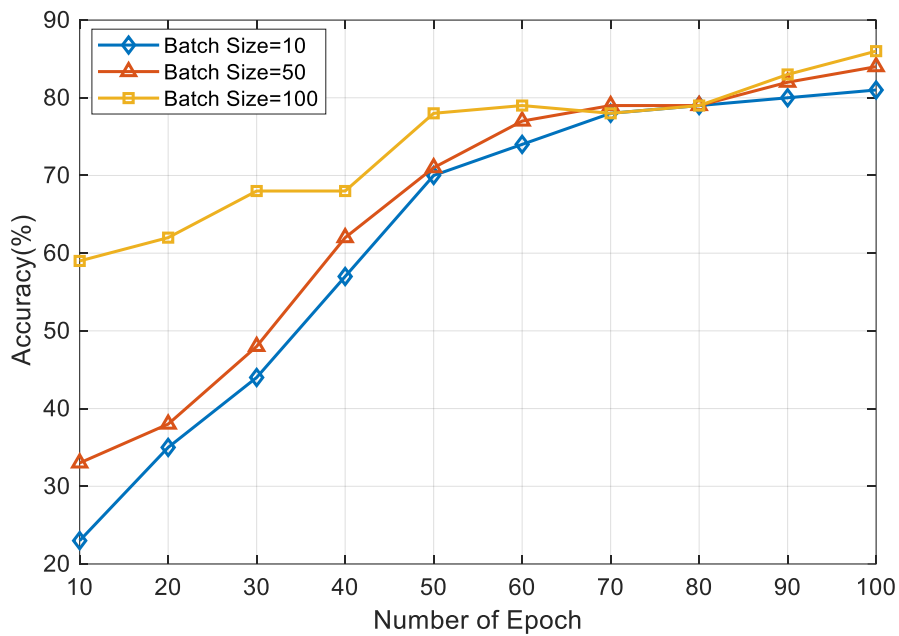

*Figure 42: Testing accuracy versus no. of epoch*

Fig.43 shows the association between the number of samples and the training time for all three algorithms. When the number of samples increases, the training time for all the algorithms increases. Where kNN shows the lowest training time while FFNN gives the highest training time. It was observed that from 200 samples to 1200 samples the training time increase exponentially. And after 1200 samples it has been saturated. When comparing the training time of FFNN it is higher compared to the others. This is due to the computational complexity of the neural network. To overcome this, stochastic gradient descent can be proposed. Stochastic gradient descent works by randomly selecting a small number m of chosen haphazardly training inputs. kNN shows less training time because it doesn't learn a discriminative function from the training data.

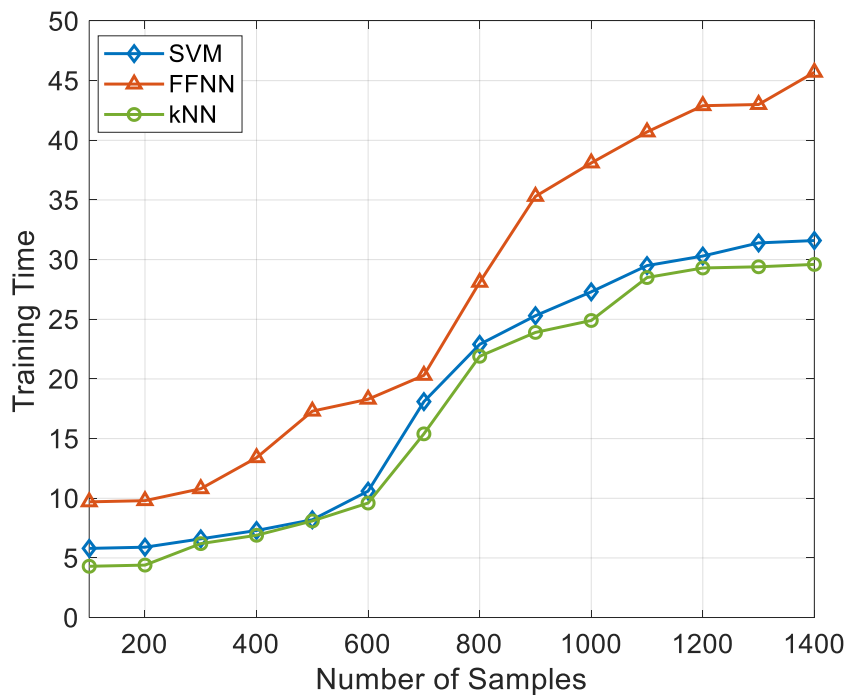

*Figure 43:Tarining time vs various machnine learning algorithms*

Fig.45 shows the relationship among the number of samples versus accuracy. The accuracy rises as the number of samples increases for all the algorithms. In FFNN up to 1000 samples, accuracy is increasing exponentially. And after 1000 samples, accuracy is saturated. However, FFNN provides the lowest accuracy compared to the other two algorithms. The smaller number of data points could be a reason for showing the lowest accuracy in FFNN. SVM and kNN up to 200 samples can see a significant increment of the accuracy. However, after 200 samples less increment. The reduction of model overfitting can be used to explain the accuracy versus the training dataset size curve. In general, as the amount of the training dataset grows, model overfitting decreases.

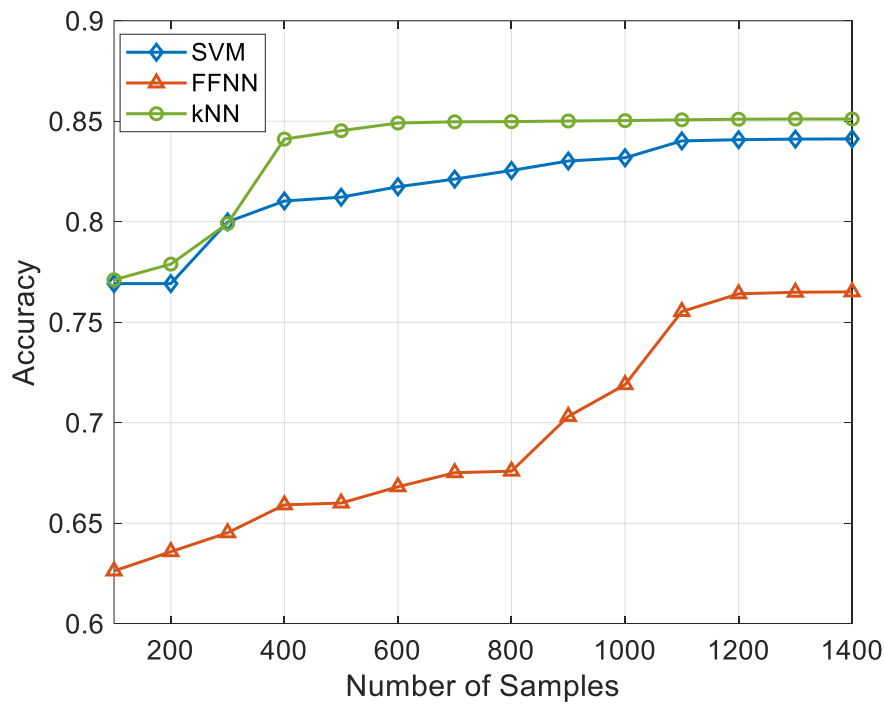

*Figure 44:Accuracy vs various machnine learning algorithms*

Fig.46 shows the relationship among the number of samples and the F1 score. By computing the harmonic mean of a classifier's precision and recall, the F1-score assimilates both into a single metric. It is mainly used to compare the effectiveness of two classifiers. As per results, the F1 score also rapidly upsurges as the number of samples increases. According to the results, kNN gives the highest F1 score, and FNN offers the lowest value. Generally, the accuracy is good for a balanced dataset. However, the utilized dataset in this work is an imbalanced dataset, where we had an unequal number of data samples for four Zones A, B, C, and D. Since the dataset is imbalance, the matrix F1 score is more suitable for evaluating this model.

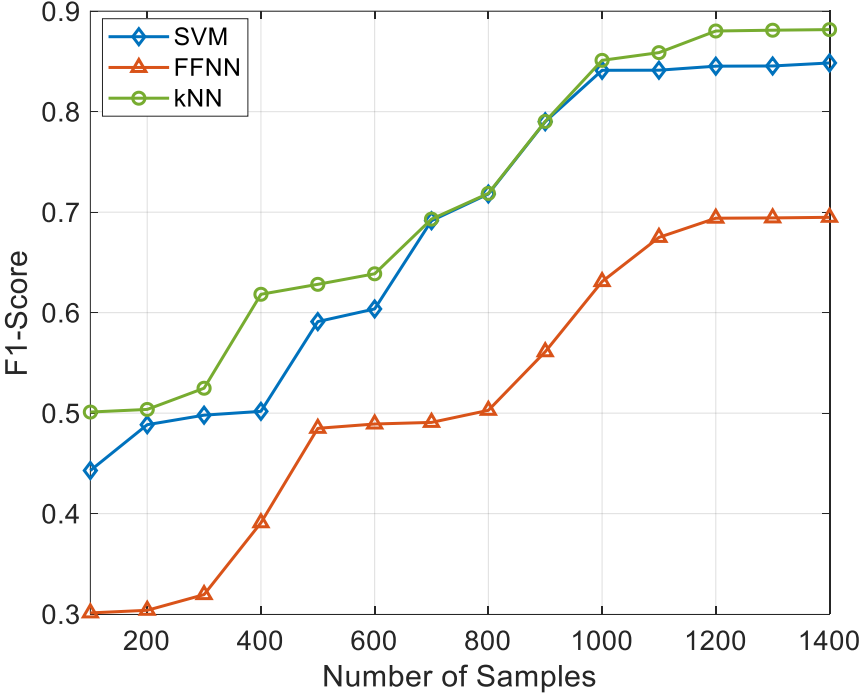

*Figure 45:F1 score vs various machine learning algorithms*

Fig.47 shows the relationship between accuracy, number of samples, and vectors. When the number of samples increases, accuracy increases irrespective of the number of vectors. When the number of vectors increases, accuracy has a significant reduction. We consider linear kernel SVM in this experiment. Also, improving the classifier's performance does not always result from adding more support vectors. The trade-off between accuracy and the number of support vectors that we previously considered true does not always hold in this specific situation. If there are already more than a particular number of support vectors—in this case, about 110—then adding another support vector results in declining marginal returns.

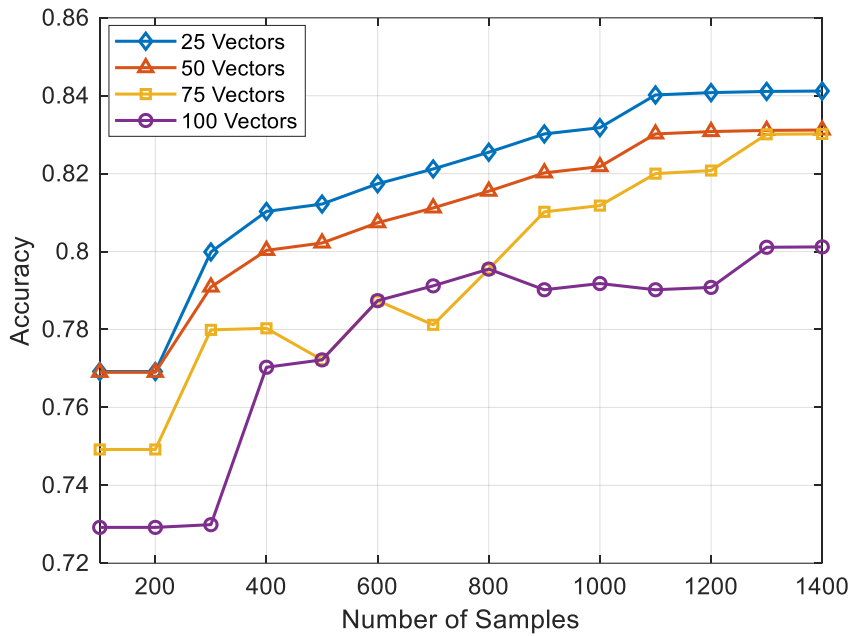

*Figure 46:Accuracy vs Number of Vectors*

Fig.48 shows the effect of the k value in kNN on the accuracy. When k increases from 2-6, it can observe an exponential increment of the accuracy. And from 6 to 20 the accuracy increases. A low number of k indicates that noise will have a more significant impact on the outcome. A considerable value makes it computationally expensive and somewhat contradicts the fundamental tenets of KNN

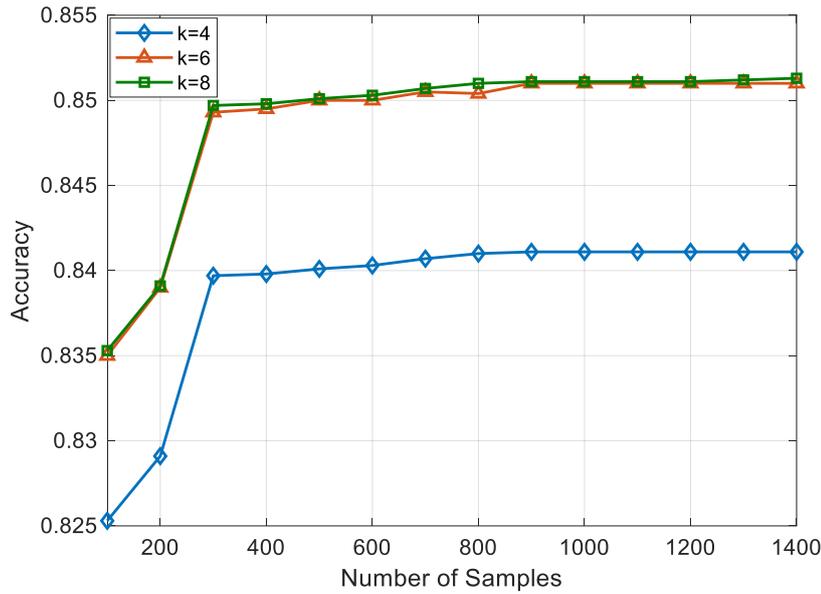

*Figure 47: Accuracy vs k value in kNN*

## 5.6. Summary

This chapter presented a simplified localization method based on BLE and supervised classifiers that could apply to indoor environments. The RSSI values were collected from thirteen different iBeacon nodes installed in an indoor environment utilized in this experiment. The RSSI data were linearized using the weighted least squares method and filtered using moving average filters to remove the outliers. Afterward, RSSI data was trained using SVM, kNN, and FFNN algorithms. During the experiments, the hyperparameters of each algorithm were tuned, and observed the impact on its accuracy. It is observed that there are significant improvements in the accuracy once the sample size is increased. Further, training time is observed against the sample size and number of epochs versus accuracy. To sum up, the kNN offers reasonably good accuracy in classifying the right zone of 85%, FFNN provides 76% when batch size is 100 under the learning rate of 0.01, and SVM provides 84% with 75 support vectors.

# CHAPTER 6:

# *6. An Ensemble Learning-Based Approach for Wireless Indoor Localization*

## 6.1. Introduction

Ensemble learning refers to combining multiple ML algorithms to have more robust predictions. Ensemble learning has several categories: bagging, stacking, and boosting. Bagging entails averaging the predictions from many decision trees fitted to various samples of the same dataset. When numerous distinct model types are fitted to the same data, stacking is used to learn how to combine the predictions effectively. A weighted average of the predictions is produced by boosting, which entails adding ensemble members sequentially that correct the predictions provided by earlier models. This novel algorithm has a bagging approach, where the prediction results from three tree-based algorithms are averaged to perform better.

Further, [19] shows the performance of tree-based supervised algorithms in localization. In this work, it has been modified the experimental testbed used in [9]. In this work, we proposed a novel ensemble learning-based approach to finding the optimal location using a combination of algorithms Decision Tree Regression (DTR), Extra Tree Regressor (ETR), and Random Forest Regressor.

The relationship between the Received Signal Strength Indicator (RSSI) and the distance is the key to any wireless-based ranging and localization system. RSSI-based position systems use three kinds of propagation models free space model, bidirectional surface reflection, and log-normal shadowing (LNSM) [52]. As an improved RSSI-based ranging model, we can use the LNSM, which can use for practical applications[53].

According to [12-18], trilateration techniques have been used widely to estimate the RSSI-based indoor localization. The time of arrival (TOA) [17] and time difference of arrival (TDOA) [14] are primarily time-based systems that need to be define with the transmission time. The angle-based arrival angle (AOA) [15] system needs the highly sophisticated directional antenna in beacon nodes to measure angles. Moreover, techniques such as AoA and TODA need additional hardware to take measurements. Considering the above range-based techniques used in the literature, RSSI is outperformed.

## 6.2. Experimental Setup and Data Collection

This experimental testbed has been set up in an electronic lab with an 8.02 square meter area spanning an open space surrounded by walls. Three reference nodes are placed at the fixed locations and one mobile sensor node. This mobile sensor node considers our target node to predict the site. The mobile sensor node was kept at 32 different locations from time to time and collected corresponding RSSI values received at reference nodes.

This experimental testbed consists of a few reference nodes and a mobile node developed using microcontrollers. The RSSI values at the three reference nodes at a particular position of a mobile node will be transmitted to an IoT cloud via a broadband router The IoT cloud platform integrates the platform with the remote localization model over the internet. Further, the cloud platform is a globally distributed MQTT broker which publishes the information acquired to a remote server. The collected data transfer between the hardware platform and the remote server is accomplished by Wi-Fi and internet technologies, respectively. The ESP8266 has been used to design and develop the mobile and reference nodes in the testbed. ESP8266 is a Wi-Fi module popular for Internet of Things applications [29]. This module has a wireless Wi-Fi transceiver that operates in the IEEE 802.11 b/g/n standard in an unlicensed frequency range of 2400-2484 MHz, supporting the TCP/IP communication protocol stack and Wi-Fi security supporting WAP3.

## 6.3. DATA PRE-PROCESSING

Prior to training the model two aprocaches for RSSI data filtering used. Firstly, proposed algorithm trained with moving averaged filtered data and secondly trained the model with filtered data using Kalman filters.

In previous experiments, we used moving average filters and the moving average filters are giving quite good smoothing to the signal. However, the works in [81-88] shows that Kalman filters are giving high accurate filtering for the signal.

In our approach for localization, we use Kalman filter to reduce RSSI errors. A process model and a measurement model make up the Kalman filter. The relationship between the state of interest and the process sounds is shown in the process model. The relationship between the state of interest and the measurement is shown by the measurement model. The projected RSSI is the state of interest in our method. The RSSI value is used as the measurement. Their connections are shown in chronological order.

The Kalman filter for RSSI estimate is designed as follows.

State of interest x is designed to be RSSI at time step t:

$$x(t) = RSSI(t) \quad (54)$$

The process model of the Kalman filter is designed as:

$$x'(t) = Ax(t) + \mathcal{E} \quad (55)$$

where A=1 in our design, $\mathcal{E}$ is the process noise.

The link between the condition of interest and the received RSSI measurement is used to create the measurement model.

$$z_t = Hx_t + T \quad (56)$$

where $Z_t$ is the RSSI measurement at time step t. H=1 in our design. $\Gamma$ is the measurement noise.

We use a time step from t-1 to t for Kalman filter update, and update the Kalman filter process for state of interest, Kalman gain, and variance from time step t-1 to t as follows.

$$\hat{x} = \hat{x}_{t-1} \quad (57)$$

$$P_t = P_{t-1} + Q \quad (58)$$

$$K_t = P'_t(P'_t + R)^{-1} \quad (59)$$

$$P_t = (1 - K_t) P'_t \quad (60)$$

Where,

$\hat{x}_t$ is the predicted state at time step t.

$\hat{x}_{t-1}$ is the state estimate at time step t-1

Q is the covariance of the process noise.

R is the covariance of measurement noise.

$P'_t$ is the predicted error variance.

$P_t$ is the updated error variance

$K_t$ is the Kalman gain at time step t.

Recursive processing is used by the filter. The stages are repeated using the previous a posteriori estimates to project or anticipate the new a priori estimates after each time and measurement update. Based on all of the prior measurements, the Kalman filter generates a current estimate in a recursive manner.

Both the filters were implemented on MATLAB 2020. The raw RSSI data shows high fluctuations; however, after filtering, those outliers have been removed. The result after using the Kalman filter is shown in Fig.50 and 51.

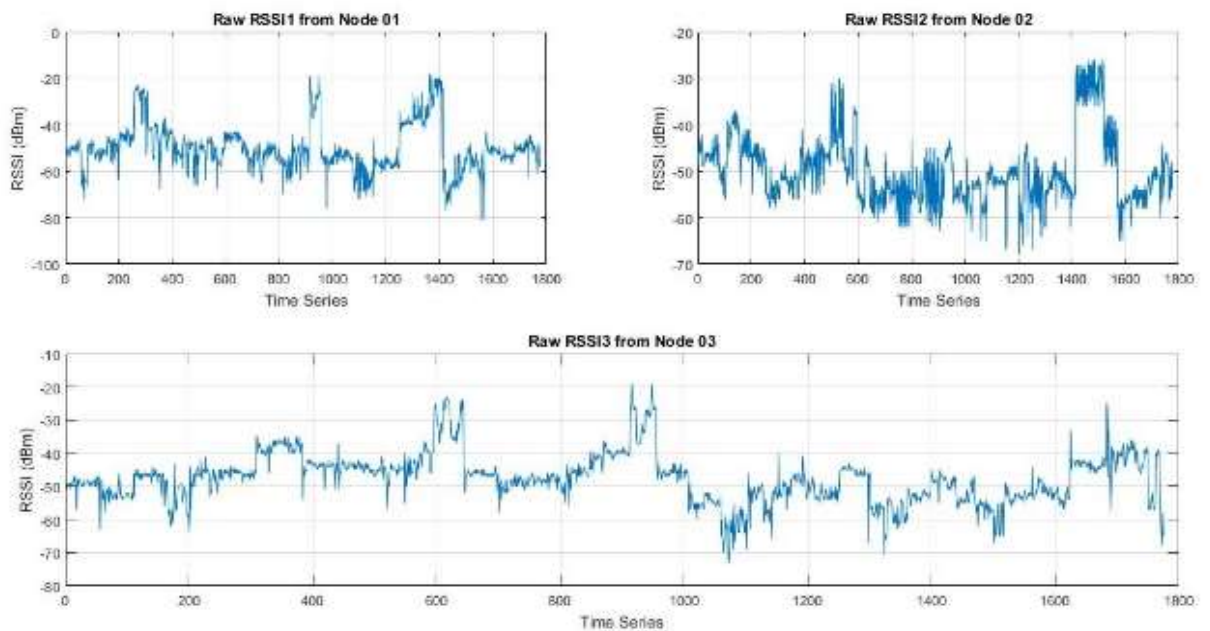

*Figure 48:RSSI data without filtering*

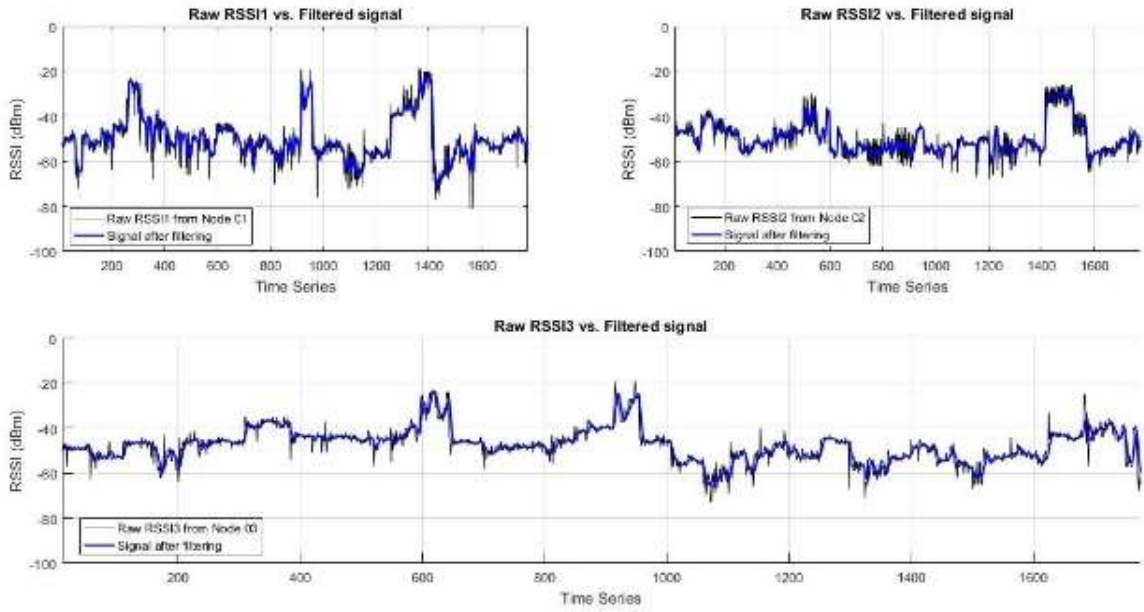

*Figure 49:RSSI data after filtering with Moving filters*

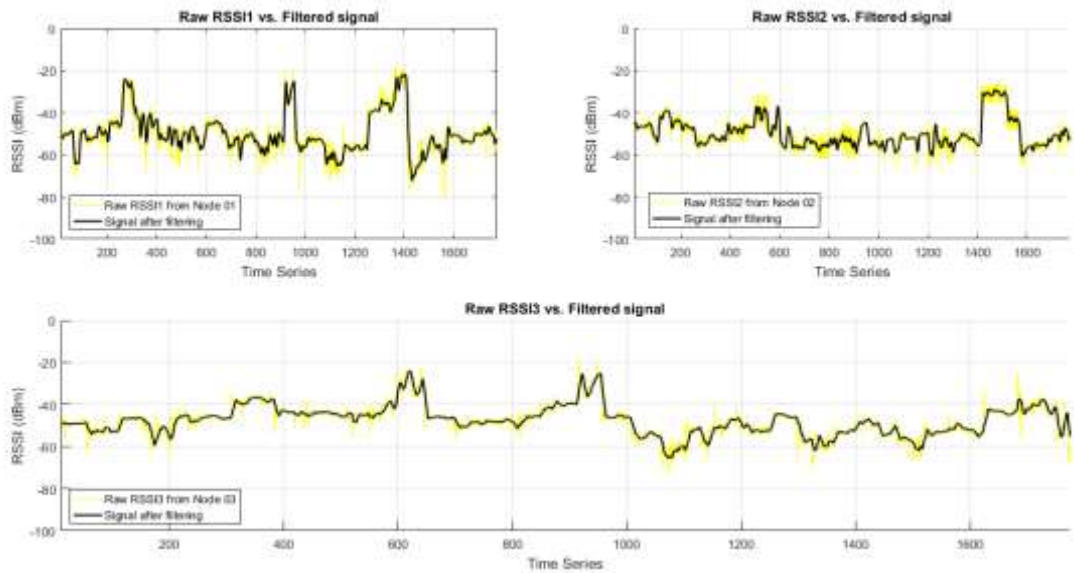

*Figure 50:RSSI data after filtering with Moving filters*

## 6.4. Model Development

Decision Tree Regressor is a structured type classifier with types of nodes. This structure consists of the root nodes and the interior nodes. The root node represents the whole sample in the dataset, and it splits the entire dataset into sub-nodes. The interior nodes represent a dataset's features, and decision rules are marked on the branches. The leaf nodes define the outcomes. There could be a linear, non-linear or complex relationship between labels and the features. The number of trees could significantly affect the prediction results. In our model, we have used 40 trees. [89-92].

Extremely Randomize Tree Regression An extremely randomized trees (or extra trees) regression algorithm [53] is a tree-based ensemble machine learning method. It is a relatively recent algorithm and was developed as an extension of the random forest algorithm. The extra trees regression (ETR) algorithm uses a classical top-down procedure to build an ensemble of unpruned classification/regression trees. ETR uses a random subset of features to train each base estimator, which is the same principle employed by the random forest regression algorithm. However, instead of selecting the most discriminative split in each node, ETR randomly selects the best feature and the corresponding value for splitting the node. Also, random forest uses a bootstrap replica to train the prediction model, whereas ETR uses the whole training dataset to train each regression tree in the forest.[37-40].

An ensemble-type machine learning technique, Random Forest Regressor is made up of several decision trees. Each tree in a random forest will be trained using a subset of the data. To arrive at the final prediction, the predictions from each decision tree will be averaged. Consequently, this approach is more effective than the decision tree technique overall. The few advantages of this algorithm include handling missing values, efficiently handling non-liner paraments, robust outliers of the dataset, and being less impacted by the noises. Our model used 100 forests during the simulations [17-19].

## 6.5. TreeLoc Algorithm

This technique has an approach to the boosting-ensemble method. An ensemble method is a technique that combines the predictions from multiple machine learning algorithms together to make more accurate predictions than any individual model.

Few works exist applying ensemble learning to indoor localization problems [102-112]. Generally, it uses a weighted averaging principle to calculate the final prediction [20]. The working of TreeLoc algorithms is shown in figure 38. The full dataset is split into three equal positions (33.33% each) and fed to the ETR, DTR, and RFR algorithms. Each algorithm's prediction of x and y coordinates is fed into the TreeLoc algorithm. Where TreeLoc will calculate the optimal coordinates again by performing the multiple linear regression. Where coordinates of x and y are generated, each algorithm will use as an independent variable and actual coordinates as the dependent variable to generate the final coordinates. Eq. 61 and 62 denote the prediction of x and y coordinates. Where a1 and a2 are constants and $W_1$, $W_2$, and $W_3$ are the weighted values. Eq.38 and 39 denote the finalized equations with all the constant values and the weights calculated through the simulations.

$$X_{finally\_predicted} = a_1 + W_1 \times x_{ETR} + W_2 \times x_{DTR} + W_3 \times x_{RFR}$$

(61)

$$Y_{finally\_predicted} = a_2 + W_1 \times y_{ETR} + W_2 \times y_{DTR} + W_3 \times y_{RFR}$$

(62)

$W_1$, $W_2$, and $W_3$ are the corresponding weighted values for ETR, RFR, and DTR, respectively.

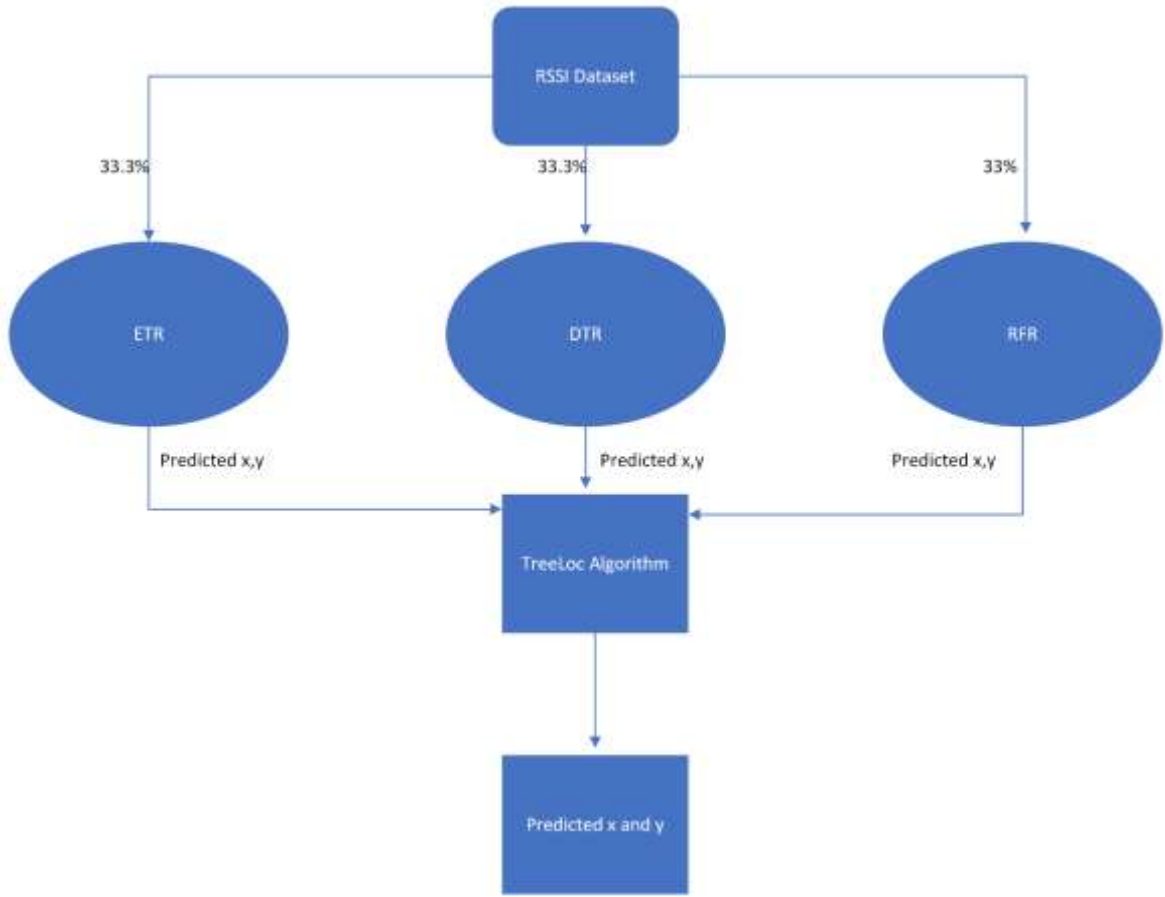

*Figure 51:zTreeLoc algorithm*

## 6.6. Model Training and Evaluation

Initially, ETR, RFR, and DTR algorithms were computed to observe the performance in location predictions. One thousand seven hundred seventy-six RSSI data was received from three reference nodes used to train the algorithms. The proposed model in above section is developed on Jupiter Notebook in a python3 environment. Computer Root Mean Squared (RMSE) and $R^2$ values are shown as per table 15 and 16.

*Table 15: Comparison of RMSE and R2 of algorithms with Moving Average Filters*

| Algorithm | RMSE | | $R^2$ | |
|---|---|---|---|---|
| | x coordinate(cm) | y coordinate(cm) | x coordinate | y coordinate |
| ETR | 8.94 | 9.48 | 0.9923 | 0.9184 |
| RFR | 29.14 | 29.09 | 0.8822 | 0.9046 |
| DTR | 29.33 | 28.52 | 0.8952 | 0.0834 |
| TreeLoc | 8.79 | 8.83 | 0.9249 | 0.9244 |

*Table 16: Comparison of RMSE and $R^2$ of algorithms with Kalman Filters*

| Algorithm | RMSE | | $R^2$ | |
|---|---|---|---|---|
| | x coordinate(cm) | y coordinate(cm) | x coordinate(cm) | y coordinate |
| ETR | 8.64 | 9.30 | 0.989 | 0.9173 |
| RFR | 29.02 | 29.00 | 0.8810 | 0.9012 |
| DTR | 23.33 | 23.52 | 0.8452 | 0.0834 |
| TreeLoc | 8.22 | 8.39 | 0.9249 | 0.9244 |

When considering the prediction results of the above three algorithms, all the algorithms performed very well in terms of accuracy. All the algorithm shows the error, which is less than 30cm. However, when observing the prediction results of all x and y coordinates in individual algorithms mentioned above, each algorithm gives the best prediction in an ad hoc way. Therefore, TreeLoc is introduced to optimize this by taking a weighted average and re-computing the coordinates by performing a multiple linear regression. As per this modeling, obtained models for final x and y are presented in Eq. 63 and 64.

$$X_{finally\_predicted} = -0.9494 + 0.8036 \times x_{ETR} + 0.5476 \times x_{DTR} + 0.5212 \times x_{RFR} \quad (63)$$

$$Y_{finally\_predicted} = -0.8348 + 0.8922 \times y_{ETR} + 0.5937 \times y_{DTR} + 0.5292 \times y_{RFR} \quad (64)$$

## 6.7. Summary

In this chapter, we proposed a novel ensemble-learning-based technique for the indoor localization problem. Popular tree-based algorithms, namely Decision Tree Regression (DTR), Extra Tree Regressor (ETR), and Random Forest Regressor (RFR), are used to build a novel TreeLoc algorithm. Though ETR is outperformed during individual simulations, RFR and DTR also give the best predictions for some samples in an ad-hoc way when observing the prediction results. In the TreeLoc algorithm, we introduced a weighted average mechanism to optimize a location. Simulation results show that the TreeLoc algorithm could provide an 8.79cm error for x coordinates and an 8.83cm error for its y coordinates when using moveing average filters and 8.22cm error for x coordinates and an 8.39cm error for its y coordinates for Kalman filters.

# Chapter 7:

# 7. *Conclusions and Future Works.*

The most significant advancements in indoor locating systems have occurred in recent years. As a result, many location-based IoT applications are developing in the market. Researchers and industry are presently investigating, developing, and improving solid indoor positioning systems.

Location-based services are considered a primary application in the Internet of Things (IoT). Over the past several decades, localization technologies have been created to offer users location and navigation services using technological advancements in digital circuitry. The Global Positioning System (GPS) was one of the first positioning systems for outdoor environments. This system is a location-based navigation system that is suitable for outdoor localization applications. GPS-based systems require specific and expensive hardware, but the smartphone era has made it feasible to utilize GPS on our handheld devices without any extra hardware. For the outdoor location, GPS has become the de facto norm. However, owing to the lack of Line-of-Sight (LOS) within buildings, GPS cannot be used in indoor environments. Localization-based systems for indoor environments are being developed since humans spend more time inside than outside. Indoor positioning systems have been developed using various signal technologies such as WiFi, ZigBee, Bluetooth, UWB..etc., depending on the context and application situation.

This thesis explores the applicability and suitability of developing an indoor positioning system using machine learning techniques. Range-based localization approaches and algorithms were chosen to explore, test, evaluate and compare in testbeds developed for this purpose. The experimental testbed was designed using Wi-Fi and existing testbeds for the wireless technologies, Zigbee, Bluetooth Low Energy (BLE), and LoRaWAN. In each testbed, RSSI values were collected using IoT cloud architectures. The collected data were pre-processed, investigating appropriate filters before training the algorithm.

The supervised regressors were investigated under every wireless technology mentioned above to estimate the geographical coordinates of a moving object sensor node and compare the performances of each supervised regressor using model evaluation matrices. Experiments were conducted using the RSSI data obtained from the Wi-Fi testbed showed that Decision Tree Regressor (DTR) was the best-outperformed algorithm compared to the rest of the tested algorithms. It was observed that once the number of reference nodes increased in the test bed, the accuracy and error were significantly improved. The number of forests in DTR matters to improve the location estimation accuracy and its significance in reducing the error. We foresee using supervised ML algorithms to improve results rather than deterministic localization based on our experiments.

Further, RSSI data collected from BLE, LoRaWAN, and Zigbee testbeds were trained using supervised learning techniques. When comparing the localization accuracy, all algorithms tested in this experiment give fairly good error values of less than one meter. When comparing the technologies, BLE outperformed based on the results, achieving the lowest error among all the supervised algorithms. One algorithm could not propose the best because different algorithms outperformed each technology. Moreover, BLE considers the most inadequate power-consuming technology. This experiment has only considered the 2D environments. A study on localization for a 3D environment would be an interesting future research direction.

Secondly, supervised classifiers FFNN, kNN, and SVM investigate indoor localization problems. Using kNN, we could achieve pretty good location classification accuracy. The RSSI values received from thirteen different iBeacon nodes were trained under a four-layered FFNN. The hyperparameters were tuned and observed the change of accuracy in each condition. The prediction model, kNN, provides reasonably good accuracy in classifying the correct zone of 85% when batch size is 100 under the learning rate of 0.01. at the same time, SVM and FFNN gave an accuracy of 84% and 76%, respectively.

Thirdly, a hybrid algorithm was developed, namely TreeLoc(Tree-based localization), as an improved localization algorithm and evaluated the performances against existing algorithms. Popular tree-based algorithms, namely Decision Tree Regression (DTR), Extra Tree Regressor (ETR), and Random Forest Regressor (RFR), are used to build a novel TreeLoc algorithm. In the TreeLoc algorithm, we introduced a weighted average mechanism to optimize a location. Simulation results show that the TreeLoc algorithm could provide an 8.79cm error for x coordinates and an 8.83cm error for its y coordinates when using moveing average filters and 8.22cm error for x coordinates and an 8.39cm error for its y coordinates for Kalman filters. Though ETR is outperformed during individual simulations, RFR and DTR also give the best predictions for some samples in an ad-hoc way when observing the prediction results.

# *8.* REFERENCES

**Annexures A: Codes for testbed implementation**

**C code for ESP 12E - Mobile Node**

```c
#include <Arduino.h>

#include <ESP8266WiFi.h>

#include <PubSubClient.h>

//broker information ; broker CloudMQTT

const char* mqtt_server ="m11.cloudmqtt.com";

const char* topic = "wifips";

int port=19585;

char message_buffer[250];

// Connect to the publishing WiFi acces point

const char* ssid = "WEERASINGHE";

const char* password = "123456789o";

void callback(char* topic, byte* payload, unsigned int length) {

  // handle message arrived

}

WiFiClient espClient;

PubSubClient client(mqtt_server,port,callback,espClient);

void setup() {

  Serial.begin(9600);

  pinMode(LED_BUILTIN,OUTPUT);

  Serial.println("hello from pointnode");

  WiFi.begin(ssid,password);

  while(WiFi.status() != WL_CONNECTED){

    Serial.print(".");

    delay(1);
```

```
  }
  Serial.println("connected!!!");
}
void loop() {
  //digitalWrite(LED_BUILTIN,HIGH);
  //Serial.print("Scanning:");
  String rssiinfo,ssidinfo;
  byte available_networks = WiFi.scanNetworks();
  String mssg="";
  for (int network = 0; network < available_networks; network++){
        ssidinfo= WiFi.SSID(network);
        rssiinfo= String(WiFi.RSSI(network));
        if(ssidinfo=="reffnode1"){
            mssg+="1:"+rssiinfo;
        }
        if(ssidinfo=="reffnode2"){
            mssg+="2:"+rssiinfo;
        }
        if(ssidinfo=="reffnode3"){
            mssg+="3:"+rssiinfo;
        }
        if(ssidinfo=="reffnode4"){
            mssg+="4:"+rssiinfo;
        }
      mssg=mssg+"#";
```

```
        delay(100);
   }
   String pubString = mssg;
   if(WiFi.status() == WL_CONNECTED){
      Serial.println(pubString);
      pubString.toCharArray(message_buffer, pubString.length()+1);
      if (client.connect("wifips","ucikymgc","zSMn6XvdKD-8")) {
         //Serial.println("Connected to MQTT server");
         //client.set_callback(callback);
         client.publish("wifips",message_buffer);
      }    }    //digitalWrite(LED_BUILTIN,LOW);    delay(100);
}
```

**C code for ESP 01 – Reference Node**

```
#include <ESP8266WiFi.h>
void setup()
{
  Serial.begin(115200);
  Serial.println();
  Serial.print("reffnode2");
  Serial.println(WiFi.softAP("reffnode2", "reffnode2") ? "Ready" : "Failed!");
}
void loop()
{
  Serial.printf("Stations connected = %d\n", WiFi.softAPgetStationNum());
  delay(1000);
```

}

**Matlab code for filter RSSI**

```matlab
load('L#23.mat');
 X=A(:,[1]);
 Y=A(:,[2]);
 Z=A(:,[3]);
reff1= smooth(X,0.1,'rloess');
reff2= smooth(Y,0.2,'rloess');
reff3= smooth(Z,0.1,'rloess');
% reff1= X;
% reff2= Y;
% reff3= Z;
subplot(3,1,1);
hist(reff1);
subplot(3,1,2);
hist(reff2);
subplot(3,1,3)
hist(reff3);
temp=[reff1,reff2,reff3];
save('point_23','temp');
```

**Annexure B: codes for ML model implementation for Supervised Regressors**

```python
import pandas as pd
from sklearn.model_selection import train_test_split
from sklearn.tree import DecisionTreeRegressor
from sklearn.metrics import mean_squared_error, r2_score
data_frame = pd.read_csv("Data.csv")
```

```python
input_data = data_frame[['RSSI1', 'RSSI2', 'RSSI3']]
label_x = data_frame[['X_Actual']]
label_y = data_frame[['Y_Actual']]

# Split data into train and test to verify accuracy after fitting the model.
input_x_train, input_x_test, label_x_train, label_x_test = train_test_split(input_data, label_x, test_size=0.2, random_state=5)
input_y_train, input_y_test, label_y_train, label_y_test = train_test_split(input_data, label_y, test_size=0.2, random_state=5)

# DTR model
DTR_x_model = DecisionTreeRegressor(max_depth=25)
DTR_y_model = DecisionTreeRegressor(max_depth=25)
# Training
DTR_x_model.fit(input_x_train, label_x_train)
DTR_y_model.fit(input_y_train, label_y_train)
# Prediction
predict_x_train = DTR_x_model.predict(input_x_train)
predict_x_test = DTR_x_model.predict(input_x_test)

predict_y_train = DTR_y_model.predict(input_y_train)
predict_y_test = DTR_y_model.predict(input_y_test)

# Training and testing accuracies
print('Training MSE X', mean_squared_error(label_x_train, predict_x_train))
print('Testing MSE X', mean_squared_error(label_x_test, predict_x_test))
print('Training MSE Y', mean_squared_error(label_y_train, predict_y_train))
print('Testing MSE Y', mean_squared_error(label_y_test, predict_y_test))
# Dataset accuracy
x_prediction = DTR_x_model.predict(input_data)
y_prediction = DTR_y_model.predict(input_data)
```

```python
print('----------------------------------------------------------------------------------------')
print('X MSE: ', mean_squared_error(label_x, x_prediction))
print('Y MSE: ', mean_squared_error(label_y, y_prediction))

print('X r2 ', r2_score(label_x, x_prediction))
print('Y r2 ', r2_score(label_y, y_prediction))

results = {'DTR_prediction_X':list(x_prediction), 'DTR_prediction_Y': list(y_prediction)}
PR_results_df = pd.DataFrame(results, columns=['DTR_prediction_X', 'DTR_prediction_Y'])

PR_results_df.to_csv('DTR_results.csv')
import pandas as pd
from sklearn.model_selection import train_test_split
from sklearn import linear_model
from sklearn.metrics import mean_squared_error, r2_score

data_frame = pd.read_csv("Data.csv")

input_data = data_frame[['RSSI1', 'RSSI2', 'RSSI3']]
label_x = data_frame[['X_Actual']]
label_y = data_frame[['Y_Actual']]

# Split data into train and test to verify accuracy after fitting the model.
input_x_train, input_x_test, label_x_train, label_x_test = train_test_split(input_data, label_x, test_size=0.3, random_state=5)
input_y_train, input_y_test, label_y_train, label_y_test = train_test_split(input_data, label_y, test_size=0.3, random_state=5)

# Build the linear model
LR_x_model = linear_model.LinearRegression()
```

```python
LR_y_model = linear_model.LinearRegression()

# Training
LR_x_model.fit(input_x_train, label_x_train)
LR_y_model.fit(input_y_train, label_y_train)

# Prediction
predict_x_train = LR_x_model.predict(input_x_train)
predict_x_test = LR_x_model.predict(input_x_test)

predict_y_train = LR_y_model.predict(input_y_train)
predict_y_test = LR_y_model.predict(input_y_test)

# Training and testing accuraciss
print('Training MSE X', mean_squared_error(label_x_train, predict_x_train))
print('Testing MSE X', mean_squared_error(label_x_test, predict_x_test))

print('Training MSE Y', mean_squared_error(label_y_train, predict_y_train))
print('Testing MSE Y', mean_squared_error(label_y_test, predict_y_test))

# Dataset accuracy
x_prediction = LR_x_model.predict(input_data)
y_prediction = LR_y_model.predict(input_data)

print('-----------------------------------------------------------------------------------------')
print('X MSE: ', mean_squared_error(label_x, x_prediction))
print('Y MSE: ', mean_squared_error(label_y, y_prediction))

print('X r2 ', r2_score(label_x, x_prediction))
print('Y r2 ', r2_score(label_y, y_prediction))
```

```python
results = {'LR_prediction_X':list(x_prediction), 'LR_prediction_Y': list(y_prediction)}
LR_results_df = pd.DataFrame(results, columns=['LR_prediction_X', 'LR_prediction_Y'])

LR_results_df.to_csv('LR_results.csv')
import pandas as pd
from sklearn.model_selection import train_test_split
from sklearn import linear_model
from sklearn.preprocessing import PolynomialFeatures
from sklearn.metrics import mean_squared_error, r2_score

data_frame = pd.read_csv("Data.csv")
input_data = data_frame[['RSSI1', 'RSSI2', 'RSSI3']]
label_x = data_frame[['X_Actual']]
label_y = data_frame[['Y_Actual']]

# Split data into train and test to verify accuracy after fitting the model.
input_x_train, input_x_test, label_x_train, label_x_test = train_test_split(input_data, label_x, test_size=0.3, random_state=5)
input_y_train, input_y_test, label_y_train, label_y_test = train_test_split(input_data, label_y, test_size=0.3, random_state=5)

# Build PR model
poly_reg = PolynomialFeatures(degree=4)
LR_x_model = linear_model.LinearRegression()
LR_y_model = linear_model.LinearRegression()
# Training
LR_x_model.fit(poly_reg.fit_transform(input_x_train), label_x_train)
LR_y_model.fit(poly_reg.fit_transform(input_y_train), label_y_train)

# Prediction
predict_x_train = LR_x_model.predict(poly_reg.fit_transform(input_x_train))
```

```python
predict_x_test = LR_x_model.predict(poly_reg.fit_transform(input_x_test))

predict_y_train = LR_y_model.predict(poly_reg.fit_transform(input_y_train))
predict_y_test = LR_y_model.predict(poly_reg.fit_transform(input_y_test))

# Training and testing accuraciss
print('Training MSE X', mean_squared_error(label_x_train, predict_x_train))
print('Testing MSE X', mean_squared_error(label_x_test, predict_x_test))

print('Training MSE Y', mean_squared_error(label_y_train, predict_y_train))
print('Testing MSE Y', mean_squared_error(label_y_test, predict_y_test))

# Dataset accuracy
x_prediction = LR_x_model.predict(poly_reg.fit_transform(input_data))
y_prediction = LR_y_model.predict(poly_reg.fit_transform(input_data))

print('-------------------------------------------------------------------------------------')
print('X MSE: ', mean_squared_error(label_x, x_prediction))
print('Y MSE: ', mean_squared_error(label_y, y_prediction))

print('X r2 ', r2_score(label_x, x_prediction))
print('Y r2 ', r2_score(label_y, y_prediction))

results = {'PR_prediction_X':list(x_prediction), 'PR_prediction_Y': list(y_prediction)}
PR_results_df = pd.DataFrame(results, columns=['PR_prediction_X', 'PR_prediction_Y'])

PR_results_df.to_csv('PR_results.csv')
import numpy as np
import pandas as pd
```

```python
from sklearn.model_selection import train_test_split
from sklearn.ensemble import RandomForestRegressor
from sklearn.metrics import r2_score,mean_squared_error

data_frame = pd.read_csv("Data.csv")

input_data = data_frame[['RSSI1', 'RSSI2', 'RSSI3']]
label_x = data_frame[['X_Actual']]
label_y = data_frame[['Y_Actual']]

label_x = label_x.values.ravel()
label_y = label_y.values.ravel()

# Split data into train and test to verify accuracy after fitting the model.
input_x_train, input_x_test, label_x_train, label_x_test = train_test_split(input_data, label_x, test_size=0.2, random_state=40)
input_y_train, input_y_test, label_y_train, label_y_test = train_test_split(input_data, label_y, test_size=0.2, random_state=40)
# Random Forest Regressor model
RFR_x_model = RandomForestRegressor(n_estimators = 100, random_state = 42)
RFR_y_model = RandomForestRegressor(n_estimators = 100, random_state = 42)

# Training
RFR_x_model.fit(input_x_train, label_x_train)
RFR_y_model.fit(input_y_train, label_y_train)

# Prediction
predict_x_train = RFR_x_model.predict(input_x_train)
predict_x_test = RFR_x_model.predict(input_x_test)

predict_y_train = RFR_y_model.predict(input_y_train)
```

```python
predict_y_test = RFR_y_model.predict(input_y_test)

# Training and testing accuraciss
print('Training MSE X', mean_squared_error(label_x_train, predict_x_train))
print('Testing MSE X', mean_squared_error(label_x_test, predict_x_test))

print('Training MSE Y', mean_squared_error(label_y_train, predict_y_train))
print('Testing MSE Y', mean_squared_error(label_y_test, predict_y_test))
# Dataset accuracy
x_prediction = RFR_x_model.predict(input_data)
y_prediction = RFR_y_model.predict(input_data)

print('-------------------------------------------------------------------------------------')
print('X MSE: ', mean_squared_error(label_x, x_prediction))
print('Y MSE: ', mean_squared_error(label_y, y_prediction))

print('X r2 ', r2_score(label_x, x_prediction))
print('Y r2 ', r2_score(label_y, y_prediction))

results = {'RFR_prediction_X':list(x_prediction), 'RFR_prediction_Y': list(y_prediction)}
RFR_results_df = pd.DataFrame(results, columns=['RFR_prediction_X', 'RFR_prediction_Y'])

RFR_results_df.to_csv('RFR_results.csv')
```

```
{
 "cells": [
  {
   "cell_type": "code",
   "execution_count": 1,
   "metadata": {},
   "outputs": [
    {
```

```
     "name": "stdout",
     "output_type": "stream",
     "text": [
      "Training MSE X 5489.0241819491985\n",
      "Testing MSE X 4963.460143192808\n",
      "Training MSE Y 4414.898851726243\n",
      "Testing MSE Y 4683.982657829282\n",
      "----------------------------------------------------------------------------\n",
      "X MSE:  5383.911374197921\n",
      "Y MSE:  4468.715612946851\n",
      "X r2   0.34642241613084634\n",
      "Y r2   0.49634156311378297\n"
     ]
    }
   ],
   "source": [
    "import numpy as np\n",
    "import pandas as pd\n",
    "from sklearn.model_selection import train_test_split\n",
    "from sklearn.svm import SVR\n",
    "from sklearn.metrics import r2_score,mean_squared_error\n",
    "\n",
    "data_frame = pd.read_csv(\"Data.csv\") \n",
    "\n",
    "\n",
    "input_data = data_frame[['RSSI1', 'RSSI2', 'RSSI3']]\n",
    "label_x = data_frame[['X_Actual']]\n",
    "label_y = data_frame[['Y_Actual']]\n",
    "\n",
    "label_x = label_x.values.ravel()\n",
    "label_y = label_y.values.ravel()\n",
    "\n",
```

```
    "# Split data into train and test to verify accuracy after fitting the model. \n",
    "input_x_train, input_x_test, label_x_train, label_x_test = train_test_split(input_data, label_x, test_size=0.2, random_state=40)\n",
    "input_y_train, input_y_test, label_y_train, label_y_test = train_test_split(input_data, label_y, test_size=0.2, random_state=40)\n",
    "\n",
    "\n",
    "# SVM model\n",
    "SVM_x_model = SVR()\n",
    "SVM_y_model = SVR()\n",
    "\n",
    "# Training\n",
    "SVM_x_model.fit(input_x_train, label_x_train)\n",
    "SVM_y_model.fit(input_y_train, label_y_train)\n",
    "\n",
    "# Prediction\n",
    "predict_x_train = SVM_x_model.predict(input_x_train)\n",
    "predict_x_test = SVM_x_model.predict(input_x_test)\n",
    "\n",
    "predict_y_train = SVM_y_model.predict(input_y_train)\n",
    "predict_y_test = SVM_y_model.predict(input_y_test)\n",
    "\n",
    "# Training and testing accuraciss\n",
    "print('Training MSE X', mean_squared_error(label_x_train, predict_x_train))\n",
    "print('Testing MSE X', mean_squared_error(label_x_test, predict_x_test))\n",
    "\n",
    "print('Training MSE Y', mean_squared_error(label_y_train, predict_y_train))\n",
    "print('Testing MSE Y', mean_squared_error(label_y_test, predict_y_test))\n",
    "\n",
```

```
      "\n",
      "# Dataset accuracy\n",
      "x_prediction = SVM_x_model.predict(input_data)\n",
      "y_prediction = SVM_y_model.predict(input_data)\n",
      "\n",
      "print('----------------------------------------------------------------------------')\n",
      "print('X MSE: ', mean_squared_error(label_x, x_prediction))\n",
      "print('Y MSE: ', mean_squared_error(label_y, y_prediction))\n",
      "\n",
      "print('X r2 ', r2_score(label_x, x_prediction))\n",
      "print('Y r2 ', r2_score(label_y, y_prediction))\n",
      "\n",
      "results = {'SVM_prediction_X':list(x_prediction), 'SVM_prediction_Y': list(y_prediction)}\n",
      "PR_results_df = pd.DataFrame(results, columns=['SVM_prediction_X', 'SVM_prediction_Y'])\n",
      "\n",
      "PR_results_df.to_csv('SVM_results.csv')"
     ]
   },
   {
    "cell_type": "code",
    "execution_count": null,
    "metadata": {},
    "outputs": [],
    "source": []
   }
  ],
  "metadata": {
   "kernelspec": {
    "display_name": "Python 3",
    "language": "python",
    "name": "python3"
```

```
    },
    "language_info": {
     "codemirror_mode": {
      "name": "ipython",
      "version": 3
     },
     "file_extension": ".py",
     "mimetype": "text/x-python",
     "name": "python",
     "nbconvert_exporter": "python",
     "pygments_lexer": "ipython3",
     "version": "3.8.5"
    }
   },
   "nbformat": 4,
   "nbformat_minor": 4
}
```

Annexure C: Codes for ANN implementation

```
clear all
close all

epoach=0:10:100;
accuracy_bs_10=[23 74.12 79.29 82.23 84.23 84.22 85.23 86.34 86.22 86.38 86.19 ];
plot(epoach,accuracy_bs_10)
title('Training Accuracy verses Epoch with Learning Rate of 0.01 ');
xlabel('Epoch');
ylabel('Acuracy(100%)')
hold on
accuracy_bs_50=[44.74 75.89 83.39 84.87 84.76 85.10 85.17 85.05 85.98 85.04 85.70];
plot(epoach,accuracy_bs_50);
```

hold on

accuracy_bs_100=[62.75 71.72 78.93 79.40 81.22 81.05 82.51 85.86 85.16 86.46 85.75];

plot(epoach,accuracy_bs_100);

hold off

grid on

legend ('Batch Size=10','Batch Size=50','Batch Size=100')

Annexure D: Codes for ensemble-learning model implementation

```python
import pandas as pd
from sklearn.model_selection import train_test_split
from sklearn.tree import DecisionTreeRegressor
from sklearn.metrics import mean_squared_error, r2_score

data_frame = pd.read_csv("Data1.csv")

input_data = data_frame[['RSSI1', 'RSSI2', 'RSSI3']]
label_x = data_frame[['X_Actual']]
label_y = data_frame[['Y_Actual']]

# Split data into train and test to verify accuracy after fitting the model.
input_x_train, input_x_test, label_x_train, label_x_test = train_test_split(input_data, label_x, test_size=0.2, random_state=5)
input_y_train, input_y_test, label_y_train, label_y_test = train_test_split(input_data, label_y, test_size=0.2, random_state=5)

# DTR model
DTR_x_model = DecisionTreeRegressor(max_depth=25)
DTR_y_model = DecisionTreeRegressor(max_depth=25)
```

```python
# Training
DTR_x_model.fit(input_x_train, label_x_train)
DTR_y_model.fit(input_y_train, label_y_train)

# Prediction
predict_x_train = DTR_x_model.predict(input_x_train)
predict_x_test = DTR_x_model.predict(input_x_test)

predict_y_train = DTR_y_model.predict(input_y_train)
predict_y_test = DTR_y_model.predict(input_y_test)

# Training and testing accuraciss
print('Training MSE X', mean_squared_error(label_x_train, predict_x_train))
print('Testing MSE X', mean_squared_error(label_x_test, predict_x_test))

print('Training MSE Y', mean_squared_error(label_y_train, predict_y_train))
print('Testing MSE Y', mean_squared_error(label_y_test, predict_y_test))

# Dataset accuracy
x_prediction = DTR_x_model.predict(input_data)
y_prediction = DTR_y_model.predict(input_data)

print('-----------------------------------------------------------------------------------')
print('X MSE: ', mean_squared_error(label_x, x_prediction))
print('Y MSE: ', mean_squared_error(label_y, y_prediction))

print('X r2 ', r2_score(label_x, x_prediction))
print('Y r2 ', r2_score(label_y, y_prediction))
```

```python
results = {'DTR_prediction_X':list(x_prediction), 'DTR_prediction_Y': list(y_prediction)}
PR_results_df = pd.DataFrame(results, columns=['DTR_prediction_X', 'DTR_prediction_Y'])

PR_results_df.to_csv('DTR_results.csv')

import pandas as pd
from sklearn.model_selection import train_test_split
from sklearn.ensemble import ExtraTreesRegressor
from sklearn.metrics import mean_squared_error, r2_score

data_frame = pd.read_csv("Data2.csv")

input_data = data_frame[['RSSI1', 'RSSI2', 'RSSI3']]
label_x = data_frame[['X_Actual']]
label_y = data_frame[['Y_Actual']]

# Split data into train and test to verify accuracy after fitting the model.
input_x_train, input_x_test, label_x_train, label_x_test = train_test_split(input_data, label_x, test_size=0.2, random_state=5)
input_y_train, input_y_test, label_y_train, label_y_test = train_test_split(input_data, label_y, test_size=0.2, random_state=5)

# DTR model
DTR_x_model = ExtraTreesRegressor(max_depth=25)
DTR_y_model = ExtraTreesRegressor(max_depth=25)

# Training
DTR_x_model.fit(input_x_train, label_x_train)
```

```
DTR_y_model.fit(input_y_train, label_y_train)

# Prediction
predict_x_train = DTR_x_model.predict(input_x_train)
predict_x_test = DTR_x_model.predict(input_x_test)

predict_y_train = DTR_y_model.predict(input_y_train)
predict_y_test = DTR_y_model.predict(input_y_test)

# Training and testing accuracies
print('Training MSE X', mean_squared_error(label_x_train, predict_x_train))
print('Testing MSE X', mean_squared_error(label_x_test, predict_x_test))

print('Training MSE Y', mean_squared_error(label_y_train, predict_y_train))
print('Testing MSE Y', mean_squared_error(label_y_test, predict_y_test))

# Dataset accuracy
x_prediction = DTR_x_model.predict(input_data)
y_prediction = DTR_y_model.predict(input_data)

print('------------------------------------------------------------------------------------')
print('X MSE: ', mean_squared_error(label_x, x_prediction))
print('Y MSE: ', mean_squared_error(label_y, y_prediction))

print('X r2 ', r2_score(label_x, x_prediction))
print('Y r2 ', r2_score(label_y, y_prediction))

results = {'DTR_prediction_X':list(x_prediction), 'DTR_prediction_Y': list(y_prediction)}
PR_results_df = pd.DataFrame(results, columns=['DTR_prediction_X', 'DTR_prediction_Y'])
```

```python
PR_results_df.to_csv('DTR_results.csv')
import numpy as np
import pandas as pd
from sklearn.model_selection import train_test_split
from sklearn.ensemble import RandomForestRegressor
from sklearn.metrics import r2_score,mean_squared_error

data_frame = pd.read_csv("Data3.csv")

input_data = data_frame[['RSSI1', 'RSSI2', 'RSSI3']]
label_x = data_frame[['X_Actual']]
label_y = data_frame[['Y_Actual']]

label_x = label_x.values.ravel()
label_y = label_y.values.ravel()

# Split data into train and test to verify accuracy after fitting the model.
input_x_train, input_x_test, label_x_train, label_x_test = train_test_split(input_data, label_x, test_size=0.2, random_state=40)
input_y_train, input_y_test, label_y_train, label_y_test = train_test_split(input_data, label_y, test_size=0.2, random_state=40)

# Random Forest Regressor model
RFR_x_model = RandomForestRegressor(n_estimators = 100, random_state = 42)
RFR_y_model = RandomForestRegressor(n_estimators = 100, random_state = 42)

# Training
RFR_x_model.fit(input_x_train, label_x_train)
RFR_y_model.fit(input_y_train, label_y_train)
```

```python
# Prediction
predict_x_train = RFR_x_model.predict(input_x_train)
predict_x_test = RFR_x_model.predict(input_x_test)

predict_y_train = RFR_y_model.predict(input_y_train)
predict_y_test = RFR_y_model.predict(input_y_test)

# Training and testing accuraciss
print('Training MSE X', mean_squared_error(label_x_train, predict_x_train))
print('Testing MSE X', mean_squared_error(label_x_test, predict_x_test))

print('Training MSE Y', mean_squared_error(label_y_train, predict_y_train))
print('Testing MSE Y', mean_squared_error(label_y_test, predict_y_test))

# Dataset accuracy
x_prediction = RFR_x_model.predict(input_data)
y_prediction = RFR_y_model.predict(input_data)

print('-------------------------------------------------------------------------------------')
print('X MSE: ', mean_squared_error(label_x, x_prediction))
print('Y MSE: ', mean_squared_error(label_y, y_prediction))

print('X r2 ', r2_score(label_x, x_prediction))
print('Y r2 ', r2_score(label_y, y_prediction))

results = {'RFR_prediction_X':list(x_prediction), 'RFR_prediction_Y': list(y_prediction)}
RFR_results_df = pd.DataFrame(results, columns=['RFR_prediction_X', 'RFR_prediction_Y'])

RFR_results_df.to_csv('RFR_results.csv')
```